# Personalizing image enhancement for critical visual tasks: improved legibility of papyri using color processing and visual illusions

**Vlad Atanasiu [1, 2] and Isabelle Marthot-Santaniello [1]**

University of Basel [1] and University of Fribourg [2], Switzerland, atanasiu@alum.mit.edu, i.marthot-santaniello@unibas.ch

**Abstract**

*Purpose:* This article develops theoretical, algorithmic, perceptual, and interaction aspects of script legibility enhancement in the visible light spectrum for the purpose of scholarly editing of papyri texts. — *Methods:* Novel legibility enhancement algorithms based on color processing and visual illusions are compared to classic methods in a user experience experiment. — *Results:* (1) The proposed methods outperformed the comparison methods. (2) Users exhibited a broad behavioral spectrum, under the influence of factors such as personality and social conditioning, tasks and application domains, expertise level and image quality, and affordances of software, hardware, and interfaces. No single enhancement method satisfied all factor configurations. Therefore, it is suggested to offer users a broad choice of methods to facilitate personalization, contextualization, and complementarity. (3) A distinction is made between casual and critical vision on the basis of signal ambiguity and error consequences. The criteria of a paradigm for enhancing images for critical applications comprise: interpreting images skeptically; approaching enhancement as a system problem; considering all image structures as potential information; and making uncertainty and alternative interpretations explicit, both visually and numerically.

*It's obvious — any fool can see it.*
Homer, *The Iliad,* 7.464 [77: 227]

## 1 Introduction

Of the images below, which would you consider more legible,

A. 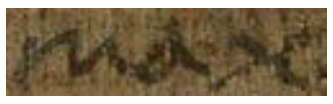 or B. 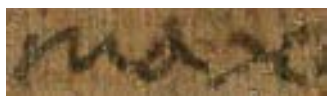 ?

If you selected B, you behaved like an engineer, confident in the benefits of a high signal-to-noise ratio. It may, therefore, come as a surprise that, in the evaluation reported in this article, three out of eight scholars deciphering these ancient Greek papyri preferred item A, because it is the original document, and originals should not be altered. This is not unreasonable: to question what exactly one is looking at is sensible. We call this attitude *critical vision,* and place it at the core of image enhancement. Our psychophysical and sociological mini-experiment also illustrates how enhancement quality depends on more factors than algorithms alone. These two principles frame the work being presented here.

*Objectives* — In narrow terms, the scope of this article is the legibility enhancement in the visible light spectrum of papyri documents for scholarly text editing. The broader objectives derive from the challenges, approaches, and lessons of this work. On the technical level, the aim is to demonstrate the utility of color processing and visual illusions in legibility enhancement. To emphasize the power of these approaches, the proposed methods will rely on a minimal algorithmic apparatus. On the theoretical level, a paradigm of legibility enhancement for computer-aided vision will be outlined. The central tenet of this paradigm is that critical vision tasks require enhancement technologies capable of making uncertainty explicit and adopting a systemic approach, that encompasses users, tasks, data, methods, tools, and interactions. The methodological objective is to develop methods and paradigms on an empirical, interdisciplinary basis.

*Relevance* — Papyri make up an extensive but underexploited trove of time capsules. The vast majority of these artifacts are accessible only as images taken in visible light, and are often characterized by poor legibility due to degraded physical documents, non-calibrated imaging, and image compression. Digital enhancement promises to reduce the time and uncertainty associated with reading difficult texts from difficult reproductions. This work may be useful in application domains other than papyrology as well, as image enhancement is a ubiquitous stage in image processing. For example, there are similarities between deciphering papyri and examining medical, forensic, and military imagery, with respect to the uncertainty surrounding their interpretation.

*Contributions* — The present article adds to the limited existing work on papyri enhancement in the visible light spectrum. It promotes the use of color processing and computationally-induced visual illusions. It works towards remedying the lack of a holistic approach to legibility enhancement. Making uncertainty explicit is also novel in this domain.

*Methodology* — The theoretical argument made in this article, the evaluation of the proposed methods, and the design of the ensuing software rely on an in-depth exploratory statistical analysis of data obtained from a user experiment. However, mathematics, algorithms, software, and hardware are never neutral when used by humans; this is especially so when the machine and the human form a closely coupled system, as in the case of legibility enhancement. Therefore, this article presents various points of view regarding how psychological and sociological factors are co-determinant in the design of legibility enhancement methods. The user-centric and context-aware approaches promoted herein are complementary to purely data-driven methods [162].

*Applications* — The principal application is an aid to human vision for use in transcribing papyri. The developed legibility software is also a tool for publishing better papyri images in print and online. Machine vision, particularly document binarization and text recognition, is the third potential application or the proposed methods, although human-readable texts are not necessarily legible to machines.

*Organization* — Section 2, "Problematic", introduces papyrology and scholarly text editing, then defines legibility enhancement trough the concept of critical vision. Section 3, "Related work", reviews the computational methods of image enhancement used in papyrology and for similar tasks in other domains. Section 4, "Methods", presents the novel legibility enhancement methods. Section 5, "Experiment", describes a user experiment to empirically understand critical legibility enhancement. Section 6, "Paradigm", discusses a type of image enhancement that makes uncertainty explicit, and summarizes the criteria of a computer-aided critical vision system. Section 7, "Implementation", shows how the paradigm and empirical findings were translated in software.

Detailed technical information that is secondary for comprehending the main arguments in this article has been set in a smaller font size to facilitate readability. For color figures, please refer to the electronic version of this article. The image evaluation dataset is part of the article's online Supplementary Material. The software resulting from this research is freely available online [16].

## 2 Problematic

### 2.1 Relevance

*Papyrus* was the quintessential portable writing surface in the Mediterranean area for five thousand years, until the end of the first millennium CE, when paper technology spread. As such, papyrus is a crucial carrier of substantial knowledge about the Egyptian, Greek, Roman, Byzantine, and other civilizations of antiquity. Figs. 1 and 2 present surviving samples of two foundational texts of Western mathematics and literature: Euclid's *Elements* and Homer's *Iliad*. The number of unpublished papyri is currently estimated to be approximately one to one and a half million, which "will keep papyrologists busy for centuries at least" [181: 644–645].

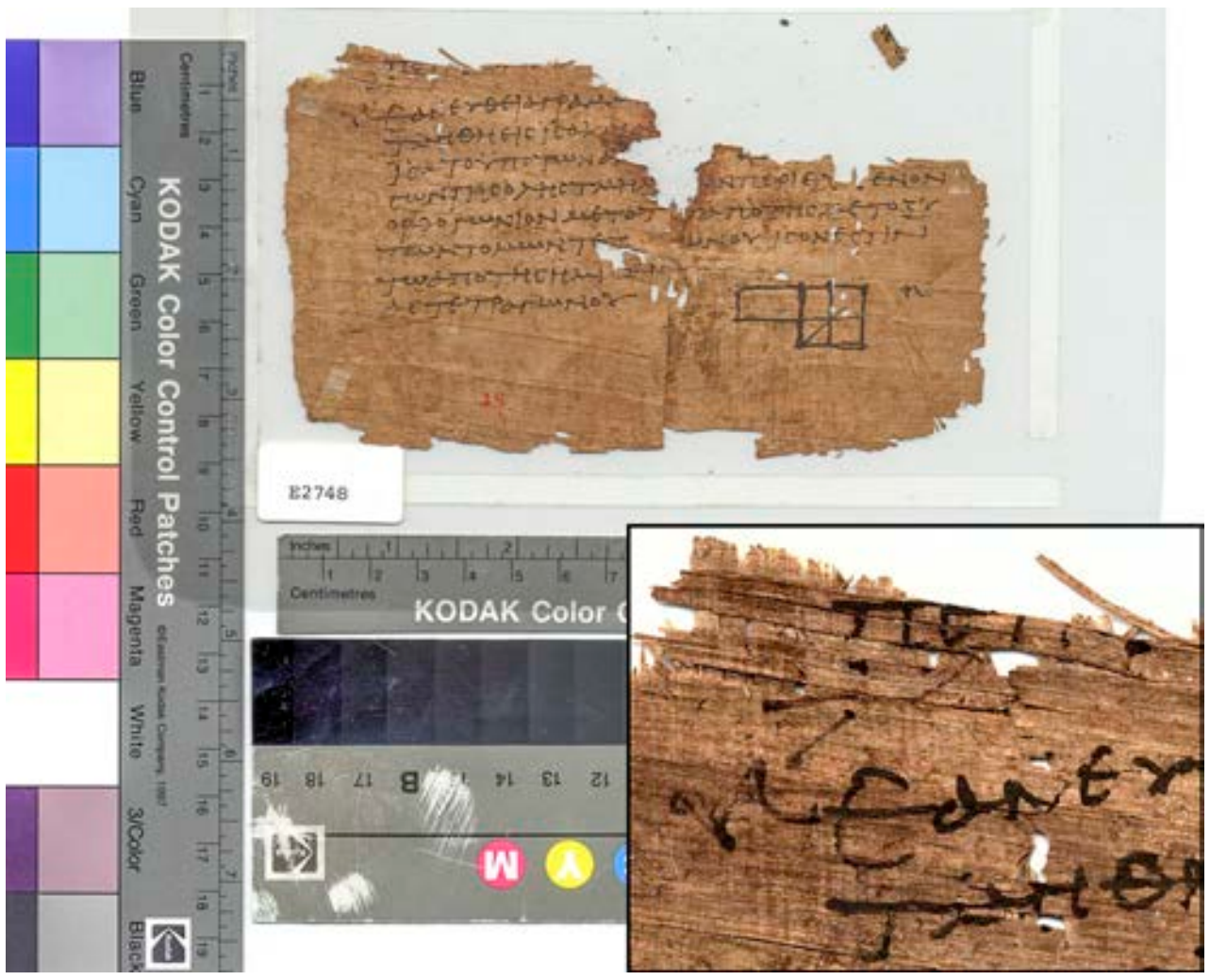

**Fig. 1** Euclid, *Elements*, Book II, Proposition 5; papyrus of the 3rd–4th century CE from Oxyrhynchus, Egypt. — Credit: P.Oxy. I 29, University of Pennsylvania Museum of Archaeology and Anthropology, CC-BY-2.5.

### 2.2 Data

*Cyperus papyrus L.* is an aquatic plant growing on the banks of the Nile River. By slicing, assemblage, and gluing via its own sap, a fine, flexible, and smooth papyrus can be manufactured, that is suitable for writing [34]. The orthogonal fiber pattern and surface roughness were once commercial criteria in the papyrus trade; today, they offer valuable hints to historians on prices, tastes, and document origins. However, the closeness in spatial frequency between papyrus texture and script complicates computational graphonomics. Reading performance is also affected, given the higher perceptual sensitivity to horizontal and vertical gratings [26: 270–271]. The spectral characteristics of papyri satisfy historical and computational aims, such as the recovery of ink traces by means of multispectral imaging. Further characteristic, inconvenient to scholars and scientist alike, include edge irregularity, holes, document fragmentation, and the presence of artifacts surrounding the papyrus, such as color charts, rulers, and shelf marks (Fig. 1).

### 2.3 Task

*Papyrology* is the study of papyri, although it also encompasses inscribed potsherds, wax tablets, metal foils, and other materials [17: xvii]. The aim of papyrology is to elucidate all aspects of the past, from the material to the spiritual. Its main source of information is the text content. *Text editing* is the process of transforming a poorly legible text in bitmap format into legible electronic text. This is not a purely mechanical task, as considerable paleographical, linguistic, and historical knowledge is required to construct a space of possible readings and then select one or more as the most probable. It involves more than "a pair of sharp eyes and a certain amount of common sense" as any person struggling with an unfamiliar script and text can attest, and it has been aptly compared to the work of a detective or a puzzle solver [161: 197, 199]. For papyrologists, every word counts, because a single misread character can, literally speaking, change a queen into a fishmonger or teleport a city across space and time. Therefore, variant readings are common and indexed, since 1922, in a dedicated publication [161: 212–213].

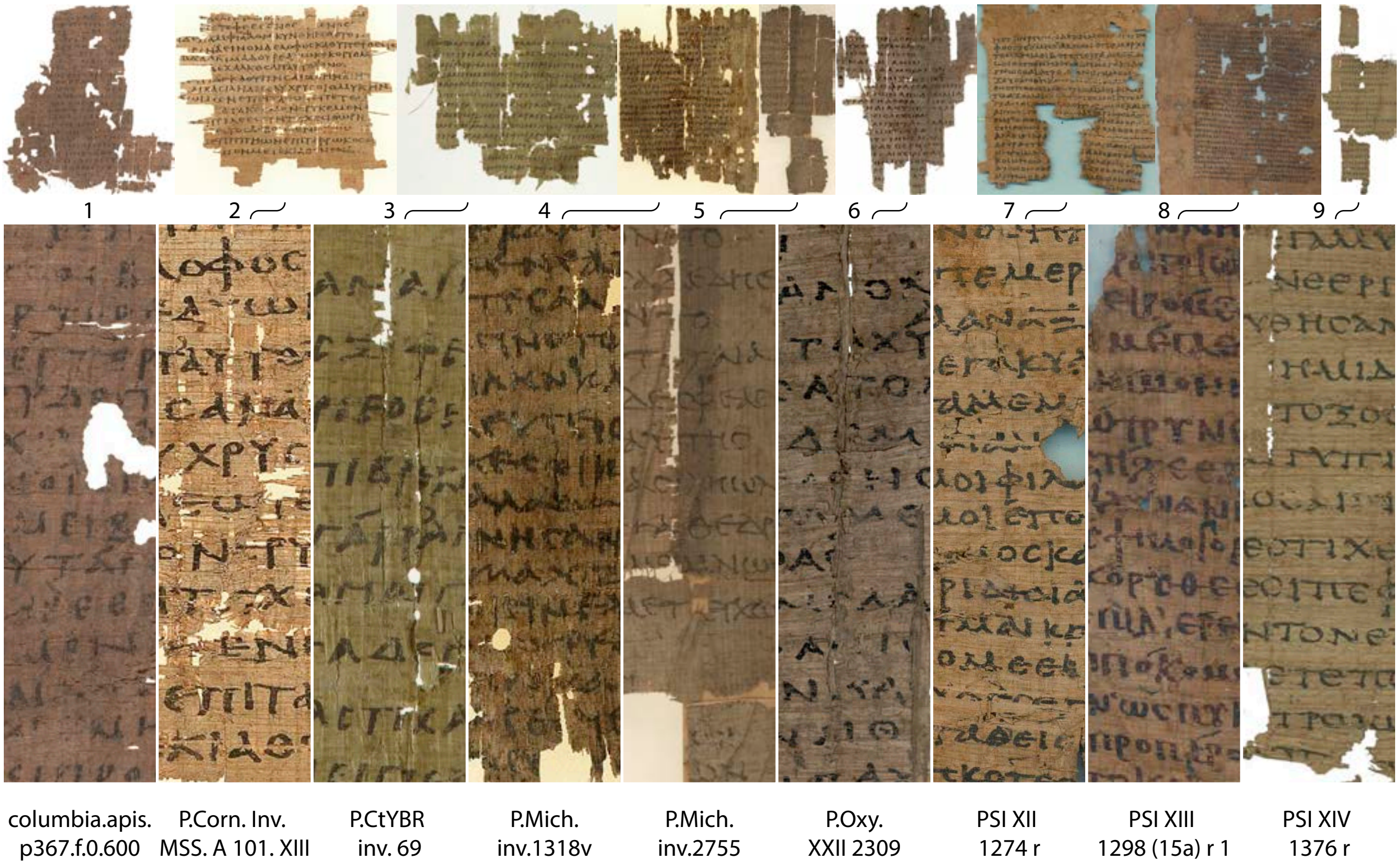

**Fig. 2** Overview and detail of the original papyri images used in the evaluation. For higher resolution images please refer to the article's online Supplementary Material. **Top:** Notice the degradation of the physical documents, the differences in image quality, and the difficulty in comparing details within a limited display space. **Bottom:** Details scaled to equalize character height, demonstrating the difficulty associated with deciphering such documents. — Credits: (1) Columbia University Library, CC BY-NC 3.0; (2, 4, 5) University of Michigan Library, Public Domain Mark 1.0; (3) Yale University Library, Public Domain Mark 1.0; (6) Courtesy of The Egypt Exploration Society and the University of Oxford Imaging Papyri Project; (7, 8, 9) Istituto Papirologico Vitelli, by permission.

The task under consideration can thus be defined as *critical vision* in (*a*) the objective sense that its outcomes are of *significant importance*, and (*b*) the subjective sense of an observer exercising *skepticism* in the interpretation of signals. Here, the observers are the text editors, who are circumspect about the reality and meaning of what is seen in images, and their readers, who doubt the editors' interpretation. The opposite of critical vision is *casual vision*, which is characterized by lower task criticality and levels of observer criticism. In psychological terms the latter is a form of fast, reflexive, and easy cognition (System 1), and the former of slow, analytical, and deliberative cognition (System 2) [83].

The two vision types call for different image processing paradigms. Casual vision is best served by binary images, designed to minimize entropy and produce trustworthy, unambiguous texts. Critical vision requires that ambiguity be maintained. In this case, no image structure is noise; every one is potential information. They provide context to develop concurrent readings and are valuable as archaeological layers of the document's history. The text on a crumpled paper, to illustrate the argument, wouldn't be identifiable as a draft if the creases would be removed for the sake of noise reduction.

## 2.4 Strategies

It is not unusual for papyrologists to take part in archaeological excavations, and, like archaeologists, to be sensitive to the context of discoveries and use various techniques to analyze data. A quote from experimental participant FRG illustrates the roles of *contextualization* and *diversification*:

> "I found it important to look at the *original*, or a normalized version of the original (*vividness*), sometimes with the *stretchlim* or *lsv* version of it for the bulk of the transcription work. To understand the state of preservation of the papyrus, the *histech*

> or *adapthistech* files have been very useful. The contrast is very good, and so I always use this type of image at the beginning of a transcription and [for] comparison with the original [...]. The *neg-vividness* and *neglsv* images have been useful to detect differences between ink and holes, so I've used them occasionally. *retinex* I expect to be useful, but not with the "Sammeltafeln" [different contents written on the same papyrus] we have in the Freiburg collection; while it is very good with large papyri, I would not use it for the transcription work throughout, only from time to time. *Locallapfilt* I have never found useful."

The user describes the creation of a coherent, rich, and justifiable interpretation based on multiple perspectives. Implicitly, he indicates that the legibility enhancement system must support the enhancement of a variety of image features, and that methods may be chosen for their good performance on a specific document and task. The strategy fits well with models of visual search [104] and information seeking [102]. In-depth user studies and theoretical models of papyrological reading strategies implemented in expert systems are discussed in [56, 155, 175, 176].

### 2.5 Duration

The text editing of papyri is a delicate and typically lengthy process, comprising document manipulation during reading, forming hypotheses regarding the content to achieve comprehension, and consulting auxiliary resources such as dictionaries. In addition to legibility, other factors affecting the amount of time required for this work include content difficulty, reader expertise, and work behavior. For example, it took the experienced participant FRG five to six hours split between two sessions to transcribe the papyrus used for training during evaluation, which was equivalent in length and legibility to the test documents presented in Fig. 2. The trainee OC, however, worked on and off for four to five weeks to transcribe a similar papyrus. AW also emphasized how, as time passed, he repeatedly modified his textual interpretation (or "belief", to use Reverend Bayes' terminology).

One implication is that dataset size and participant number is limited by necessity in papyrological legibility evaluations. A further consideration is the feasibility and appropriateness of experiments conducted in controlled laboratory settings. Not only is their organization complicated by their length, but the cognitive process of activity followed by rest phases is also impractical to reproduce. Under these constraints, the "homework" approach is more practical and realistic.

## 3 Related work

### 3.1 Papyrological practice

Three approaches to script enhancement characterize common papyrological practice: interactive image manipulation, spectral decorrelation, and imaging techniques [142, 35].

*Interactivity* — Image editors such as Adobe Photoshop allow easy and quick enhancement of papyri for a wide range of users. The flexibility of interactive image editing promotes solutions adapted to specific document qualities and user interests. Manipulations beyond the most basic tasks, however, require greater expertise, and are time-consuming.

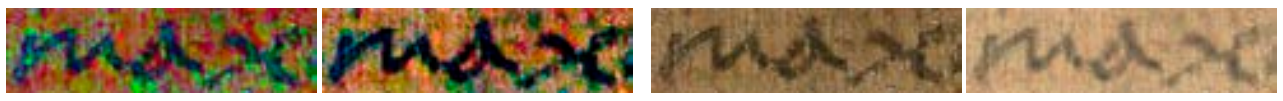

**Fig. 3** Enhancement with (left to right) DStretch options *rgb* and *lbk*, and the novel methods *vividness* and *lsv* (detail of papyrus 4 of Fig. 2).

*Decorrelation* — In the early 2000s, Jon Harman created DStretch, a software for enhancing pre-historical rock art from the American Southwest. This has since become a commonly used tool among papyrologists [71, 142]. DStretch decorrelates spectral bands by using either eigenvectors or principal components [8, 66]. The process can be performed in various color spaces, to highlight different features. An image is obtained with pixels that are ideally well-separated in terms of color values, and represented as pseudo-colors. The method yields excellent results for images with a Gaussian distribution, which is characteristic of inscriptions such as walls, potsherds, or stones. DStretch is also considered user-friendly by papyrologists and can be used in a field-work context, as a smartphone application. One limitation of the decorrelation stretch method is its poorer performance for images with a non-Gaussian distribution, as in the case of papyri reproductions. Moreover, the features emerging from decorrelation are not necessarily meaningful. Fig. 3 shows how DStretch induces both achromatic contrast and chromatic noise, in comparison with two of our novel methods.

*Imaging* — Most papyri images are obtained through color or monochrome photography; these are the images of principal interest in this article. However, the specific needs of paleography have also lead to the use or pioneering of many sophisticated techniques. Ultraviolet and infrared imaging, which can reveal ink traces in recycled and degraded documents invisible to unaided human vision, are used to analyze individual papyri [180, 60, 5]. With the advent of commercial narrow-band hyperspectral scanners, large-scale digitization has also become possible [94, 91, 41]. Reflectance Transformation Imaging retrieves the three-dimensional texture of the papyrus surface by illuminating it from a multitude of angles, thus providing a level of depth information that is inaccessible through classic photography [142]. X-ray and terahertz-based imaging have been used for the non-destructive reading of subsurface inscriptions in papyri and mummy cartonnage [105, 64]. These methods have been coupled with computed tomography methods for the virtual unrolling of the Herculaneum papyri, carbonized during the famous eruption of Mount Vesuvius in 79 CE [7, 112].

### 3.2 Computer science research

The authors have found very few articles that specifically examine the enhancement of papyri legibility from visible light images. Sparavigna [174] applies an edge detector based on the magnitude of dipole moments to increase the contrast of shape outlines, and thereby further enhancing the effect of a similar edge detection neural mechanism already present in the human visual system, known as the Mach bands [26: 362–363]. The anisotropic filtering of the image in the frequency domain was tested as a preprocessing step to suppress papyri fibers and gaps; unfortunately, this also removed character segments collinear with the fibers.

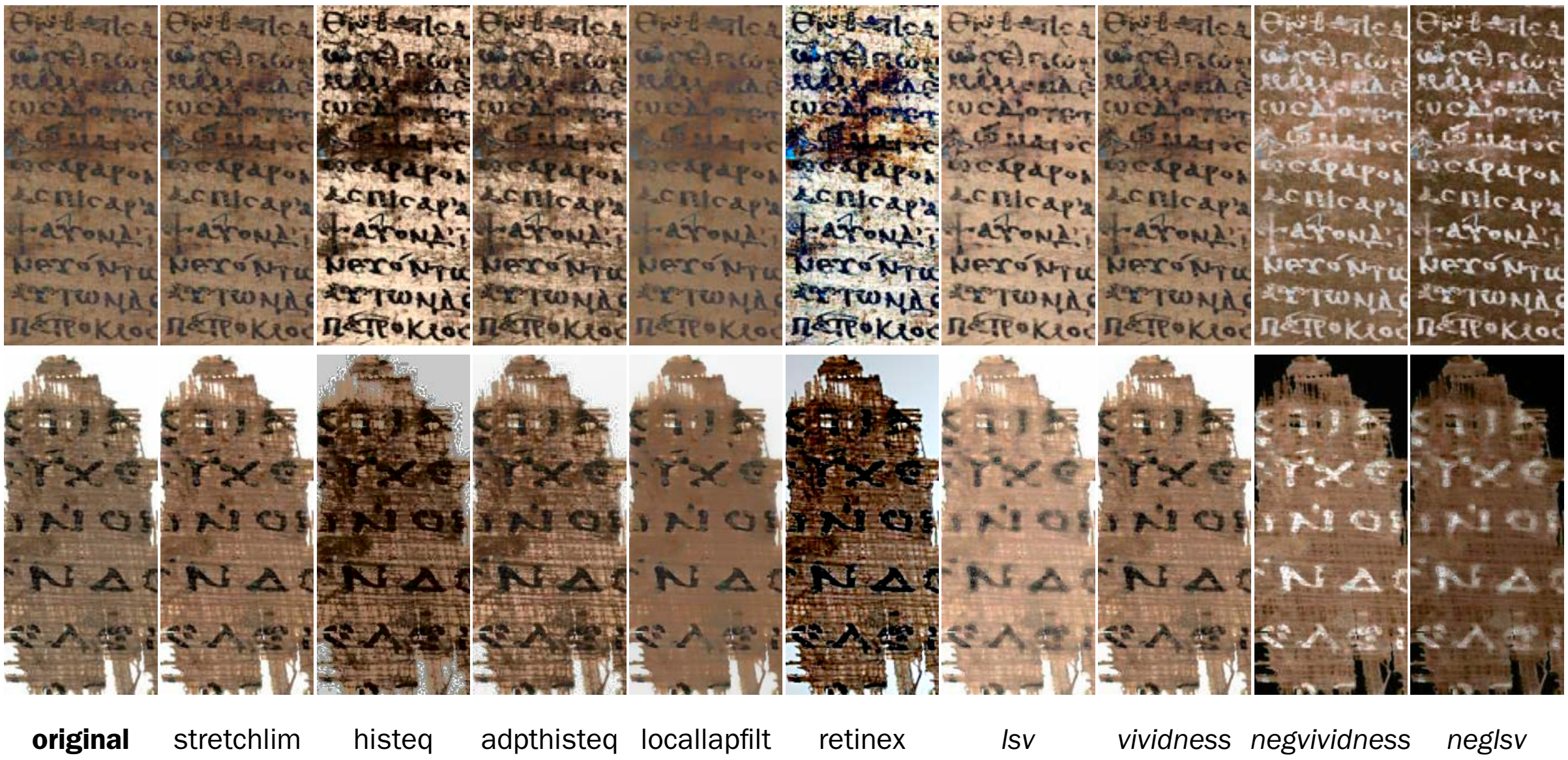

**Fig. 4** Enhancement results on papyri PSI XIII 1298 (15a) r 1 (top) and P.Oxy.XXII 2309 (bottom). From left to right: the original; five standard enhancement methods; and the new methods proposed in this article (italics). Note the different attenuation of the smudge (top row) and papyrus texture (lower row). The image labels refer to the eponymous methods described in Sections 4, "Methods", and 5.1, "Algorithms", and are abbreviations of (from left to right) original image, stretch limits (*stretchlim*), histogram equalization (*histeq*), adaptive histogram equalization (*adapthisteq*), local Laplacian filters (*locallapfilt*), retina and cortex (*retinex*), lightness, saturation, and value (*lsv*), vividness (*vividness*), negative vividness (*negvividness*), and negative lsv (*neglsv*). Retinex and vividness are established names in the literature, while stretchlim, histeq, adapthisteq, and locallapfilt carry over the function names given in the Matlab programming language, which was the development environment in the work presented herein. All ninety evaluation images are available in the article's online Supplementary Material.

The best papyri enhancements are obtained using multispectral imaging, as noted above in the papyrological practice overview. Moreover, in conjunction with multispectral images, color processing has also been used for papyri legibility enhancement specifically and for document processing more generally, as will be shown in the next section. We will here widen the survey to encompass other methods, topics, and applications of potential interest to this article, within the field of document processing and beyond, and conclude by discussing the applicability of machine learning.

Because noise — whether biological, mechanical, electrical, or digital — is a fundamental issue in communication, substantial research has aimed at improving document legibility by suppressing a great variety of what are usually considered manifestations of *noise*, including: ink bleed-through [46], see-through [48], foxing [163], shadows [95], termite bites [150], cross-outs [30], photocopied low-contrast carbon copies [36], low resolution raster images [141], and background texture interference [133] (for a history of image denoising, see [111]). The type of visual *media* can also dictate the typology of enhancement methods (e.g., methods for pictures and for movies differ in whether the time dimension is available as a source of contextual information for optimizing image processing) [24]. In applied science contexts, such as in the photography and video equipment industry, there is interest in developing enhancement methods predicated on an understanding of the *nature of noise*, e.g., optical, mechanical, and electronic noise sources in cameras [139]. Advances in image quality measurement [87, 191] have benefited from research into *visual perception* and *neuroscience* [187, 19], as well as models of *scene statistics* [188]. The role of *tasks* in image enhancement is of particular interest in cultural heritage applications [179]. The *systemic* and *critical* approaches to document analysis advocated in this article have been the subject of exemplary research over more than a half-century in two domains, each with specific aims, constraints, and solutions. First, the legibility of flight documentation and instrumentation plays a critical role in *aviation* performance and safety; here, optimization has been approached mainly through psychophysical experimentation [55]. Second, research on the enhancement of *medical* radiographic images stands out in terms of the extent to which the impact on clinical diagnosis of technologies and perception have been investigated, including the role of visual illusions [158, 23].

In addition to methods, prominent areas of focus in this field are datasets, benchmarks, and groundtruthing [117]. Visual confirmation of the attenuation of conspicuous artifacts is a typical means of comparing methods, supplemented with numerical characterization if feasible (i.e. given the availability of reference images or appropriateness of reference-free image quality measurements). User evaluations are rare, and vary from a few participants for historical documents [28, 12] to thousands in online campaigns for industrial applications [63, 144]. Legibility has been systematically studied since the early 20th century in experimental psychology (notably for the design of traffic signage and car plates, flight documentation and instrumentation, and typography for visually impaired [114, 14]), in works on teaching handwriting [159], as well as in relation to optical character recognition [68] and document image quality [4].

While the use of machine learning for image enhancement is well established [177], its application to document image enhancement, in particular historical, is still rare [72, 74, 130, 131, 115]. Notwithstanding results comparable to the state of the art, the challenges it faces are considerable. Historical data is scarce and heterogeneous, and groundtruthing is time-consuming, which hinders the generalizability of the learned models [54]. In respect to the task of critical vision, an implementable machine learning methodology has yet to be devised that can deal with images that are ambiguous, and their interpretations multiple, variable, or unknown. For scholarly applications, the resulting enhancements have to be, furthermore, trustworthy.

## 4 Methods

### 4.1 Justification

The proposed methods are rooted in color science and visual psychology, and thus stand in contrast to the mathematics- and data-driven methods typically utilized in the document analysis domain. This section presents the rationale for and benefits of these methods.

*Color image processing* is a highly interdisciplinary area of computer science, owing to the multifarious aspects of color, as diverse as electromagnetic, perceptual, genetic, and linguistic [62, 97]. The perception of shape as the prominent aspect of documents has resulted in color receiving moderate attention in document image analysis. However, if the image resolution is low, or the text and background interfere with each other, as often occurs with papyri, color processing becomes an excellent alternative to spatial and frequency domain processing. Application examples of color-based document processing include historical document image enhancement [118], denoising, and binarization [184], page object segmentation [9], as well text extraction from typographic documents [172], web images [86], and cartographic maps [183]. Fig. 3 illustrates the outcomes of the proposed and comparison methods, highlighting interactions between color and shape.

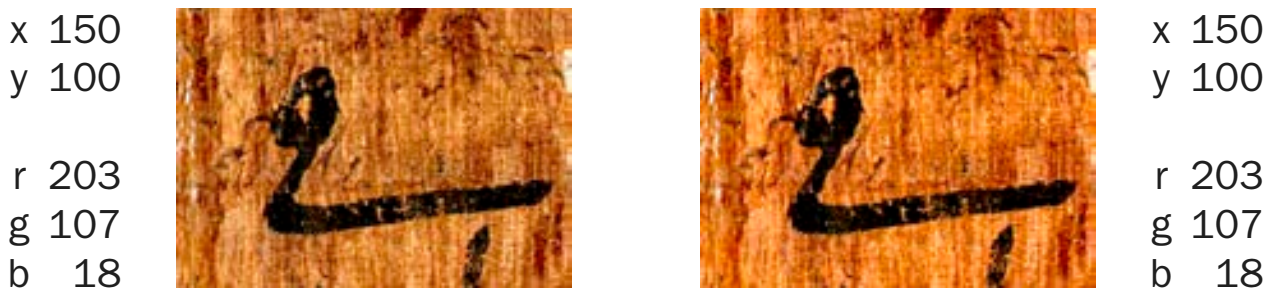

**Fig. 5** Color values at the same location in an image with sRGB (left) and Adobe RGB (1998) (right) embedded color profiles (detail of Fig. 7).

One far-reaching tenet of color research [84, 182] is that the technological, cognitive, and perceptual modalities through which visual reality is apprehended are not necessarily consistent with one another. In the negative polarity method, for example, physically and numerically identical gray-level differences are perceptually distinct. Because such effects, experienced as paradoxical, are the core of some proposed methods, this fact is emphasized through the use of the term *visual illusion* (see definitions, typologies, theories, history, and samples in [147, 164]). Effects appropriate for enhancement were chosen among color illusions, since geometrical illusions are difficult to elicit without degrading legibility, and that illusions of stereopsis, apparent motion, and temporal illusions are tiring to use for extended periods of work.

The use of color and illusions shifts the burden of image enhancement from the computer to the human visual system. This *perceptual image processing* approach makes possible to create algorithms that are both *mathematically straightforward*, and considered by the experimental participants useful. Researchers also have noticed that simplicity is beneficial to legibility enhancement quality [178: 404, 416], and the likelihood of implementation of methods in software and their use by the end-users [100: 920].

### 4.2 Chroma contrasting by gamut expansion

In Fig. 5, the picture on the right appears more saturated, and the text is more clearly separated from the background. Yet the color values of both image files are identical. The perceptual difference results from the files being tagged with different color profiles, meaning that the same values represent different locations in the color space. The embedded profile of the left image is sRGB IEC 61966-2.1, a standard designed for displays with a limited amount of reproducible color, as were typical in the 1990s. The space defined by these colors is called a color *gamut* (Fig. 6). The larger Adobe RGB (1998) gamut was designed to meet the needs of photographers and high-fidelity reproduction. Humans perceive far more colors than are present in either of these gamuts.

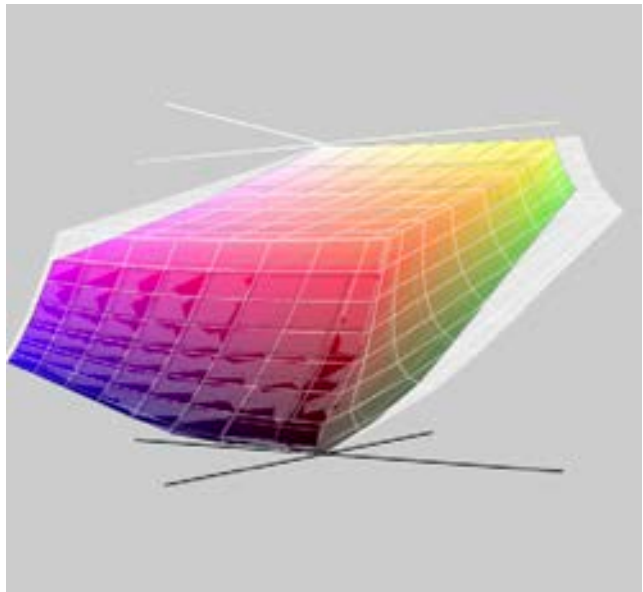

**Fig. 6** The sRGB color gamut volume represented within the Adobe RGB (1998) gamut (gray hull) in the CIELAB color space. The vertical dimension represents lightness, $L^*$, while the horizontal axes correspond to the red–green, $a^*$, and the blue–yellow, $b^*$, opponent colors.

Numbers alone do not define colors; a coordinate system is also necessary. When a color space is replaced by a wider one, the gamut expansion increases the detectability of differences without affecting the values. The elegance of this legibility enhancement method resides in not requiring any numerical manipulation from software developers.

The underlying psychophysics is based on stronger chroma increasing the perceived level of luminance (*brightness*), which is known as the Helmholtz–Kohlrausch effect [52, 45]. This effect is leveraged as an unintended application of gamut mapping to improve the legibility of papyri.

A color space assignment and perceptual effects occur whenever an image file is read and visualized. Therefore being aware of their impact on legibility is important. In the case of papyri, the issue is compounded by so few reproductions having embedded color spaces. The most widespread space, sRGB, is assumed to apply in such situations.

The work of converting numerical values to visible color is left to the color management system of the image viewers, operating systems, and displays. For several reasons, transforming the numerical values to the destination color space may be desirable, instead of solely specifying it. For example, applications might disregard color space profiles, and thus deprive users of the enhancement effect. Nonetheless, the transforms among color spaces (*gamut mapping*) are user- and application-specific. The "*perceptual intent*" shifts out-of-gamut (OOG) color values to in-gamut values, thereby preserving perceptual color appearance; by contrast, the "*colorimetric intent*" strives to preserve numerical fidelity, usually by clipping outlier OOG values to the destination's gamut boundaries. This is a key topic in color management research [153, 165]. The legibility enhancement method suggested here can be made more robust by using standardized color conversion profiles, such as developed by the International Color Consortium (ICC). The relevant code in the Matlab programming language is as follows:

```
source_profile = 'sRGB Profile.icc';
destination_profile = ‘AdobeRGB1998.icc’;
C = makecform(‘icc’, destination_profile, source_profile);
I = imread('input_image.tif');
J = applycform(I, C);
imwrite(J,'output_image.tif', 'ColorSpace', 'rgb')
```

The results shown in Fig. 7 exhibit the change in chroma and color distribution. On the left is the original image with sRGB profile and its color distribution, and on the right the image after conversion to Adobe RGB (1998).

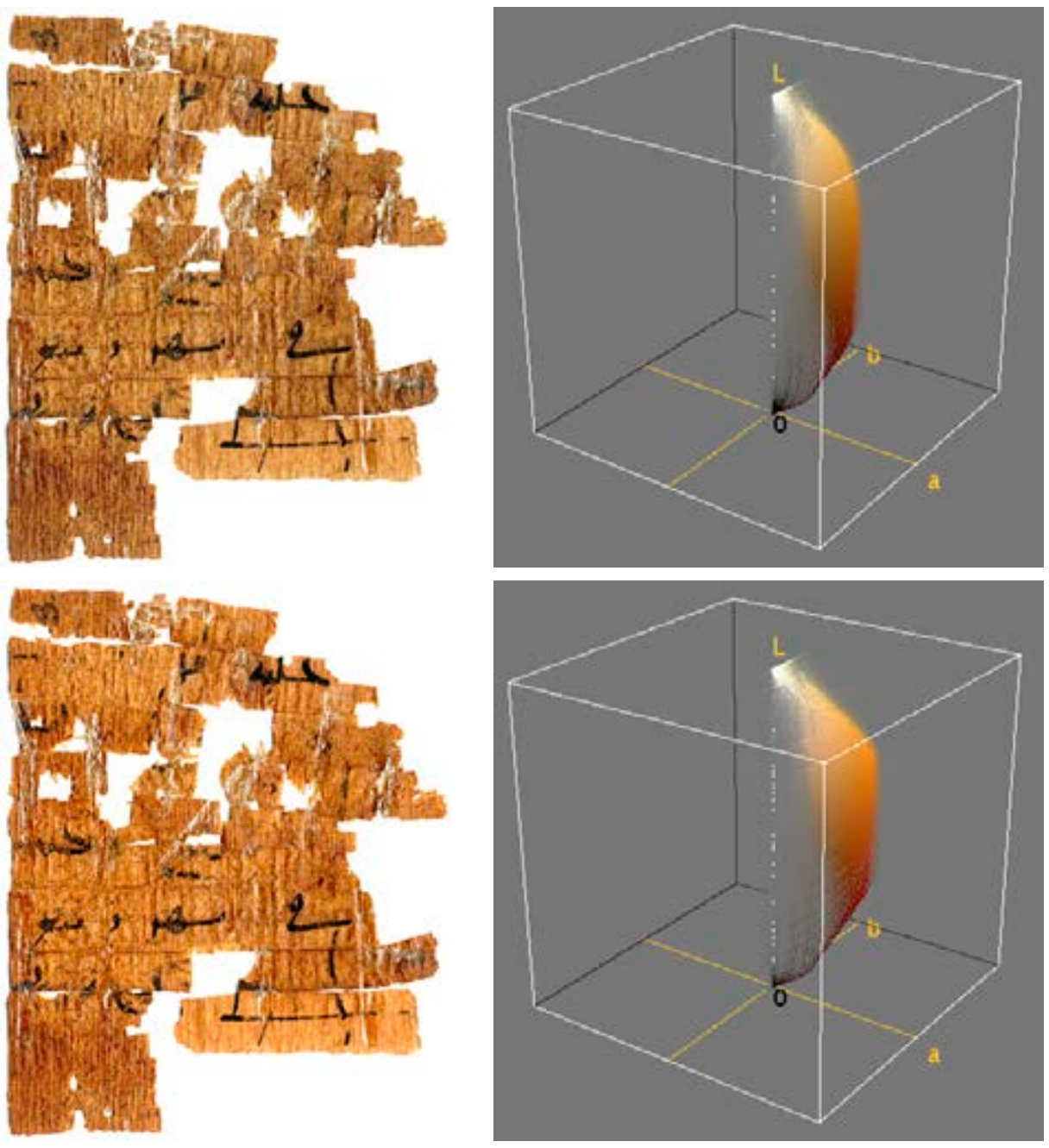

**Fig. 7** Color distribution of a papyrus image with native sRGB color profile (top), and following conversion to Adobe RGB (1998) (bottom). — Credits of papyrus image: Sorbonne Université, Institut de Papyrologie, P.Sorbonne Arabe 201 1500a.

## 4.3 Lightness contrasting by stretching

*Lightness* refers to perceived reflectance, and is an achromatic component of color, together with brightness [157: 2766–2767]. It is also a principal source of information about shapes for the human visual system, as can be ascertained via image decomposition into achromatic and chromatic colors:

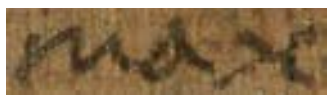 = 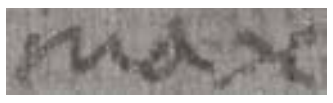 + 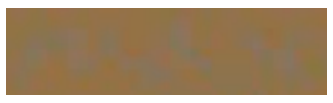 .

Hence, increasing the lightness contrast substantially improves legibility. The enhancement formula (*stretchlim*) is

$$L^{*\prime} = [L^* - \min(L^*)] \,/\, [\max(L^*) - \min(L^*)]\,, \quad (1)$$

where $L^*$ is the lightness in the CIELAB color space, bound to the [0, 100] range and corresponding to the approximate number of perceptually noticeable lightness levels under optimal conditions [51: 24, 202]. CIELAB was developed by the Commission Internationale de l'Éclairage (CIE) as a perceptually uniform color space allowing for the quantitative description and manipulation of color along fundamental phenomenological dimensions, such as lightness, chroma, and hue. Its perceptual proprieties often make CIELAB a better candidate for color processing than the Red, Green, and Blue (RGB) color space of common digital images. CIELAB is not without its shortcomings and more accurate color appearance models exist and are developed [166; 51: 201–210]; however, for applications where faithful color reproduction is not the main goal, CIELAB remains adequate. The conversion between RGB and CIELAB is parametrized by a triplet that specifies the illuminant of the scene reproduced in the image [40]. D65 (noon daylight, 6504 K) provided in practice the greatest improvement to papyri legibility.

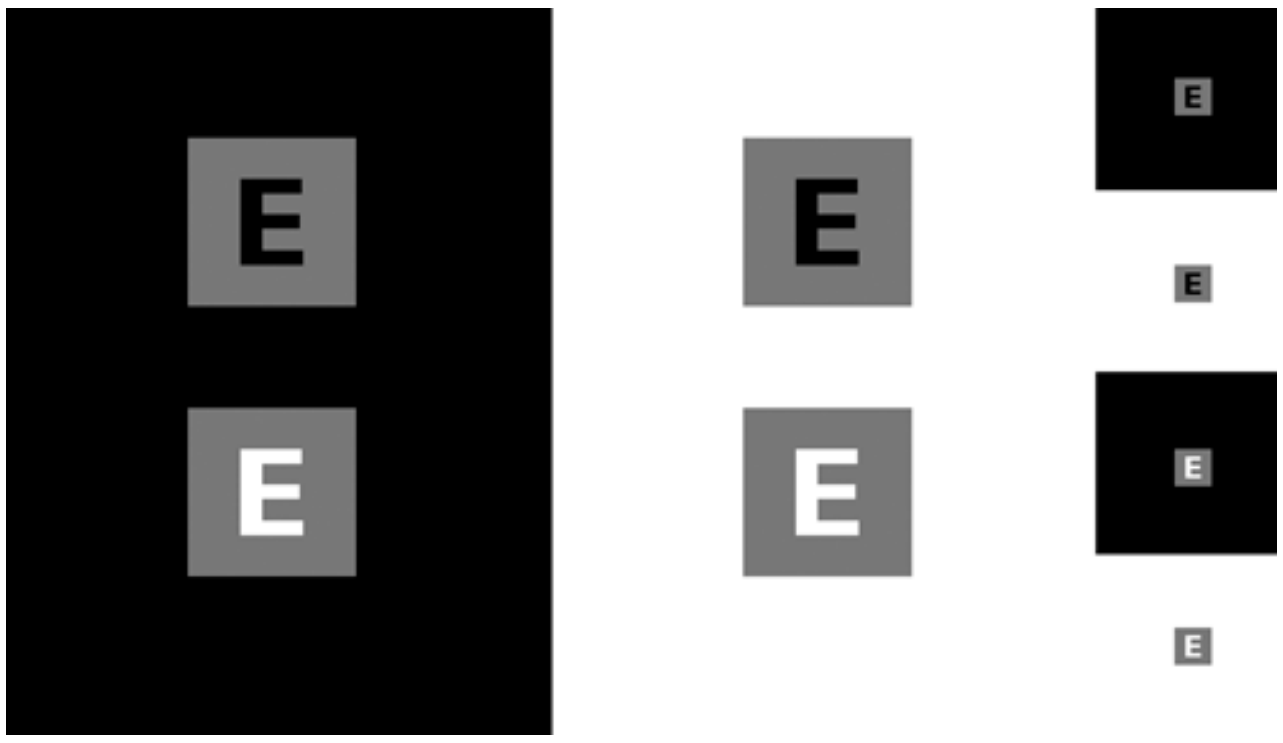

**Fig. 8** Legibility varies with foreground/background and background/surround polarity.

## 4.4 Lightness contrasting by negative polarity

The characters in Fig. 8, with CIELAB lightness values of 0 (black) or 100 (white), are set on squares of value 50 (gray). Despite identical numerical difference between figure and ground, the perceived contrast is stronger for light-on-gray configurations (bottom) and weaker for dark-on-gray (top); dark surrounds (left), and diminishing the stimuli (right), increase the effect. This occurs despite the perceptual linearity of CIELAB lightness and the use of calibrated displays.

This contrast illusion, known as *lightness induction,* is modulated by factors including asymmetrical polarity gain, non-linear focus/surround contrast gain, and spatial frequency [156; 157; 26: 354–356, 358–359, 370–371]. It belongs to the broader category of *color constancy*, the neural and ecological basis of which has been a matter of debate since antiquity [65, 2]. Furthermore, the *irradiation effect*, well-known to astronomers, makes light figures in dark surrounds (e.g. stars in the night sky) appear larger [190, 143].

Taken together, these effects let us posit that reversing the polarity of the image (i.e., so that the script appears lighter), will improve legibility. The empirical testing of this hypothesis as applied to papyri images is reported in Section 5, "Experiment". The formula for negative polarity (*neg*), $L^{*\prime}$, is

$$L^{*\prime} = 100 - L^{*}, \qquad (2)$$

where $L^*$ is the image lightness in the CIELAB color space.

The method is appropriate for papyri, a document type of medium lightness inscribed with dark ink, and typically imaged on a light background (see Fig. 2).

## 4.5 Selective contrasting by vividness colorization

*Chroma* is a dimension of the CIELAB color space that is orthogonal to lightness and hue (Fig. 9). Chroma modification effects a change between achromatic and chromatic color. Another dimension is *vividness*, which defines a concomitant change in lightness and chroma [22]. Vividness can be used as a model of how a dark ink film applied to the papyrus reduces both the amount of reflected light, thereby changing the lightness, and acts as an achromatic filter. Thus, vividness helps to distinguish ink from papyrus better than lightness. They can be substituted one for the other to improve legibility. Enhancement based on vividness (*vividness*) preserves the papyrus appearance, because the chroma and hue values remain unchanged (Fig. 3). The process differs from stretchlim and gamut expansion in that the transform is non-linear and selective: that is, more vivid colors are more strongly emphasized, while achromatic color values do not change. The enhanced lightness, $L^{*\prime}$, is given by

$$L^{*\prime} = (L^{*2} + a^{*2} + b^{*2})^{-1/2}, \qquad (3)$$

where $L^*$ is the lightness, and $a^*$ and $b^*$ are the chromatic components in CIELAB ($L^{*\prime}$ values exceeding 100 are clipped). The mathematical expression of vividness corresponds to the $l^2$-norm of the CIELAB image values.

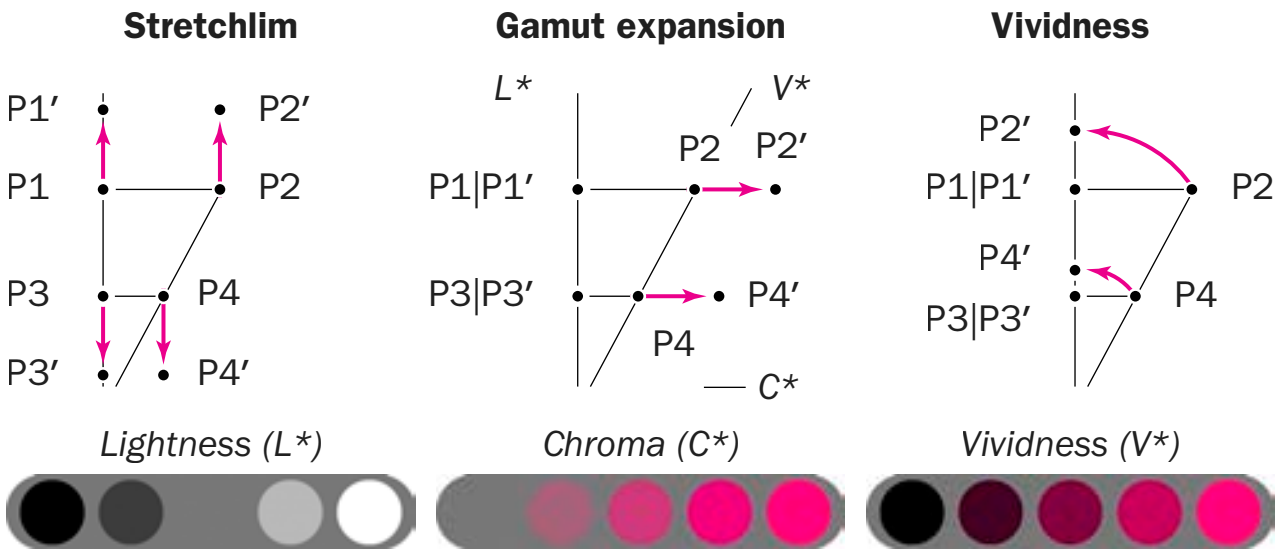

**Fig. 9** Comparison of three enhancement methods described in this article. The vertical diagram axis represents the CIELAB lightness, $L^*$, the horizontal axis the chroma, $C^*$, and the diagonal the vividness, $V^*$. The variation of color along these axes is illustrated in the bottom part of the figure. For each enhancement type, the diagrams show the location of four exemplary points in a vertical section of the CIELAB color space before and after enhancement. Thus, stretchlim displaces all points towards the extremities of the lightness axis; gamut expansion displaces the points along the chroma axis (except for those situated directly on the lightness axis); finally, vividness applies a rotation transform resulting in all points being collinear with the lightness axis.

Because the vividness method modifies the image in a pixel-wise fashion, it increases the distinguishability of ink and papyrus even in the presence of objects such as color charts, which complicate the use of more sophisticated methods such as principal component analysis [27].

## 4.6 Background attenuation by difference of saturation and value

The hue, saturation, and value (HSV) color space is not a perceptual color space like CIELAB, rather, it is a simple mathematical transformation of the RGB color space [170]. Nonetheless, in HSV the distributions of saturation and value in papyri images roughly overlap (Fig. 10). On the basis of a simple difference between them, it is possible to attenuate the non-inked papyrus surface and, thus, enhance text legibility. Because the HSV value only approximates lightness, we fuse the difference of saturation and value with CIELAB lightness (e.g., by taking the mean) and accordingly obtain an image with sharper edges (Fig. 11).

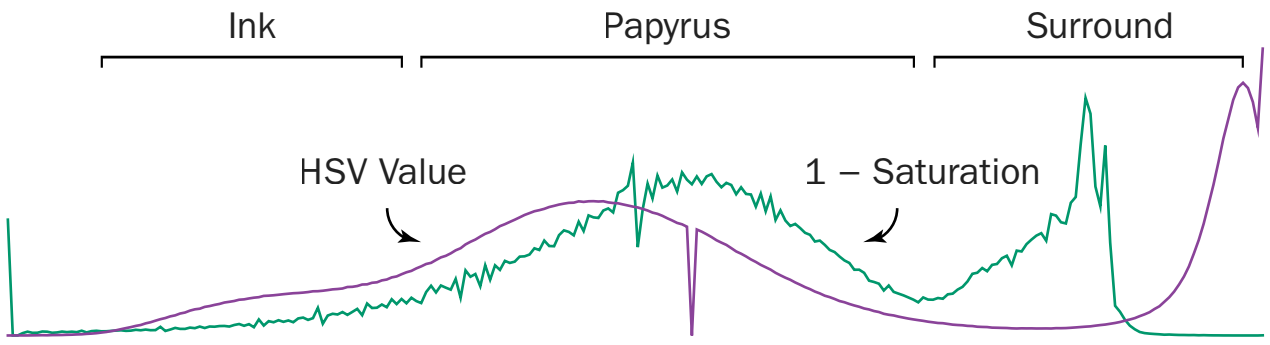

**Fig. 10** The figure depicts an example of the color profile of a papyrus in the HSV color space (the "value" channel and the complement of the "saturation" channel), along with the object classes dominating specific bands of the spectrum (ink pixels, non-inked papyrus surface, and the imaging area surrounding the papyrus).

The algorithm (*lsv*) represents a blind separation of a mixture of distributions. It consists in replacing the CIELAB lightness, $L^*$, by pixel-wise operations with $L^{*\prime}$, as follows:

$$L^{*\prime} = 100\,{}_0\lfloor 1 - ({}_0\lfloor 100 - L^* \rceil^1 + {}_0\lfloor V + S - 1 \rceil^1)/2 \rceil^1 , \qquad (4)$$

$$V = \max(R, G, B)\,, \qquad (5)$$

$$S = [V - \min(R, G, B)] \,/\, V\,, \qquad (6)$$

where $L^{*\prime}$ is the enhanced lightness; $V$ and $S$ are the value and saturation image channels in the HSV color space, derived from the $R$, $G$, and $B$ values of the RGB color space; and ${}_0\lfloor \cdot \rceil^1$ denotes normalization to the [0, 1] range.

## 4.7 Hue contrasting by hue shift

*Rationale* — The spectral sensitivity of the human visual system is uneven: it peaks in the green band in bright light, and in blue–green in dim light; it is lowest in the red band [26: 28–29, 118–119, 208–209]. As papyri have a brownish tint, a shift towards hues of higher acuity would be expected to improve legibility. The images in Fig. 12 result from applying this rationale. The first is the original, the second has all chromatic information discarded, and the remainder have the means of their CIELAB hue shifted to 246°, 162°, and 24°, which correspond to the loci of the primary colors [50: 343].

In addition for stronger acuity, a host of perceptual effects combine to make blue a compelling target for a hue shift. The increase in perceived lightness effected by an increase in chroma, as described above, is lowest for yellow and stronger for blue [52]. Moreover, the visual field for blue is broader than that for both green and red, thereby allowing for better comparison between focus and surround [26: 108–109]. A blue text background has also been found to improve performance in analytic tasks [121]. Blue is furthermore robust to the most common colorblindness types [135].

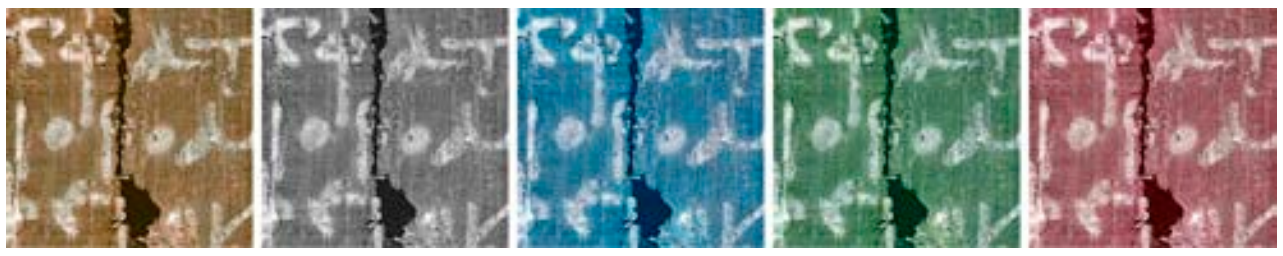

**Fig. 12** From left to right: original, lightness channel $L^*$, and hue-shifted papyri; all have negative lightness polarity (detail of papyrus 4 in Fig. 2).

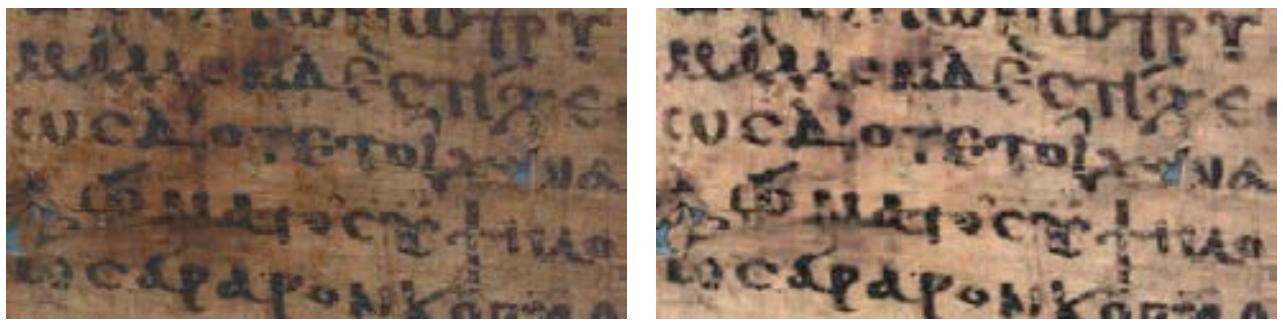

**Fig. 11** Papyrus before (left) and after (right) enhancement with the *lsv* method (detail of papyrus 8 in Fig. 2).

Hue shift can be profitably combined with lightness polarity reversal: (*a*) lightness induction reinforces hue contrasting; (*b*) the asymmetry of the color space geometry reduces the dynamic range of chroma in the blue direction with respect to yellow for lightness levels above the medium [127: 133–169].

Achromatic vision can resolve finer details than chromatic vision [128]. Because high spatial frequency is a characteristic of both script and papyri, using only the lightness channel for the decipherment task might make sense. However, the addition of color allows for ink to be better distinguished from stains, shadows, and other entities. Color vision in general improves image segmentation [69, 29].

*Implementation* — The shift of the papyrus towards blue (*blue* method) is accomplished via the following equation:

$$h_i{}' = h_i - h_p + h_b\,, \qquad (7)$$

where $h_i$ and $h_i{}'$ are the input and output hue in the cylindrical CIELAB color space, with $h = \tan^{-1}(b^*/a^*)$ [51: 204]; while $h_p$ is the centroid of the papyrus hue values and $h_b$ = 246° is the locus of blue in the hue dimension. An initial observation is that this is an ideal method, given the non-homogeneities in CIELAB. Second, the centroid of the papyrus hue is non-trivial to determine in the presence of other scene objects, such as color charts. Third, the location of maximum blue sensitivity varies across individuals, genders, ages, visual ecologies, and cultures, among other factors [49].

A simpler solution is available. This consists in changing the sign of the values in the a* and b* CIELAB channels:

$$a^{*\prime} = -a^* \quad \text{and} \quad b^{*\prime} = -b^*\,. \qquad (8)$$

This is made possible because the brown hue of papyri has blue as its opponent color. It is this method, applied in conjunction with negative lightness, vividness, and gamut expansion, that was employed in the user evaluation. A comparison of the hue shift and sign change methods is shown in Fig. 13.

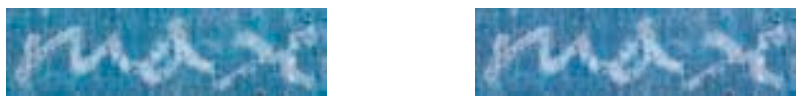

**Fig. 13** Hue shift (left) vs sign change (right), plus negative polarity (detail of papyrus 4 in Fig. 2).

### 4.8 Dynamic range increase with CIELAB retinex

The retinex theory (from *retina* and *cortex*) provides an explanation for the constancy of color perception across a wide range of illumination conditions. Essentially, it expresses lightness as the product of ratios of spatial neighborhoods. As an image enhancement technique, it reveals details in dark and light areas. Many variants have been proposed since Land's and McCann's original work, and the outcomes differ according to image quality, content, and intent [107, 119]. The process is typically performed on each color channel individually, or on a single intensity image derived from the three RGB values, as a substitute of lightness. Pursuing our thinking within the framework of color processing, we find that performing retinex in a perceptual color space (CIELAB) is beneficial to legibility enhancement. It results in a concomitant increase in contrast in several perceptual dimensions discussed in previous sections: lightness, saturation, vividness, and hue. Consequently, significant features may become visible. In Fig. 14, the presence of annotations in a different ink (☜) is striking in the CIELAB retinex process, and further details are revealed by the blue negative. The result, in terms of appearance, insights, and pitfalls, is analogous to watching the *Iliad* unfold in Technicolor rather than in black and white. The *cinematic metaphor of image processing* sums up the rewards and implications of color processing and visual illusions in support of critical vision.

*Methodology* — Conversion from RGB to CIELAB in the sRGB color space. Multiscale retinex with color restoration (MSRCR) [81, 140, 160] at three scales with Gaussian kernels of standard deviation 15, 80, 250, and a saturation of the 2.5% lowest and highest pixel values applied to the (*b*) RGB and (*c*) $V^*$ (vividness), $a^*$ (blueish–yellowish), and $b^*$ (greenish–reddish) dimensions. Output normalization of (*c*) to the [0, 100], [−128, 127], and [−128, 127] ranges, respectively, and back-conversion to RGB. (*d*) Blue negative applied after CIELAB retinex, with $V^{*\prime} = 100 - V^*$, $a^{*\prime} = -a^*$, $b^{*\prime} = -b^*$.

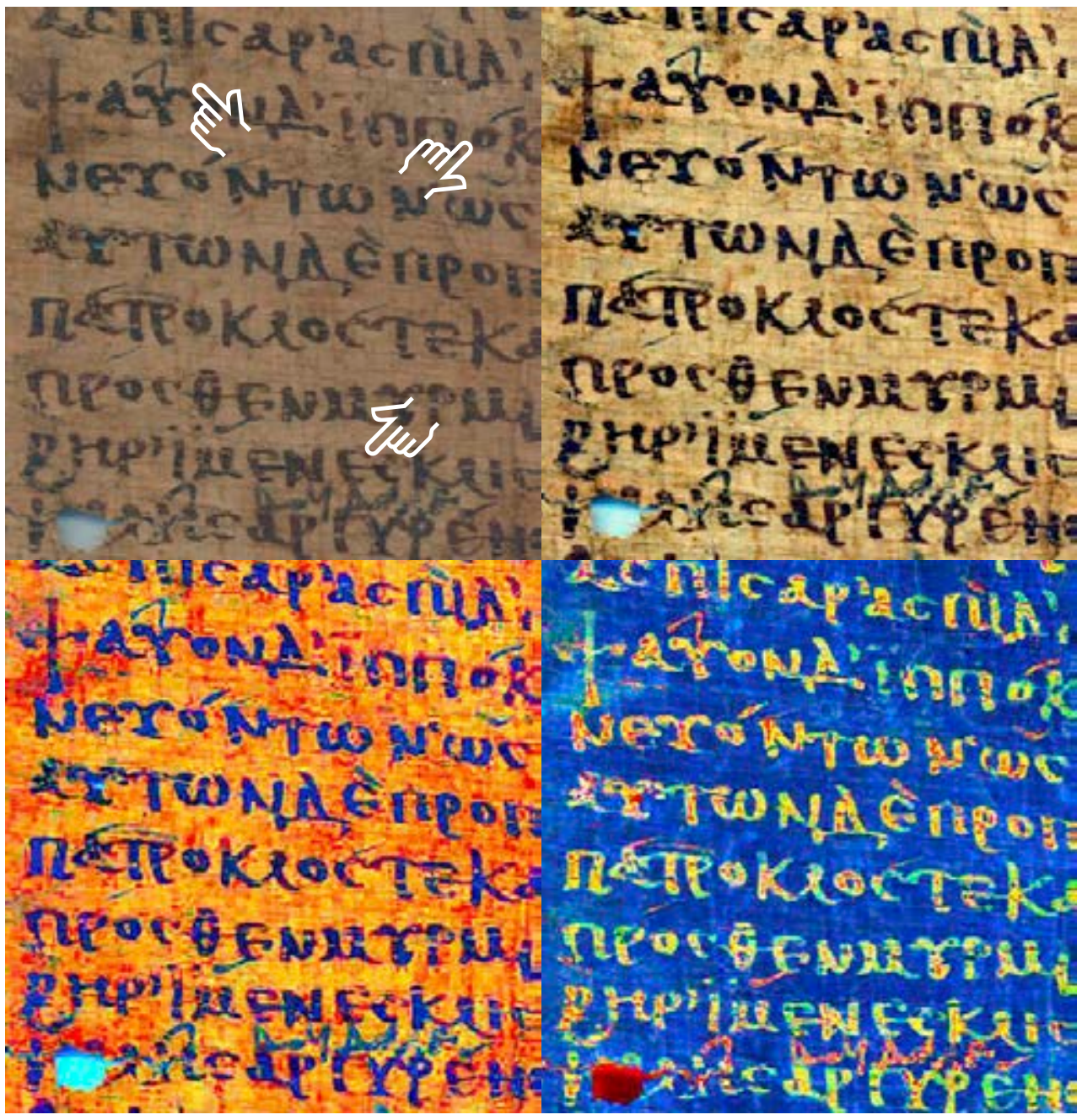

**Fig. 14** NW: Original image, with three faint annotations indicated by an index. NE: Enhancement by retinex in the RGB color space. SE: Retinex enhancement in the CIELAB color space. SW: Blue-shifted and negative of the SE image (detail of papyrus 8 in Fig. 2).

### 4.9 Cross-spectral colorization

Although the proposed methods are designed for visible light images, their use may be extended to multispectral images.

*Background* — Multispectral images consist of a set of single-channel images taken in various bands of the light spectrum; hence, each one is perceptually achromatic. Image fusion, which is a common additional processing step, is based on component analysis, wavelet transform, and other dimensionality reduction techniques [12, 123, 110, 80, 70, 173].

The visualization of multispectral images and images from bands outside the visible spectrum (VIS) may be improved through the use of color processing and by integrating multiple images into a single image. The problem, however, is that non-VIS images have no natural color correspondence. Moreover, the number of dimensions of multispectral images (dozens or even hundreds) usually surpasses that which is perceptible by the trichromatic human visual system.

Many single- and multi-channel "pseudocoloring" solutions have accordingly been developed [146, 186, 39, 194], ranging from analog optical [193] to digital frequency domain coloring [3] to the ubiquitous "colormaps" [1]. However, the perceptual intricacies that must be considered even when colormapping single grayscale images [154, 98] indicate considerable scope for additional research in this area.

In the document processing domain, the pseudocoloring of multispectral and non-VIS images has been used to increase the contrast between text and substrate. For example, George *et al.* [61] used the first three components of the principal and independent component analysis, respectively, of 31 spectral bands of an ostracon as input in the RGB color space. Easton *et al.* [47] inserted the ("invisible") ultraviolet-dominant grayscale image (long-wave ultraviolet light illumination) into the green and blue dimensions of the (visible) RGB color space, as well as the infrared-dominant grayscale image (tungsten illumination) into the red RGB dimension, thereby facilitating the legibility of the Archimedes parchment palimpsest. Gargano *et al.* [60] applied an overlay technique while working with papyri, which consisted in etching the binarized infrared image (or, alternatively, the red VIS channel) containing the document text into the visible light color image.

Our solution does not rely on binarization, which presents problems when applied to papyri [149] and may introduce artifacts and suppress critical contextual information during the interpretation of difficult texts. The rationale is that a single-channel image from the non-visible spectrum (non-VIS), or a fusion of several bands, contains features of interest that are not or poorly identifiable in the visible spectrum. Conversely, the VIS image will contain contextual information, supporting the interpretation of the image. Because the "end-user" of the visualization process is the human visual system, we use the perceptual CIELAB color space rather than the physical RGB space.

The procedure consists in transforming the VIS image from RGB to CIELAB, inserting the non-VIS image into the lightness dimension $L^*$, and then converting back to RGB.

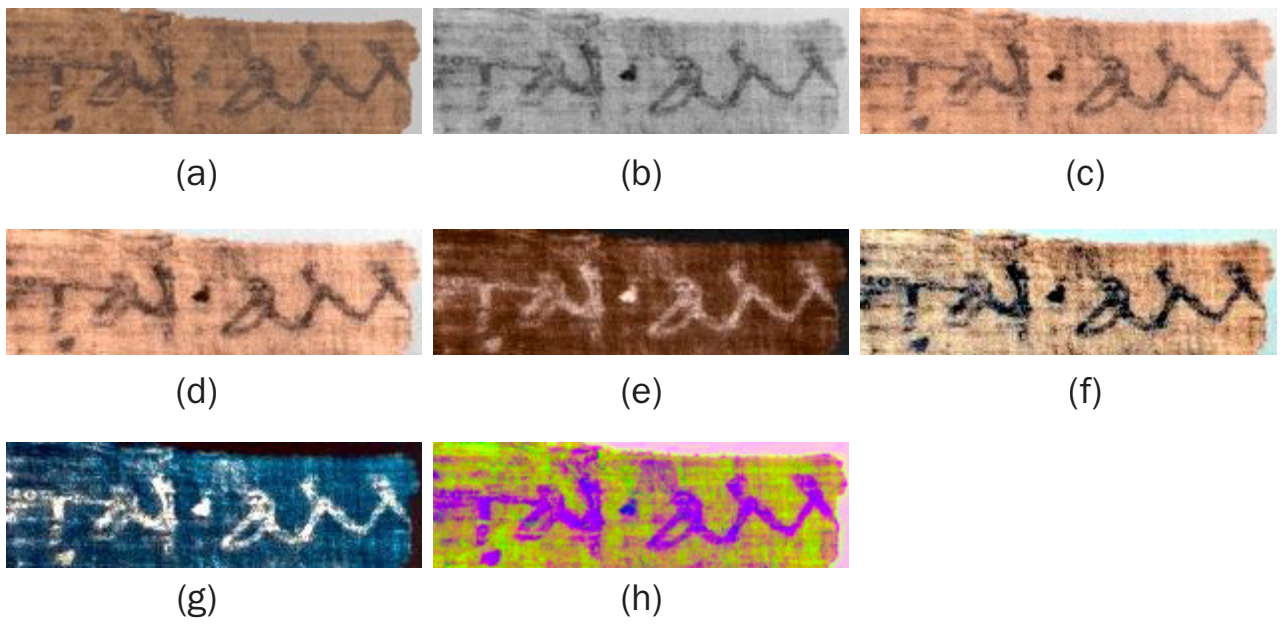

**Fig. 15** Example of legibility enhancement by cross-spectral colorization. (a) Papyrus imaged in visible light; (b) infrared image; (c) the infrared, colorized with the chromatic channels of the visible light image and enhanced with the gamut expansion and stretchlim methods; (d) vividness enhancement of the image (c); (e) negative of image (d); (f) MSRCR-RGB retinex of image (c); (g) negative blue-shift of image (f); (h) and the decorrelation stretch of the blue, red, and infrared bands.

We thereby obtain an image where the details (rendered as lightness) come from outside the visible spectrum, while the context (rendered as chroma and saturation) comes from within it. The use of color allows for visual discrimination between the two types of information, while the lesser spatial acuity of chromatic perception relative to achromatic perception [154] confers saliency to the script with respect to the background. Since we now have a color image, we may apply the methods presented above for further enhancement. Fig. 15 illustrates the improved legibility resulting from the cross-spectral colorization process.

### 4.10 Method integration

The proposed methods lead to better results when used concurrently. A papyri legibility enhancement workflow is presented in Algorithm 1. The vividness method is taken as an example, with gamut expansion, dynamic range stretching, negative polarity, and blue shift also included.

**Algorithm 1.** Papyrus legibility enhancement

▷ input
**read RGB image values**

▷ color gamut expansion (Section 4.2)
**convert color profile using ICC profiles**
· source color space: sRGB IEC 61966-2.1
· target color space: Adobe RGB (1998)
· source render intent: perceptual
· target render intent: perceptual

▷ use a perceptual color space for enhancement (§ 4.3)
**convert color space from RGB to CIELAB**
· source color space: sRGB IEC 61966-2.1
· whitepoint: D65

▷ vividness enhancement (§ 4.5)
**replace lightness by vividness** (Eq. 3)
$L^{*\prime} = (L^{*2} + a^{*2} + b^{*2})^{-1/2}$

▷ dynamic range increase (§ 4.3)
**stretch dynamic range of lightness to bounds** (Eq. 1)
$L^{*\prime} = [L^* - \min(L^*)] \,/\, \{\max[L^* - \min(L^*)]\}$

▷ negative polarity (§ 4.4)
**reverse lightness polarity** (Eq. 2)
$L^{*\prime} = 100 - L^*$

▷ blue shift (§ 4.7)
**change sign of values in chromatic channels** (Eq. 8)
$a^{*\prime} = -a^*$ and $b^{*\prime} = -b^*$

▷ back-conversion to primary colors space
**convert color space from CIELAB to RGB**
· source color space: sRGB IEC 61966-2.1
· whitepoint: D65

▷ manage out-of-gamut values
**clip values to the [0, 1] range**
$I' = \min(0, \max(1, I))$

▷ output
**save image** in TIFF or JPEG format

▷ make color profile explicit
**embed ICC color profile in image file**
· color space: sRGB IEC 61966-2.1

## 5 Experiment

*Objectives* — An experiment was carried out with papyrologists as participants in pursuit of four objectives: (1) Test the hypothesis that the proposed methods outperform existing ones. (2) Use the results to support the decision as to which methods to implement. (3) Explore the data to discover further potentially relevant facts and operationalize the findings. (4) Design the experiment to gain insight into the realistic usage conditions of the legibility enhancement software.

### 5.1 Setup

*Participants* — The experiment participants were two female and five male graduate students, and one male faculty member participating in the Eucor Tri-national Papyrology Webinar 2019–2020 held by the Universities of Basel (Switzerland), Freiburg (Germany), and Strasbourg (France). All were trained in Ancient Greek. Participation in the experiment counted towards their grades. The sample is more representative than suggested by its small size, if it is considered relative to the total number of papyrologists (the International Association of Papyrologists had 484 members in May 2020), and its international heterogeneity.

*Stimuli* — The stimuli consisted of a training and an evaluation set of papyri images. The training set comprised eight images of unpublished papyri from the University of Freiburg Library, whereas the evaluation set comprised the nine publicly available papyri color images from the ICDAR 2019 Competition on Document Image Binarization (DIBCO) [149] (Fig. 2). The quality of the original images in the training set was similar to that of the noisiest evaluation images. The DIBCO images represent typical qualities of papyri documents and reproductions, and the authors participated in their selection process. Each original method was processed with ten of the methods described in the previous section, thus bringing the total of images to 80 and 90 in the training and evaluation set, respectively (the evaluation images are available in the online Supplementary Material).

*Algorithms* — Four novel enhancement algorithms (*vividness, lsv, negvividness*, and *neglsv*) and five comparison algorithms (*stretchlim, histeq, adapthisteq, locallapfilt*, and *retinex*) were used. Given the variability of the lightness dynamic range in the test images, and to facilitate a bias-free comparison of methods, all images had their CIELAB lightness normalized in a preprocessing step, while their gamut was also expanded as described above. All methods were implemented in the Matlab (R2020b) programming environment, except for retinex, which was produced with the ImageJ software package. The choice of comparison methods was based on their status as classical enhancement methods (stretchlim, histeq, adapthisteq), having outstanding edge-preserving performance (locallapfilt), and being a prime choice for achieving color contrast equalization (retinex). — Input images had no embedded color space information, except for P.Oxy.XXII 2309, which was in sRGB. All methods converted the input images to the Adobe RGB (1998) color space with D65 whitepoint and saved them to sRGB to expand the color gamut. Lightness was obtained from CIELAB. — The novel algorithms were those described in Section 4, "Methods". *Negvividness* and *neglsv* apply negative lightness to *vividness* and *lsv*. Due to late-breaking research *hue shift* and *CIELAB retinex* were not included in the experiment. — *Stretchlim* [118] expands the dynamic range of an image's CIELAB lightness to its bounds, [0, 100]. — *Histeq* [79, 118] is a global enhancement method that equalizes the spread of the histogram values of an intensity image over the available bit-depth dynamic range via linearization of its cumulative density function, thereby improving its contrast. — *Adapthisteq* (contrast-limited adaptive histogram equalization, CLAHE) [197, 118] is a local enhancement method similar to histeq, in that it equalizes the image histogram; however, it does this on small regions (here, 8-by-8 pixels). Accordingly, contrast for local structures is increased. — *Locallapfilt* [134, 118] is a state-of-the-art edge-preserving image enhancing method using pyramids of locally-adapted Laplacian filters. Depending on parametrization, it can be used to either smooth or sharpen. Smoothing was used here, with parameters $\sigma = 0.4$, $\alpha = 2$, and $\beta = 1$, derived empirically for their appropriateness to papyri images. — *Retinex* images were produced with the ImageJ implementation (RGB color space, uniform level, scale 240, scale division 3, and dynamic 1.20) [73].

*Procedure* — Each participant transcribed the text of a different image from the training set as part of their webinar examination. Afterward, they were asked to rate the processed images according to how useful they were for text transcription, using the following system: X: "*I used only one image for the transcription, this one*"; A: "*Very useful: it was the primary image used for transcription*"; B: "*I sometimes used this image*"; N: "*I didn't use this image*". (The results were similar to those obtained from the evaluation set.) This experimental step in real use conditions acquainted participants with the opportunities provided by the various processing methods for scholarly text editing.

During the evaluation step, all participants worked on all images of the evaluation set (Fig. 2). The first task involved rating the images for a hypothetical transcription on the same scale as used for the training set. The second task involved the ranking of the same images by utility. Given the effort and time invested in a scholarly transcription of ancient papyri, requiring participants to provide a transcription of all evaluation images it would have been impractical.

Several further questions were asked of the participants regarding how they organized the images and interacted with the computer to compare the versions (e.g., side by side, overlaid, or printed); whether they zoomed into the images; whether they used a desktop, laptop, tablet, or other device; how many displays were used and what sizes they were; what the operating system and image display application were used; and whether they could provide any other information about using the processed images for the transcription task.

*Setting* — The experimental instructions were sent by email to the participants, who downloaded the images to their computers and eventually sent back the completed questionnaires. Given the virtual nature of the papyrology webinar, the participants never physically convened in the same location as the experimenters, thus precluding testing in controlled laboratory conditions. This constraint, however, turned out to be a great opportunity. The homework "field study" enabled *in situ* simulation of the use of then future legibility enhancement software in realistic conditions in a variety of dimensions, and thus led to findings that could not have been obtained in a laboratory setting. In addition to leaving the choice of evaluation environment and time to participants, the setting also maintained their familiarity with the software and hardware employed. Data organization on personal computers, human–computer interaction patterns developed in coexistence with specific operating systems and applications, as well as the display size and number are only some of the factors that would have biased the results if participants had to work in the unfamiliar laboratory settings.

To more accurately simulate real-life conditions, the image filenames were also not anonymized. The identity of these images is indeed known to real users, since they select methods by name in the enhancement software interface. Furthermore, the methods employed in this evaluation produce images that are sufficiently peculiar to make them identifiable at a glance (such as the negative polarity method), which defeats the purpose of anonymization. The differentiated practices that may develop from this knowledge are part of normal user behavior and represent the insights that the evaluation aims to develop. If anonymization was employed, these real behaviors would not have been captured, meaning that the evaluation would have simulated a non-existent condition. An illuminating example of this rationale concerns the special treatment given by the experimental participants to the original image, which was ranked as the most reliable image for the transcription task despite broad acknowledgment of the better script–background contrast in the enhanced images.

Similar considerations to ours have led to remote evaluations in two other legibility enhancement evaluations of historical documents [28, 12]. Large-scale image and video quality assessments with 8100 and 985 participants from multiple countries, rating 1162 and 3000 images respectively, found the results of laboratory and unconstrained online evaluations to be equivalent [63, 144]. Methodologies for crowdsourcing image quality assessment have also been actively investigated [78, 152], while the "ecological validity" of laboratory and field experimentation has been debated for over half a century in fields such as psychology, sociology, and economics [76, 90, 124, 57].

*Frame* — The experiment measured the preferences for legibility enhancement methods in the context of specific documents and available hardware and software, developed in real-life conditions by participants as they gained expertise in using these methods for scholarly text transcription. This statement is necessary to clarify how legibility enhancement was framed by the experimental design. It is not (as would be possible with a different design) transcription errors or duration of task completion that are measured to judge the utility of methods for enhancing legibility, but rather their utility as manifested in their rating and ranking by papyrologists.

## 5.2 Results

Table 1 presents the ratings and rankings of the image processing methods used in the experiment. Providing this data *in extenso* is essential if we are to make visible the statistical features quantified, summarized, and discussed in the next section. For example, while user behavior heterogeneity is apparent from Table 1, it is obfuscated in Table 3 through statistical data aggregation.

**Table 1.** Ranking and rating of legibility enhancement methods.

| | **AW** ♂ 2 displays | **FG** ♀ 3 displays | **FRG** ♂ 1 display | **LG** ♂ 2 displays | **OC** ♂ 2 displays | **OR** ♀ 1 display | **SR** ♂ 2 displays | **SN** ♂ 1 display |
|---|---|---|---|---|---|---|---|---|
| **columbia.apis. p367.f.0.600** | A adapthisteq<br>**neglsv**<br>retinex<br>B **negnorm**<br>**norm**<br>stretchlim<br>**lsv**<br>*original*<br>histeq<br>N locallapfilt | A *original*<br>⎡ adapthisteq<br>histeq<br>locallapfilt<br>**lsv**<br>B **neglsv**<br>**negnorm**<br>**norm**<br>stretchlim<br>⎣ retinex | A adapthisteq<br>B **norm**<br>**neglsv**<br>*original*<br>N **lsv**<br>stretchlim<br>histeq<br>retinex<br>**negnorm**<br>locallapfilt | A **neglsv**<br>B **lsv**<br>**negnorm**<br>**norm**<br>stretchlim<br>*original*<br>locallapfilt<br>N retinex<br>adapthisteq<br>histeq | A **norm**<br>**neglsv**<br>N *original*<br>retinex<br>adapthisteq<br>histeq<br>locallapfilt<br>**lsv**<br>**negnorm**<br>stretchlim | A *original*<br>B retinex<br>**neglsv**<br>**negnorm**<br>N adapthisteq<br>histeq<br>locallapfilt<br>**lsv**<br>**norm**<br>stretchlim | A **negnorm**<br>B locallapfilt<br>**lsv**<br>**neglsv**<br>**norm**<br>stretchlim<br>*original*<br>N adapthisteq<br>histeq<br>retinex | A **lsv**<br>B **neglsv**<br>**negnorm**<br>locallapfilt<br>*original*<br>**norm**<br>stretchlim<br>adapthisteq<br>N retinex<br>histeq |
| **P.Corn. Inv. MSS. A 101. XIII** | A retinex<br>**negnorm**<br>*original*<br>adapthisteq<br>stretchlim<br>B histeq<br>**neglsv**<br>**norm**<br>N locallapfilt<br>**lsv** | A *original*<br>⎡ adapthisteq<br>histeq<br>locallapfilt<br>**lsv**<br>B **neglsv**<br>**negnorm**<br>**norm**<br>stretchlim<br>⎣ retinex | A *original*<br>B **negnorm**<br>**norm**<br>retinex<br>histeq<br>N stretchlim<br>**lsv**<br>**neglsv**<br>adapthisteq<br>locallapfilt | X **negnorm**<br>N *original*<br>retinex<br>adapthisteq<br>histeq<br>locallapfilt<br>**lsv**<br>**neglsv**<br>**norm**<br>stretchlim | X stretchlim<br>N *original*<br>retinex<br>adapthisteq<br>histeq<br>locallapfilt<br>**lsv**<br>**neglsv**<br>**negnorm**<br>**norm** | X *original*<br>N retinex<br>adapthisteq<br>histeq<br>locallapfilt<br>**lsv**<br>**neglsv**<br>**negnorm**<br>**norm**<br>stretchlim | A adapthisteq<br>B *original*<br>histeq<br>**negnorm**<br>locallapfilt<br>**lsv**<br>**norm**<br>retinex<br>stretchlim<br>N **neglsv** | A **negnorm**<br>B *original*<br>**lsv**<br>stretchlim<br>locallapfilt<br>**norm**<br>**neglsv**<br>adapthisteq<br>histeq<br>retinex |
| **P.CtYBR inv. 69** | A retinex<br>**norm**<br>*original*<br>B **negnorm**<br>adapthisteq<br>stretchlim<br>N histeq<br>**lsv**<br>**neglsv**<br>locallapfilt | A *original*<br>⎡ adapthisteq<br>histeq<br>locallapfilt<br>**lsv**<br>B **neglsv**<br>**negnorm**<br>**norm**<br>stretchlim<br>⎣ retinex | A **norm**<br>B *original*<br>histeq<br>retinex<br>**neglsv**<br>N stretchlim<br>adapthisteq<br>**lsv**<br>**negnorm**<br>locallapfilt | A **neglsv**<br>B **negnorm**<br>**lsv**<br>*original*<br>**norm**<br>stretchlim<br>N retinex<br>adapthisteq<br>histeq<br>locallapfilt | A **norm**<br>locallapfilt<br>N *original*<br>retinex<br>adapthisteq<br>histeq<br>**lsv**<br>**neglsv**<br>**negnorm**<br>stretchlim | A *original*<br>B **neglsv**<br>retinex<br>N adapthisteq<br>histeq<br>locallapfilt<br>**lsv**<br>**negnorm**<br>**norm**<br>stretchlim | A **negnorm**<br>B *original*<br>stretchlim<br>**lsv**<br>**norm**<br>locallapfilt<br>**neglsv**<br>adapthisteq<br>N retinex<br>histeq | A **negnorm**<br>B *original*<br>locallapfilt<br>**lsv**<br>**norm**<br>stretchlim<br>N **neglsv**<br>adapthisteq<br>retinex<br>histeq |
| **P.Mich.inv.1318v** | A retinex<br>adapthisteq<br>**neglsv**<br>**norm**<br>*original*<br>B histeq<br>**negnorm**<br>stretchlim<br>N **lsv**<br>locallapfilt | A *original*<br>⎡ adapthisteq<br>histeq<br>locallapfilt<br>**lsv**<br>B **neglsv**<br>**negnorm**<br>**norm**<br>stretchlim<br>⎣ retinex | A **norm**<br>B stretchlim<br>histeq<br>*original*<br>**negnorm**<br>retinex<br>N **lsv**<br>**neglsv**<br>adapthisteq<br>locallapfilt | A **negnorm**<br>B **neglsv**<br>**norm**<br>*original*<br>N retinex<br>**lsv**<br>adapthisteq<br>stretchlim<br>histeq<br>locallapfilt | A **norm**<br>**lsv**<br>N *original*<br>retinex<br>adapthisteq<br>histeq<br>locallapfilt<br>**neglsv**<br>**negnorm**<br>stretchlim | A *original*<br>B retinex<br>N adapthisteq<br>histeq<br>locallapfilt<br>**lsv**<br>**neglsv**<br>**negnorm**<br>**norm**<br>stretchlim | B **neglsv**<br>**lsv**<br>**norm**<br>**negnorm**<br>adapthisteq<br>stretchlim<br>retinex<br>N *original*<br>locallapfilt<br>histeq | A **lsv**<br>B **negnorm**<br>**neglsv**<br>**norm**<br>locallapfilt<br>*original*<br>N retinex<br>stretchlim<br>adapthisteq<br>histeq |
| **P.Mich.inv.2755** | A retinex<br>**norm**<br>**neglsv**<br>B adapthisteq<br>stretchlim<br>*original*<br>N histeq<br>**lsv**<br>**negnorm**<br>locallapfilt | A *original*<br>⎡ adapthisteq<br>histeq<br>locallapfilt<br>**lsv**<br>B **neglsv**<br>**negnorm**<br>**norm**<br>stretchlim<br>⎣ retinex | A **norm**<br>B histeq<br>**neglsv**<br>stretchlim<br>retinex<br>N **lsv**<br>*original*<br>**negnorm**<br>adapthisteq<br>locallapfilt | B retinex<br>**lsv**<br>**norm**<br>stretchlim<br>*original*<br>N **neglsv**<br>**negnorm**<br>adapthisteq<br>locallapfilt<br>histeq | X **norm**<br>N *original*<br>retinex<br>adapthisteq<br>histeq<br>locallapfilt<br>**lsv**<br>**neglsv**<br>**negnorm**<br>stretchlim | X *original*<br>N retinex<br>adapthisteq<br>histeq<br>locallapfilt<br>**lsv**<br>**neglsv**<br>**negnorm**<br>**norm**<br>stretchlim | A **negnorm**<br>B **neglsv**<br>**norm**<br>adapthisteq<br>**lsv**<br>stretchlim<br>*original*<br>retinex<br>N histeq<br>locallapfilt | A **negnorm**<br>B *original*<br>**neglsv**<br>**lsv**<br>**norm**<br>stretchlim<br>adapthisteq<br>N locallapfilt<br>histeq<br>retinex |
| **P.Oxy.XXII 2309** | A retinex<br>**norm**<br>adapthisteq<br>**negnorm**<br>*original*<br>B stretchlim<br>**neglsv**<br>N histeq<br>locallapfilt<br>**lsv** | A *original*<br>⎡ adapthisteq<br>histeq<br>locallapfilt<br>**lsv**<br>B **neglsv**<br>**negnorm**<br>**norm**<br>stretchlim<br>⎣ retinex | A **norm**<br>B stretchlim<br>**negnorm**<br>retinex<br>adapthisteq<br>N histeq<br>locallapfilt<br>**lsv**<br>**neglsv**<br>*original* | A *original*<br>B **negnorm**<br>**norm**<br>stretchlim<br>N **neglsv**<br>retinex<br>adapthisteq<br>locallapfilt<br>histeq<br>**lsv** | A **norm**<br>**neglsv**<br>N *original*<br>retinex<br>adapthisteq<br>histeq<br>locallapfilt<br>**lsv**<br>**negnorm**<br>stretchlim | X *original*<br>N retinex<br>adapthisteq<br>histeq<br>locallapfilt<br>**lsv**<br>**neglsv**<br>**negnorm**<br>**norm**<br>stretchlim | A **negnorm**<br>B **norm**<br>stretchlim<br>*original*<br>locallapfilt<br>**lsv**<br>adapthisteq<br>retinex<br>**neglsv**<br>N histeq | A **negnorm**<br>B *original*<br>stretchlim<br>**norm**<br>retinex<br>**lsv**<br>N **neglsv**<br>locallapfilt<br>adapthisteq<br>histeq |
| **PSI XII 1274 r** | A retinex<br>adapthisteq<br>**neglsv**<br>**norm**<br>*original*<br>B stretchlim<br>**lsv**<br>**negnorm**<br>histeq<br>N locallapfilt | A *original*<br>⎡ adapthisteq<br>histeq<br>locallapfilt<br>**lsv**<br>B **neglsv**<br>**negnorm**<br>**norm**<br>stretchlim<br>⎣ retinex | A **norm**<br>B **neglsv**<br>retinex<br>histeq<br>N stretchlim<br>*original*<br>**lsv**<br>adapthisteq<br>**negnorm**<br>locallapfilt | A **neglsv**<br>B **negnorm**<br>**norm**<br>*original*<br>**lsv**<br>stretchlim<br>N retinex<br>locallapfilt<br>adapthisteq<br>histeq | X **norm**<br>N *original*<br>retinex<br>adapthisteq<br>histeq<br>locallapfilt<br>**lsv**<br>**neglsv**<br>**negnorm**<br>stretchlim | A *original*<br>B retinex<br>**neglsv**<br>**negnorm**<br>N adapthisteq<br>histeq<br>locallapfilt<br>**lsv**<br>**norm**<br>stretchlim | A **lsv**<br>B stretchlim<br>locallapfilt<br>**negnorm**<br>**norm**<br>adapthisteq<br>**neglsv**<br>retinex<br>N *original*<br>histeq | A **lsv**<br>B locallapfilt<br>*original*<br>**norm**<br>stretchlim<br>**negnorm**<br>**neglsv**<br>N retinex<br>adapthisteq<br>histeq |
| **PSI XIII 1298 (15a) r 1** | A retinex<br>**neglsv**<br>*original*<br>B **lsv**<br>**negnorm**<br>adapthisteq<br>**norm**<br>N histeq<br>stretchlim<br>locallapfilt | A *original*<br>⎡ adapthisteq<br>histeq<br>locallapfilt<br>**lsv**<br>B **neglsv**<br>**negnorm**<br>**norm**<br>stretchlim<br>⎣ retinex | A **lsv**<br>B *original*<br>retinex<br>histeq<br>**norm**<br>**neglsv**<br>N **negnorm**<br>stretchlim<br>adapthisteq<br>locallapfilt | B **neglsv**<br>adapthisteq<br>**lsv**<br>stretchlim<br>**negnorm**<br>**norm**<br>N *original*<br>retinex<br>histeq<br>locallapfilt | X **norm**<br>N *original*<br>retinex<br>adapthisteq<br>histeq<br>locallapfilt<br>**lsv**<br>**neglsv**<br>**negnorm**<br>stretchlim | X *original*<br>N retinex<br>adapthisteq<br>histeq<br>locallapfilt<br>**lsv**<br>**neglsv**<br>**negnorm**<br>**norm**<br>stretchlim | A **neglsv**<br>B **lsv**<br>**norm**<br>**negnorm**<br>stretchlim<br>adapthisteq<br>retinex<br>N *original*<br>histeq<br>locallapfilt | A **lsv**<br>B *original*<br>**negnorm**<br>**neglsv**<br>**norm**<br>stretchlim<br>N retinex<br>locallapfilt<br>adapthisteq<br>histeq |
| **PSI XIV 1376 r** | A **norm**<br>**neglsv**<br>**lsv**<br>adapthisteq<br>B **negnorm**<br>stretchlim<br>retinex<br>*original*<br>N histeq<br>locallapfilt | A *original*<br>⎡ adapthisteq<br>histeq<br>locallapfilt<br>**lsv**<br>B **neglsv**<br>**negnorm**<br>**norm**<br>stretchlim<br>⎣ retinex | A **lsv**<br>**neglsv**<br>**norm**<br>histeq<br>retinex<br>N adapthisteq<br>**negnorm**<br>stretchlim<br>*original*<br>locallapfilt | B **neglsv**<br>**negnorm**<br>**lsv**<br>*original*<br>N adapthisteq<br>histeq<br>locallapfilt<br>**norm**<br>retinex<br>stretchlim | A **norm**<br>adapthisteq<br>**neglsv**<br>N *original*<br>retinex<br>histeq<br>locallapfilt<br>**lsv**<br>**negnorm**<br>stretchlim | A *original*<br>B retinex<br>**neglsv**<br>**negnorm**<br>locallapfilt<br>N adapthisteq<br>histeq<br>**lsv**<br>**norm**<br>stretchlim | A **lsv**<br>B **norm**<br>**negnorm**<br>stretchlim<br>**neglsv**<br>locallapfilt<br>*original*<br>adapthisteq<br>retinex<br>N histeq | A **negnorm**<br>B **neglsv**<br>*original*<br>**lsv**<br>**norm**<br>stretchlim<br>adapthisteq<br>locallapfilt<br>N histeq<br>retinex |

*Note:* "X" denotes methods used to the exclusion of all others, "A" those that were of primary use, "B" those that were of secondary use, and "N" (on gray background) those that were of no use. Brackets indicate tied rankings.

**Table 2.** Utility of enhancement methods.

| *Methods* | *Utility category counts* | | | |
|---|---|---|---|---|
| | *Exclusive X* | *Primary A* | *Secondary B* | *No use N* |
| original | 4 | 22 | **28** | 18 |
| | | *Proposed methods* | | |
| lsv | 0 | 10 | 30 | **32** |
| vividness | 3 | 16 | **41** | 12 |
| neglsv | 0 | 13 | **38** | 21 |
| negvividness | 1 | 12 | **36** | 23 |
| *Total proposed* | *4* | *51* | ***145*** | *88* |
| | | *Comparison methods* | | |
| stretchlim | 1 | 1 | **42** | 28 |
| histeq | 0 | 0 | 22 | **50** |
| adapthisteq | 0 | 9 | 25 | **38** |
| locallapfilt | 0 | 1 | 23 | **48** |
| retinex | 0 | 8 | **33** | 31 |
| *Total comparison* | *1* | *19* | *145* | ***195*** |

*Utility category percentages, column-wise*

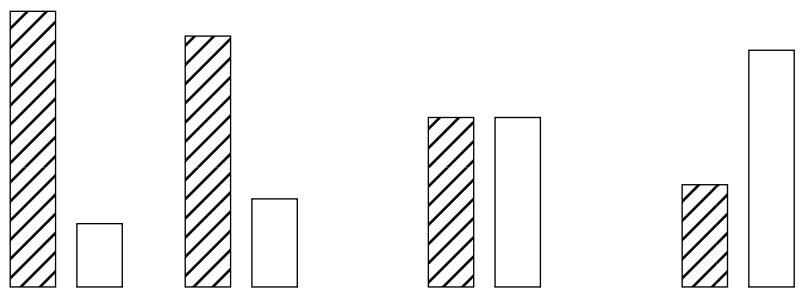

| | X | A | B | N |
|---|---|---|---|---|
| Proposed methods | **80** | **73** | **50** | 31 |
| Comparisons | 20 | 27 | **50** | **69** |

**Table 3.** Overall ranking of enhancement methods by utility ratings.

**A. Ranking by Centroids method**

| *Rank* | *Method* | *Rating* | *Spread* (1) |
|---|---|---|---|
| 1 | *original* | 31.1 | |
| 2 | **vividness** | 29.1 | |
| 3 | **neglsv** | 11.1 | |
| 4 | **negvividness** | 10.1 | |
| 5 | **lsv** | −2.9 | |
| 6 | retinex | −3.9 | |
| 7 | stretchlim | −5.9 | |
| 8 | adapthisteq | −9.9 | |
| 9 | locallapfilt | −27.9 | |
| 10 | histeq | −30.9 | |

(1) Graphical representation of the ratings distribution.

**B. Ranking by Majority Judgement method**

| *Rank* | *Method* | *Rating* | *Spread* | *Category* |
|---|---|---|---|---|
| 1 | *original* | 2.37 | | Secondary use + |
| 2 | **vividness** | 2.26 | | Secondary use + |
| 3 | **neglsv** | 1.71 | | Secondary use − |
| 4 | **negvividness** | 1.68 | | Secondary use − |
| 5 | stretchlim | 1.61 | | Secondary use − |
| 6 | retinex | 1.57 | | Secondary use − |
| 7 | **lsv** | 1.56 | | Secondary use − |
| 8 | adapthisteq | 1.48 | | No use |
| 9 | locallapfilt | 1.33 | | No use |
| 10 | histeq | 1.31 | | No use |

**C. Aggregation of rankings by ROD method**

| *Rank* | *Method* | *Rating* | *Spread* | *Optimization* |
|---|---|---|---|---|
| 1 | *original* | ∞ | | *original* |
| 2 | **vividness** | 73.908 | | **vividness** |
| 3 | **neglsv** | 2.091 | | **neglsv** |
| 4 | **negvividness** | 1.779 | | **negvividness** |
| 5 | stretchlim | 0.483 | | stretchlim |
| 6 | retinex | 0.475 | | retinex |
| 7 | **lsv** | 0.460 | | **lsv** |
| 8 | adapthisteq | 0.204 | | adapthisteq |
| 9 | locallapfilt | 0.008 | | locallapfilt |
| 10 | histeq | 0.000 | | histeq |

## 5.3 Analysis

### 5.3.1 How good are the proposed methods?

A visual inspection of Table 1 suggests that, as a group, the proposed methods ranked overall higher than the comparison methods. This impression is substantiated by the distribution of ratings of the two method sets according to their utility: the proposed methods are prevalently of primary use, whereas the comparison methods are mostly of no use (Table 2).

### 5.3.2 How do the methods differ?

The methods' effects on legibility were examined by combining the participants' feedback with a visual inspection of the enhanced images (Fig. 3) and understanding of the algorithms. The *original* image was typically low-contrast, and some portions of the script were faded or missing. *Stretchlim* (i.e., normalization of lightness) was often insufficient. *Histeq* created high local contrast variation, because it is based on global contrast improvement. *Adapthisteq* strongly enhanced the papyrus structure, which interferes with the script. *Locallapfilt* had very low contrast, and its smoothness was too strong; parametrization for individual images becomes thus necessary, although this is difficult to do automatically and inconvenient when done manually. *Retinex* performed best out of all methods at revealing the script, creating very sharp edges. However, reading was also made difficult because of interference from the papyrus structure, and there was also a strong vignetting effect that sometimes completely obscured parts of the image. *LSV* had excellent smoothing properties, but often damaged script structures. *Vividness* gives the most balanced results, although the contrast could be improved.

### 5.3.3 Which methods should be implemented in software?

To determine the relative utility of the individual methods, their ratings were aggregated across all participants and documents in an overall ranking (Table 3). This ranking was obtained with two different ranking methods, after which the outputs were aggregated and optimized. The top-1 enhancement method is vividness, followed by the negative method. The least useful were locallapfilt and histeq.

*Methodology* — While ranking has been extensively investigated in statistics [11], we found methods with more useful properties for the task at hand in the domains of operations research, marketing, sports, and voting theory. These methods are non-parametric, and thus facilitate the handling of data with unknown distributions. They take categorical and ordinal data as input, which corresponds to the experimental ratings and rankings. They also differ in terms of their perspectives, with the mean-based methods being competitive and rewarding the individual performance of the enhancement methods, while the median-based method is consensus-based and satisfies the majority of users. These methods are briefly described below; the code was published in [15].

The *Centroids* ranking is based on matrix analysis [108: 127–129]. The $n$-by-1 vector $\boldsymbol{r}$ of the aggregated rating scores of $n$ items with individual scores $s$ (or ranks $k$, for which $s = n + 1 - k$) is defined as follows:

$$\boldsymbol{r} = \text{round}\,[\,(\boldsymbol{S} - \boldsymbol{S}^{\mathbf{T}})\,\boldsymbol{e}\,/\,n\,]\,, \tag{9}$$

where $\boldsymbol{S}$ is an $n$-by-$n$ matrix, with $\boldsymbol{S}_{ij} = \sum(s_i)$ for $i \neq j$ and $\boldsymbol{S}_{ij} = 0$ for $i = j$ (diagonal of zeros), $\boldsymbol{S}^{\mathbf{T}}$ is the transpose of $\boldsymbol{S}$, and $\boldsymbol{e}$ is an $n$-by-1 vector of ones. (Note the row-wise averaging, from which the method derives its name.) Using Table 3.A as an example, the items correspond to the enhancement methods; their corresponding ranks are obtained from Table 1, after which application of the above equations results in the values listed in the rating column of Table 3.A. Their ranking is obtained by sorting the rating vector $\boldsymbol{r}$ in descending order of magnitude.

The *Ratio of Offense and Defense* (ROD) [108: 176–197] takes as input the same score matrix $\boldsymbol{S}$ used by the Centroids method, from which it derives the ratings vector $\boldsymbol{r}$ through the following equation:

$$\boldsymbol{r} = \boldsymbol{o}\,/\,\boldsymbol{d}\,, \tag{10}$$

where the methods' "offensive" strength is $\boldsymbol{o} = \boldsymbol{R}\boldsymbol{e}$, while their "defensive" strength is $\boldsymbol{d} = \boldsymbol{R}^{\mathbf{T}}\boldsymbol{e}$. (The terminology comes from the world of sports, where this ranking method originates.)

In *Majority Judgment* each elector grades each candidate (instead of choosing a single one), and rates and ranks are obtained from mean scores, resulting in the elections of those candidates that are the most acceptable to the largest number of electors, i.e. consensual [18]. The algorithm begins by iteratively removing the median values $\{m_1, m_2, \dots, m_p\}$ from the ratings $\{r_1, r_2, \dots, r_p\}$ of each candidate $\{c_1, c_2, \dots, c_n\}$, and concatenating them lexicographically. For example, the ratings vector $\{7, 9, 9, 11, 11\}$ is reordered by medians into $\{9, 9, 11, 7, 11\}$, and concatenated to generate the numerical value 9.09110711. The resulting scalars $\{v_1, v_2, \dots, v_n\}$, called "majority values", are then sorted in descending order of magnitude to yield the candidate rankings $\{k_1, k_2, \dots, k_p\}$, where $v(k_1) \succ v(k_2) \succ \dots \succ v(k_p)$. This procedure was applied to the enhancement method utility ratings $\{$X, A, B, N$\}$ of Table 1 in order to generate the rankings and ratings presented in Table 3.B. While ranking is an ordinal representation and rating is numeric, verbal qualifiers have their own benefits. For example, the utility of the ranked enhancement methods is easily recognizable from the category column of Table 3.B, which uses the spelled-out names of the four utility classes {Exclusive, Primary, Secondary, No use} corresponding to the ratings. The plus and minus signs following the class names indicate whether a method belongs to the upper or lower range of a utility class.

Empirical evidence suggests that aggregated opinions may outperform individual ones, an approach formalized by *ensemble methods* [103]. When comparing the results of the Centroids and Majority Judgment methods, differences can be seen in ranking and rating, reflecting the methods' differing outlooks. To obtain a balanced result, the two rating vectors are aggregated, first by deriving the matrices $\boldsymbol{R}$ of pairwise positive differences from each of the $n$ (in our case $n = 2$) rating vectors $\boldsymbol{r}$:

$$\boldsymbol{R}_{p(ij)} = [\boldsymbol{r}_{p(i)} - \boldsymbol{r}_{p(j)}] \text{ for } [\cdot] > 0, \text{ and } \boldsymbol{R}_{p(ij)} = 0 \text{ for } [\cdot] \leq 0\,, \tag{11}$$

where $p = \{1, 2, \dots, n\}$. The matrices are then normalized and averaged:

$$\bar{\boldsymbol{R}} = \sum[\,\boldsymbol{R}_p\,/\,\sum(\boldsymbol{R}_p\boldsymbol{e})\,]\,/\,n\,, \tag{12}$$

where $\boldsymbol{e}$ is an $n$-by-1 vector of ones. Finally, the aggregated scores vector $\boldsymbol{r}$ is obtained using the ROD method: $\boldsymbol{r} = \bar{\boldsymbol{R}}\boldsymbol{e}\,/\,\bar{\boldsymbol{R}}^{\mathbf{T}}\boldsymbol{e}$. This procedure produced the ratings shown in Table 3.C. The approach is akin to graph analysis, as commonly used in e. g. website ranking methods, since negative differentials between pairwise scores are set to zero, meaning that the matrix $\bar{\boldsymbol{R}}$ reflects the dominance of each enhancement method.

A further refinement, named *Kemenization*, allows for the aggregated ranking to be optimized [88: 9–23; 108: 175–176]. It consists in switching item $k_i$ with item $k_{i+1}$ in the aggregated vector $\boldsymbol{r}$ if item $k_{i+1}$ occurs more frequently among the ranks $i$ of the input vectors $\boldsymbol{r}_p$. From the optimization column of Table 3.C, we can observe that the procedure results in no modification of the aggregated ranking.

### 5.3.4 Are the evaluated methods sufficient?

The fact that multispectral images typically provide superior legibility indicates that additional enhancement methods are desirable and sometimes indispensable (e.g., when reading carbonized papyri [125, 171]). Nevertheless, participants attempting further image manipulation, such as in Photoshop, could not usefully improve legibility: "The enhanced pictures were most often so good that this didn't help as much as I hoped" (SN). Also, one method, i.e. vividness, was substantially better than the others (Table 3). Finally, over-reliance on the original image was apparent: "I like to work on the original version, and I find it's not always necessary to look at the others" (OR); "I mainly worked with the 'base' image (vividness or original). This represented 95% of my work. . . . The other . . . I briefly tried to use them, but quickly abandoned." (OC). This final comment was reflected in the numerous cases in which methods were deemed not useful (42% of $n = 720$ in Table 2; swaths of gray in Table 1). Provided that the DIBCO dataset is representative of papyri, the conclusion is that the amount and type of novel and existing methods made available are satisfactory on average.

### 5.3.5 Enhancement methods are complementary and their utility is context-dependent

Table 1 reveals a substantive variability of method ranking and ratings within participants, between participants, and between documents. Retinex, for example, is the first choice for participant AW, but the last for OC (Table 4). The statistical analysis of agreement and spread quantitatively confirms the striking variability (Table 5 and Fig. 16). Various distributions are present: unimodal, bimodal, and uniform (Fig.

**Table 4.** Intra-class ranking of enhancement methods by utility ratings (optimized aggregation by ROD method; ties in brackets).

**A. Within–user ranking** (column-wise aggregation of ratings from Table 1; ties in brackets)

| AW | FG | FRG | LG | OC | OR | SR | SN |
|---|---|---|---|---|---|---|---|
| retinex | *original* | **vividness** | **negvividness** | **vividness** | *original* | **negvividness** | **negvividness** |
| *original* | **lsv** | retinex | **neglsv** | **neglsv** | retinex | **lsv** | **lsv** |
| **vividness** | **vividness** | histeq | *original* | stretchlim | **neglsv** | **neglsv** | *original* |
| adapthisteq | **neglsv** | **neglsv** | **vividness** | **lsv** | **negvividness** | adapthisteq | **vividness** |
| **neglsv** | **negvividness** | *original* | **lsv** | adapthisteq | locallapfilt | **vividness** | stretchlim |
| **negvividness** | stretchlim | **lsv** | stretchlim | locallapfilt | **lsv** | stretchlim | **neglsv** |
| stretchlim | histeq | **negvividness** | adapthisteq | *original* | **vividness** | retinex | locallapfilt |
| **lsv** | adapthisteq | stretchlim | locallapfilt | **negvividness** | stretchlim | *original* | adapthisteq |
| histeq | locallapfilt | adapthisteq | retinex | histeq | histeq | locallapfilt | retinex |
| locallapfilt | retinex | locallapfilt | histeq | retinex | adapthisteq | histeq | histeq |

**B. Within–document ranking** (row-wise aggregation of ratings from Table 1; ties in brackets)

| columbia.apis. p367.f.0.600 | P.Corn.Inv. MSS.A 101.XIII | P.CtYBR inv.69 | P.Mich. inv.1318v | P.Mich. inv.2755 | P.Oxy.XXII 2309 | PSIXII 1274r | PSIXIII 1298 (15a) r1 | PSIXIV 1376r |
|---|---|---|---|---|---|---|---|---|
| **neglsv** | *original* | *original* | **vividness** | **vividness** | *original* | **vividness** | *original* | **neglsv** |
| *original* | **negvividness** | **vividness** | *original* | *original* | **vividness** | **neglsv** | **lsv** | **lsv** |
| **vividness** | stretchlim | **negvividness** | **negvividness** | stretchlim | **negvividness** | *original* | **neglsv** | *original* |
| **negvividness** | retinex | **neglsv** | **neglsv** | **neglsv** | stretchlim | **negvividness** | **vividness** | **vividness** |
| **lsv** | adapthisteq | stretchlim | retinex | retinex | retinex | **lsv** | **negvividness** | **negvividness** |
| adapthisteq | **vividness** | locallapfilt | **lsv** | **negvividness** | **neglsv** | retinex | retinex | adapthisteq |
| stretchlim | histeq | retinex | stretchlim | **lsv** | adapthisteq | stretchlim | stretchlim | retinex |
| locallapfilt | **lsv** | **lsv** | adapthisteq | adapthisteq | **lsv** | adapthisteq | adapthisteq | stretchlim |
| retinex | **neglsv** | adapthisteq | histeq | histeq | locallapfilt | histeq | histeq | locallapfilt |
| histeq | locallapfilt | histeq | locallapfilt | locallapfilt | histeq | locallapfilt | locallapfilt | histeq |

16A). Clearly, *no single method is adequate for every participant–document pairing*. The implication here is that an optimal text transcription can be obtained only by optimizing individual user/document/method configurations, and not by the method that performs best on average; in other words; the choice of enhancement method is best left to the user.

A second insight from Table 1 is provided by participant FG, who stated that all methods were equally useful. Other comments concur: "the images . . . definitely complemented each other" [FG]. Therefore, not only is no single method is the best, but *there is an explicit need for the concurrent use of multiple methods to optimize text transcription*.

The above precept is even more pertinent as image quality varies locally (e.g., AW ranked the method utility differently for the recto and verso of the same papyrus), and local information is critical to legibility. These findings further support the conclusions drawn from the overall method ranking. Rather than implementing a universal solution for legibility enhancement, *multiple methods should be offered to users*.

These results are unsurprising. The DStretch software, for example, offers more than fifty parameterizable methods. A recent comparison of the performance of eight optical, X-ray, and terahertz-based imaging approaches for recovering text within Egyptian mummy cartonnage concluded that it is only by "carefully selecting, optimizing and combining" them that success may be achieved [64: 1]. Similarly, the organizers of the Document Image Binarization competition explicitly state that "no binarization algorithm is good for all kinds of document images" [44, 113]. In industries dealing with color imaging and reproduction, the impossibility of a universal algorithm for color conversion (gamut mapping) is also acknowledged. Various strategies are therefore utilized to select appropriate methods, on the basis of (for example) the user's subjective intention (maintaining saturation or overall color appearance [127: 3–5, 107–109, 194]), the linguistic dimension (preserving the names given to colors [127: 218–219]), or the semantic content (e.g., skin color [127: 216]).

**Table 5.** Agreement of ratings of enhancement methods, measured with Kendall's coefficient of concordance, *W*.

**A. Inter-class agreement**

| | |
|---|---|
| Between users | 0.376 |
| Between documents | 0.698 |

**B. Within-user agreement**

| | | |
|---|---|---|
| 1. | FG | 1 |
| 2. | OR | 0.727 |
| 3. | AW | 0.616 |
| 4. | SN | 0.607 |
| 5. | OC | 0.500 |
| 6. | LG | 0.474 |
| 7. | SR | 0.453 |
| 8. | FRG | 0.386 |

**C. Within-document agreement**

| | | |
|---|---|---|
| 1. | P.Oxy.XXII 2309 | 0.376 |
| 2. | PSI XIII 1298 (15a) r 1 | 0.286 |
| 3. | P.Corn. Inv. MSS. A 101. XIII | 0.283 |
| 4. | P.Mich.inv.2755 | 0.280 |
| 5. | columbia.apis.p367.f.0.600 | 0.280 |
| 6. | PSI XIV 1376 r | 0.246 |
| 7. | P.CtYBR inv. 69 | 0.243 |
| 8. | P.Mich.inv.1318v | 0.236 |
| 9. | PSI XII 1274 r | 0.220 |

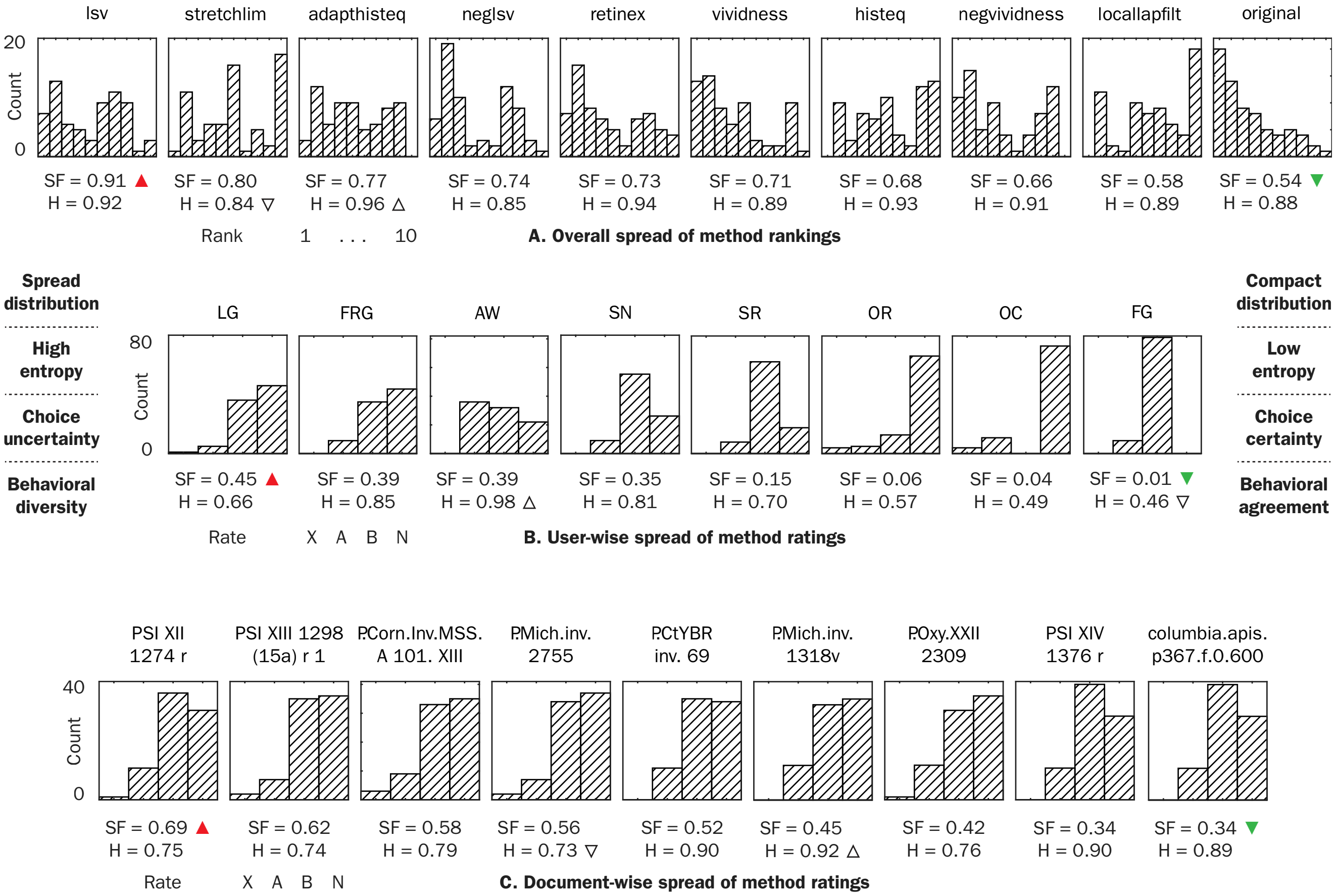

**Fig. 16** Spread of enhancement method rankings and ratings, measured with spatial flatness $SF$ and Shannon entropy $H$.

*Comparison* — Our findings appear to differ from those obtained by two recent legibility enhancement evaluations of historical documents, which explicitly indicate the *homogeneity* of agreement between participants. Brenner and Sablatnig [28] produced an annotated dataset of 250 multispectral images from 20 paper and parchment manuscripts, with selected regions of interest graded for their legibility by 20 scholars; the settings were both laboratory and remote, and the interclass correlation coefficient (ICC) [96] was used for statistical analysis. Arsene *et al.* [12] had seven participants evaluate 15 pseudocolor multispectral fusion methods for their utility in enhancing the legibility of one parchment palimpsest page; a laboratory setting and ICC analysis were used.

There are evident similarities and differences in terms of intent and experimental setup between the above-mentioned works and the present research. We do however consider that the main reason for the behavioral *heterogeneity* observed in our study stems principally from three factors. First, all three evaluations contend with small sample sizes and unconstrained settings, which may amplify variance (while large datasets may hide diversity). Second, we adopted the following methodological "mantras": do not reject outliers, but question their significance (cf. the case of the original image); examine the raw data (Table 1) before aggregating it (Tables 2 to 5); do not assume normality for the data distribution (while ICC does). Third, the reported behavioral differences may reflect the evaluators' differing outlooks on which experimental aspects to operationalize. The focus is on one hand on personalization and contextualization, and on the other hand on generalization. The choice depends on the situation and affordability: enhancement for critical vision benefits from local optimization, while global optimization is more efficient for casual vision.

*Methodology* — *Kendall's coefficient of concordance W* is non-parametric and distribution-independent statistic of agreement between multiple rankings, appropriate for the multimodal nature the analyzed data [89: 94–116]. The results in Table 4 were obtained by adjusting for ties and by correcting for small sample sizes, followed by a $\chi^2$ test of significance for large number of ties. The hypothesis $H_0$ that the between-user and between-document agreements are due to chance is rejected in favor of the alternative, with $p = 0.001$ and $p = 6.124\text{e}{-9}$, respectively, at an $\alpha = 0.1$ significance level. The coefficient of concordance, $W$, for $m$ ranked items and $n$ rankings is defined as

$$W = 12\,S \,/\, [\,m^2(n^3 - n) + 2 - m \textstyle\sum (t^3 - t)\,]\,, \qquad (13)$$

where $S$ is the sum of squares of rank deviations from the mean, and $t$ is the length of runs of tied ranks. The formula for the $\chi^2$ value is:

$$\chi^2 = 12\,S \,/\, [\,m\,(n^2 + n) + 2 - \textstyle\sum (t^3 - t)/(n-1)\,]\,. \qquad (14)$$

*Spatial flatness SF* is the complement of the spectral flatness [196; 31: 104–105], which is itself the ratio of the geometric and arithmetic mean of the magnitude of the Fourier transform $\mathcal{F}$ of a signal $x$ of length $n$:

$$SF = 1 - n\,(\textstyle\prod |\,\mathcal{F}(x)\,|\,)^{1/n} \,/\, \textstyle\sum |\,\mathcal{F}(x)\,|\,. \qquad (15)$$

Spatial flatness varies between 0 for an impulse and 1 for a uniform distribution. It is used in signal processing and is implemented as one of the descriptors of the MPEG-7 multimedia content description standard, with applications including classification into noise and tonal sounds [82: 29–32]. We applied this method to characterize histograms. It is a powerful, non-parametric statistic that informs on (1) the *coverage* of a range of possible values by actual values in a vector (e.g., to how many different ranks experimental participants assigned a method), (2) the *variance* of the distribution's amplitude (e.g., whether one rank is preferred over the others), and (3) the *homogeneity* of distribution (e.g., whether participants exhibit consistent or eclectic behavior). Such a rich description is made possible by the properties of the frequency domain, where regularities in signals are much easier to investigate than in the spatial domain. For the sake of comparison, Fig. 16 gives the Shannon entropy, $H$, of the ranking histograms. The "peakiest" histogram in Fig. 16A has the lowest $SF$ value, but the lowest $H$ value is that of a histogram wherein the values are fairly evenly distributed.

### 5.3.6 From facts to action: an operationalization pattern

The findings on rating variability derived from Table 1 were operationalized via the following *design pattern*: measurement ▹ interpretation ▹ implications ▹ implementation.

The variability of ratings within columns represents the *within-participant variability* and is indicative of two behavioral attitudes: *eclecticism* in the use of enhancement methods and *adaptability* to document particularities. The prevalence of high eclecticism may translate to a decision to accommodate this behavior by simplifying the access of users to the various enhancement methods, e.g., by tiling the interface windows. Conversely, low eclecticism may lead to methods organized in depth,via several menu levels. High adaptability may be considered an incentive to invest in developing image quality metrics, in order to predict the enhancements best suited to a given document, and to rank them dynamically in the software interface. The measurements of agreement in Table 5B show a wide distribution range between eclectic and conservative participants (FG has a uniform stimulus responses), and adaptable and rigid behaviors. The conclusion is that both *depth* and *flat interface designs* are needed, as well as that developing image quality models is desirable.

The principal factors in the level of *within-document rating agreement* in Table 5C are the document and image quality, script legibility, and user experience. This information could aid in evaluating the significance of enhancement method performance. In relating document rankings to their images in Fig. 2, no correlations with either factor are found. For example, moderate agreement characterizes both low and high document or image quality (PSI XIV 1376 r vs columbia.apis.p367.f.0.600), whereas low agreement is obtained for all script cursivity levels (PSI XII 1274 r vs P.Mich.inv.1318v).

Comparison of *between-user* and *between-document rating agreement* may reveal the dominant factor driving the observed behavioral heterogeneity in the rating of the enhancement methods, owning either to user variability or to document quality variability. The results would have bearing on whether to invest more in *human–computer interaction* issues or *enhancement algorithms*. The values in Table 5A show a greater agreement in between-document ratings than between-user ratings, and thus support the focus on interaction facilities, more than on enhancement algorithms. This empirical finding is a surprising conclusion to an article devoted to algorithms, and thus merits further exploration.

### 5.3.7 Approaching legibility enhancement as a system

The preceding findings detailed the *high heterogeneity of user behavior* and the *large number of factors* affecting which legibility enhancement methods users choose to utilize. The following model of the legibility enhancement process aims to help better understand these phenomena.

The SYSTEM takes *bitmap text* as input and outputs *electronic text*. During this process the *entropy* of the text is lowered (i.e., a cryptic text has been read). The system is composed of *interacting elements* and their *properties*. The principal aspects are: (1) the TASK of critical reading, (2) the IMAGES with various degrees of content *affordance*, (3) the TOOLS creating an ergonomic environment (such as image viewing software and displays), (4) the METHODS of legibility enhancement that are variously useful for the users, and (5) the USERS themselves, along their level of expertise in interacting with the system components. The BEHAVIOR of users emerges from the processes within the system, and is characterized by an idiosyncratic *eclecticism* in how they use the methods and the tools, as well as by contextual *adaptability* to the specificities of documents and their digital reproductions. LEGIBILITY is a property not only of images alone, but also of user *expertise* and tool *ergonomy*. Likewise, the *utility* of a legibility enhancement method is a property of the interaction among all elements of the system; therefore, its measurement is relative to the overall context.

### 5.3.8 Contextual system optimization is preferable

Image interpretation in scholarly transcription, and even more so in medical, forensic, and military contexts, can have significant repercussions on people's lives (e.g., the role of the Rosetta Stone transcription in the Egyptian tourism industry). Optimal performance on a case-by-case basis, rather than considered statistically, is thus critical. This can be achieved only if the "variables" of the legibility system (i.e., methods and tools) are attuned to the "constants" (i.e., user, document, and reproduction). In short, choosing the *globally* optimal method for all combinations of the system states will yield suboptimal performance in most cases; only when several analysts express themselves on the same document can their opinions be fused into a unique solution that might surpass individual solutions. To use an analogy, biometric security systems are optimized for individual users, whereas physical keys are universally functional for any user; evidently, the average of biometric signals from multiple users will result in very poor authentication performance.

The evidence in respect to legibility enhancement thus favors the *personalization* of methods and tools to specific users or demographics, the *contextualization* to the task, and the *adaptation* to individual documents and reproductions.

One simple implementation of a legibility enhancement system optimized to local conditions would be to offer a spectrum of methods and then *leave the choice to the users*. More sophisticated techniques are the ranking of methods according to their suitability for the document in question, and adaptation to the behavior of a given user. A fully-fledged context-aware critical visual enhancement system (e.g., autonomous robots) would have at its disposal a multitude of enhancement methods to choose from and combine in the function of the observed distortion types, scene content, environmental conditions, technological resource application domains, and user demographics and individualities.

*Related work* — Digital photo editing is a research area where adaptation to individual and demographic preferences has been proposed as a way to develop image enhancement processes beyond the stage of universal filters [85]. Typical features of interest are the lightness transfer function (S-curve), temperature, and tint, while more complex edits may be modeled using deep learning [92].Taking user intent and image content into account are some of the challenges of this application type.

Demographics and individual adaptation have long been of interest for medical applications of analog and computational image enhancement. For example, macular degeneration is a common ophthalmologic pathology manifested by a blurred spot in the visual field. The detrimental impact of this condition, including the handicap it presents for reading, may be mitigated by bandpass filtering based on psychophysical models of low vision [137], a technique that could be implemented in an MPEG image decoder for real-time applications [93]. The progressive yellowing of the eye's crystalline lens with aging is another example of the demographic dimension of image enhancement. It leads to a darker field of vision, a condition for which it has been proposed to use lightness enhancement by means of non-linear tone mapping and high-pass filtering (sharpening) [126]. The several types and degrees of colorblindness (affecting about eight percent of males and one percent of females of European ancestry [135: 222–223; 120]) have spurred the development of a further class of demographic-specific image enhancement techniques intended to simulate color vision deficiency for the normal-sighted [185, 116], as well as enhancing the color perception of the colorblind [151, 53, 195].

One challenge associated with medical-related image enhancement concerns the wide variety and individuality of pathologies that need to be accounted for in software [59]. Furthermore, the suboptimal ergonomics of the hardware in which this software embedded is often a reason why they fail to be used [136]. Consequently, a number of creative design solutions have been proposed, such as the integrated visual and audio representations of images [67]. An additional persistent problem remains the difficulty of evaluating the effectiveness of the experimental enhancement methods outside the laboratory, as part of the daily life of the visually impaired [138]. This aspect has considerable significance, given the far greater disparity for pathological applications than for non-pathological ones between laboratory and real-life environments (in a similar manner as we suggest to be the case for paleographic transcription).

#### 5.3.9 The primacy of the "original" image

Overall, papyrologists ranked original images as the most useful for transcription. Participant SN is unambiguous: "I'm convinced that the original picture is the most important reference point in discussing a text." Why should this be so, and what might it mean for software development?

In many domains — papyrological, medical, or journalistic — the more removed the information from its source, the less trustworthy it is considered (cf. witness vs hearsay evidence, primary vs secondary documentary sources). The dogma of the "original" is a socio-professional specificity inculcated early on in papyrological education. As sensible this is in the context of physical documents examined visually, digital reproductions are, however, inherently "manipulated" images, and an image at the exit point of an imaging system is no more faithful than, say, an image calibrated during post-production. Hence, suitable documentation should clarify alterations by the enhancement methods, and thus preserve the confidence of users in that they understand the source of features observed in images.

The "low toner effect" might have contributed. Psychologists found that, counterintuitively, a job application letter of poor print quality was more trustworthy than a pristine print. The proposed explanation was that low legibility demands more cognitive effort, which is interpreted as a reflection of the high value of the letter content [132: 149–151]. The effect was reproduced with mathematical exercises more often correctly solved when in an uncommon faded typeface [10].

It would however be misleading to rely solely on the overall ranking of enhancement methods (Table 3) and thereby conclude that the original is the top-performing image under all conditions. For example, between-user variability is substantial: whereas OR consistently rates the original as being of "exclusive use" or "primary use", OC rejects it as "not useful" (Table 2). The software should thus be able to adapt to user preferences, as well as document specificities.

#### 5.3.10 The unexpected usefulness of negative images

The trivial negative method is noteworthy not only for its excellent performance as a top-2 enhancement method after vividness (Table 2), but also for its *discriminatory qualities*. It helps to ascertain whether a specific pixel cluster might be ink ("I used the negative image to understand whether I was imagining an ink trace or it was real." [OC]), and to distinguish between ink and holes in the papyrus, a recurrent issue in reading papyri reproductions. One further point of practical relevance regarding this method is that it is applicable to *monochrome images*. Some papyri survive or are remotely accessible only in this format; other (such as the charred papyri) have little chromatic information.

The finding that negative polarity is considered by users to improve legibility is of interest because it contradicts the prevailing *psychophysical evidence*, which consistently associates better legibility with positive contrast (i.e., dark script on light background) [32, 145, 109, 99, 26: 206–207]. The possible cause of the disparity is that these studies typically use binary black and white stimuli, rather than ternary black, gray, and white stimuli, as in the present experiment.

Negative polarity is known to be preferable only for special conditions, such as the degeneration of visual acuity in aging [26: 244–243] or disease [109]. Also, aeronautical displays with negative polarity help to maintain the adaptation of the visual system to a dark environment during night flight [148]. Closer to papyrology is the similar case of readers repeatedly switching between the high luminance of a computer screen and the low luminance of paper documents [20]. These exceptions have a context of *low vision* in common, whether permanent or transitory, that demands increased visual and cognitive efforts. In other words, they are akin to critical reading tasks, such as the decipherment of ancient documents, characterized by high visual and semantic entropy.

In terms of *visual ecology*, positive contrast has dominated the history of writing, albeit mostly for practical reasons such as the cost of colored substrate. Negative polarity texts are often prestige objects, for example, the Rosetta inscription that features a light script with crystalline sparkle on a dark stone [122]; the word “God” calligraphed in gold on a green eight-meter-high panel in the Aya Sofia mosque of Istanbul [25: 504]; and the 12th-century “Blue Sutra” of Japan, in gold on blue paper [58: 36, 68–69]. These examples are also sources of inspiration for how contrast — and thus legibility — might be further improved, using displays with special material properties, such as high-reflectance or fluorescence.

#### 5.3.11 Smooth human–computer interaction is critical

Participant AW reported: “Even with the split-screen option, and using two monitors, I felt like my devices lacked the optimal Graphical User Interface to maximize the usefulness of those pictures. Comparing the same zoomed part of text through several images proved, for example, to be quite costly in terms of time and clicks, and I feel this disturbed the workflow.” Five of eight participants juxtaposed two or more image variants during transcription; three used two displays, one used three. Two participants printed the images (as more documents fit on a table than on a computer display).

These explicit and implicit statements demonstrate how interaction with the images can impact the advantages derived from image processing. Indeed, without good interaction, there is little enhancement as far as the user is concerned. What, then, is *good interaction*? Observing the participants is informative in this regard: good interaction involves a wide visual field, the synchronization of manipulations across multiple images, the understanding of how different enhancement methods have varying utility for transcription, and a state of mental flow during task completion [42].

#### 5.3.12 Gender may affect legibility enhancement

One female participant in the experiment had perfect agreement in her method ratings, while the second female had the second highest agreement score (Table 5B). If they are not statistical outliers, then the gender disparity in behavior could have implications, given the increased proportion of females in papyrology (202 [57%] out of 356 participants at the International Congress of Papyrology in 2019 were female, compared to 3 [5%] out of 57 at the congress in 1937 [37, 6]). For example, females have been found to have greater chromatic discrimination ability, whereas males are more sensitive to lightness variation [189]. Therefore, the enhancement method through chromatic contrasting might improve legibility more for females, while the negative polarity method could be more appropriate for males. A larger sample size would be needed to further the investigation of the topic.

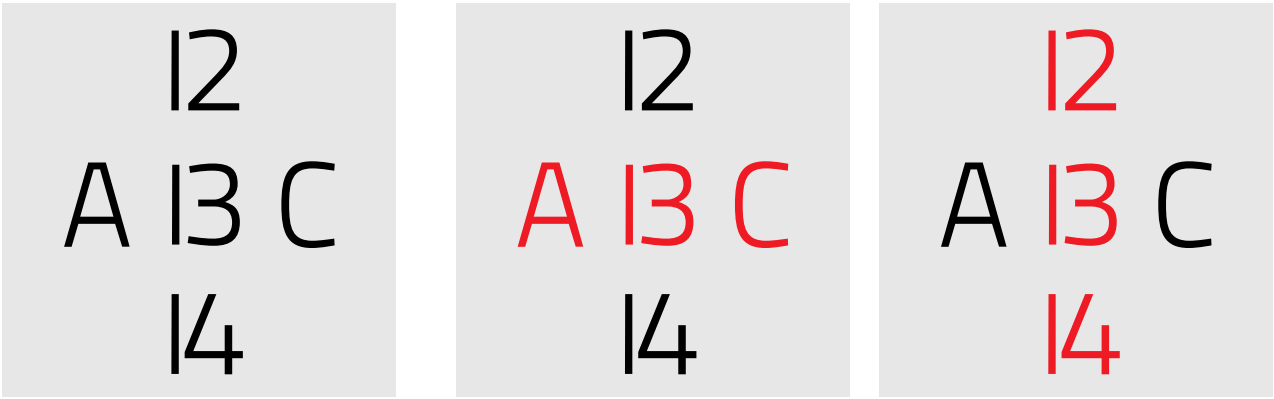

**Fig. 17** The shapes in the center of the squares are perfectly legible, but ambiguous: do they represent the letter “B”, the figure “13”, or something else? Coloring helps differentiate between alternatives. Such shapes are said to be “multistable stimuli” [101], and have been the source of well known illusions, e.g. Arcimboldo’s face made of books [164: 131].

However, this section is representative of an issue with broader significance. In addition to gender, there are numerous other perceptual variability factors (age-related, genetic, cultural, and climatic, among others [49: 149–244]) that indicate the utility of demographically tailored and even personalized image enhancements (*in a similar fashion to how corrective glasses are fitted individually*) rather than those calibrated to the statistical mean of the general population.

## 6 Paradigm

A paradigm of legibility enhancement for critical applications is provided herein, representing a step towards developing computer-assisted critical vision systems. This paradigm helps in defining the goals, methods, and limits of legibility enhancement, and is rooted in the problem analysis and experiments carried out in the present research. The preceding sections have circumscribed the notion of critical vision, established that potential information should not be suppressed, and advocated a systems approach to the design of enhancement software. A further element will be discussed below, namely uncertainty; this element will help us to operationalize the paradigm and link the field of image enhancement to the vast body of interdisciplinary research on decision making under uncertainty.

### 6.1 Make uncertainty explicit

*Requirements* — Because vision is purposeful, we might sometimes stare at things without seeing them when there is no reason to seek them out (“inattentional blindness” [169]). Two converse problems concern our tendency to believe what we see and see what we believe (“confirmation bias” [168: 61–75]). Generally speaking, the ideal enhancement system would inform us about objective uncertainty in images and guard against subjective cognitive biases during their interpretation. Uncertainty is important and its presence should be made explicit. The system should therefore aid in its detection and analysis, represent it visually and numerically, and preserve it for verification. *Uncertainty management should be at the core of the enhancement system and its components, from the algorithms to the interface.*

*Examples* — One illustrative analogy is the use of white material for pottery reconstruction in archeology to distinguish between missing and original parts. An example of script polysemy and the role of context in reading is shown in Fig. 17, wherein the central shapes can be interpreted as either letters or numbers depending on the reading direction. Mundane cases where an individual seeks to mentally solve visual ambiguities include faded maps or letters and frightening shadows at night, while imagery-based clinical diagnosis and military decisions represent domains in which uncertainty is both pervasive and critical. For the facial identification of suspects, police organizations have developed mechanical and computational aids for using witness recollections to produce faces composited from a catalog of elements, such as eyes, nose, and mouth, assembled by likelihood into a coherent whole [43].

*Rationale* — A paradigm called *conservative preprocessing* was proposed by Chen, Lopresti, and Nagy in a rare work on the theory of image enhancement [38]. It consists in preserving the original by means of reversible transformations and the use of derivative representations as a proxy. Other researchers have noted that applying the concept of noise suppression outside the casual reading task might be misplaced: for example, bleed-through removal may unwittingly obliterate the only trace of text that is inaccessible because the page is glued to the support, was not scanned, or belongs to a missing second or third page [179]. Such invasive enhancement can thus destroy historical, forensic, and conservation-related information in documents and reproductions [129, 60]. Problems of this kind are common in digital library projects. Technical factors can also make enhancement problematic; for example, because band-pass filtering is known to frequently suppress script parts [133: 445], while neural networks can "hallucinate" character-like shapes [21: 1221], should the user not be made aware of the possibility of erroneous readings?

*Conceptualization* — Let us here define uncertainty in the context of image analysis and consider possible solutions. As discussed above, image enhancement is the outcome of interacting stimuli, observers, tasks, contexts, and technologies. Multiple concurrent enhancements therefore appear to be better adapted to the high combinatorial variability potential of such a SYSTEM.

To clarify the uncertainty reduction PROCESS, we draw on cognitive psychology (in particular, decision-making in critical contexts, such as those encountered by medical emergency services, firefighters, police, and military personnel [168]). The process consists of an *interpretation* stage, in which possible stimuli of interest are detected and separated from the surrounds (building "situational awareness"), various options as to their identification are formulated (integration into coherent "gestalts"), and their likelihood is estimated (multidimensional weighting and/or ranking); this is followed by a *decision* stage, in which a choice of interpretations is made. For our proposed method, it is noteworthy that (1) *alternatives* to the visual stimulus are produced, (2) the visualization is *mental*, rather than physical, and (3) it involves mental *manipulations* of the alternative images.

Establishing a taxonomy of common FACTORS leading to uncertainty helps us to find appropriate remedies. The inability to detect visual targets (*lack* of information) and vaguely defined targets (*fuzzy* information) are types of uncertainty that arise early in the interpretation process and that may be countered by more and sharper imagery; that is, alternative views. Decisions regarding *implausible* interpretations may be solved by considering the target's compatibility with its visual context, which shows the utility of retaining context and even using alternative backgrounds. The uncertainty developing from the need to choose between *contradictory* interpretations demands an increased capability to compare alternatives. Entities need not necessarily be noisy to make interpretation difficult (viz. readily distinguishable characters in unknown scripts); to deal with *novelties*, information exogenous to the image and observer must be supplied. Finally, an *overabundance* of potential interpretations may become overwhelming and result in "decision inertia" [168: 37–41]. Information visualization theory suggests for such cases filtering the amount of information, through selection or ranking [167].

In terms of LIMITATIONS, he number of items processable at a given time in human memory is however finite and can degrade under time pressure and stress; similarly, the visual field size is also limited and increasingly blurred as it moves away from the fovea. Thus, methods designed to *increase* the number of alternatives must be accompanied by mechanisms to *limit* visible alternatives and *manage* all available stock.

*Operationalization* — We propose to make uncertainty materially explicit through the physical rendering and cueing of mental interpretations of images (as a literal application of the notion that "seeing is believing"). When the alternatives are known, highlighting or inpainting of objects may be used to find fits between potential targets and a given context, as in a puzzle game where the pieces are materializations of mental images. For unknown alternatives, solutions may be elicited by purposive or random alterations at the global or local level, statically or as a movie, similar to the shimmering reflection on an undulating water surface, or searching for optimal shapes by sketching, or tilting a hologram to observe it from different perspectives (our deliberate use of metaphors is a method for problem solving, as well as for communication [106, 75]). Machine learning may help shape prediction and synthesis, while human–machine interaction techniques are used to guide the image exploration. The result is much more than an image enhancement algorithm: it is a *computer-assisted critical vision system*. Current research on such systems includes applications for, e.g., medical radiographic diagnosis [158: 359–414].

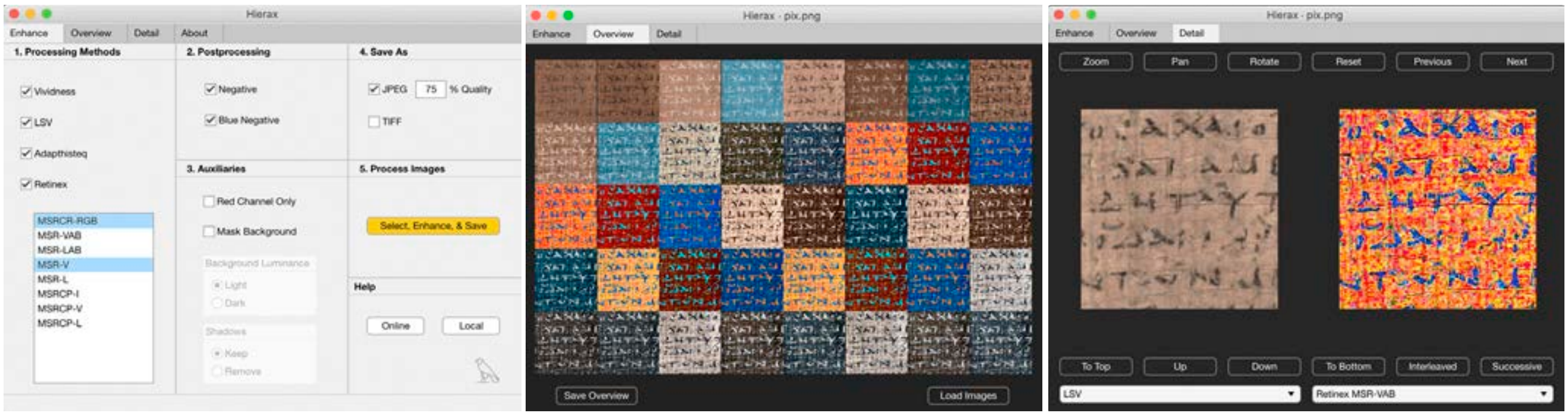

**Fig. 18** The graphical user interface of the Hierax software for papyri legibility enhancement, which implements the methods described in this article. Method selection and parametrization (left); overview of all processed images (center); image comparison and interaction panel (right).

This method operates as illustrated using Fig. 17, in which two images were derived from the original; each of these make explicit a different interpretation through the perceptual differentiation of distinct semantic classes. If semantic grouping became unfeasible, the random alternation of individual shapes between two colors would be the next best option (also revealing a less obvious solution when only the three vertical bars are red, in that they may represent the upper-case letter “I”, the lower-case letter “l”, or the figure “1”). The inpainting technique presented in this example consists in removing the gaps between adjacent central shapes to strengthen their identity as the letter “B”. In the real case of papyri, the most common character allographs of interest may be statically inpainted or flickered. This is a complex and possibly contentious intervention in the image structure; thus, in our implementation, we avoided segmentation by opting to present variants of the whole scene.

*Frameworks* — The presented method for making uncertainty explicit in view of helping decision-making may be understood in information theoretical terms as finding a balance between maximal (= quandary) and minimal choice (= no or unknown alternative) by comparing image alternatives with different entropies for different entities within the images. The heterogeneity of user behavior observed in our experiment suggests that the optimum is contextually determined by a combination of factors.

Bayesian theory offers an additional useful theoretical framework, in that establishing the identity of the visual target is dependent on both the prior assumptions of the observed shape having the hypothesized identity and the likelihood of observing the hypothetical entity in the given context. In this sense, the explicit presentation of multiple alternatives for comparison is a means of updating and optimizing one’s beliefs about the identity of visual stimuli.

### 6.2 Summary of criteria

Our work has yielded a set of core criteria that define our paradigm of legibility enhancement for critical applications::

1. Critical vision — Legibility enhancement for critical applications is a matter of critical vision. It consists in exercising skepticism about the interpretation of images.
2. Potential information — Any image structure should be viewed as potential information that should not be suppressed in critical vision applications; interpretations can arise from connections made between image parts.
3. Uncertainty explicitation — Support for critical vision is realized by making alternative image interpretations materially explicit. Visual target identity depends on both the shape and its context.
4. Systems approach — Legibility enhancement for critical applications is a system problem. Optimization depends on tasks, data, users, tools, and their interactions.

## 7 Implementation

The research findings were implemented in the open-source software Hierax [16]. To account for the heterogeneity of user behavior and the complementarity of enhancements, a range of methods were offered: our novel methods, and adapthisteq and retinex. Since interaction with the images was found to affect the usefulness of enhancements, the software includes an image viewer. Our design methodology conforms to elements of the so-called information visualization “mantra” for graphical user interfaces: overview, filter, detail, and context [167]. One interface tab presents an overview of one image enhanced using the selected methods (Fig. 18); through the spatial juxtaposition of patterns, the resulting mosaic facilitates understanding of the characteristics of different enhancement methods. A second tab allows for detailed side-by-side comparison, with zooming, panning, and rotation being synchronized; the diptych view is intended for the interpretation of details in close coupling of vision and motor control. Rapid switching between images via the keyboard is a temporal image analysis modality that leverages attentional capture through flicker, introduced to help detect potential features of interest [192]. The mosaic, diptych, and flicker image analysis techniques implement the paradigm of uncertainty explicitation at the whole scene level. These facilities were appreciated by the paleographers who beta-tested the software. Hierax has already been employed in a real-world setting, to enhance the 70 papyri of the University of Basel Library for a forthcoming online catalog.

## 8 Conclusions

This work aimed to improve the legibility of ancient papyri for text editions. Novel enhancement methods were developed on the basis of color processing and visual illusions. A user experiment demonstrated that these methods outperform classical and more complex enhancement methods. Future work could examine whether the new methods could also improve reading by machines.

The experiment also yielded unexpected findings. Critical for the software, user behavior was found to be *heterogeneous* in type, and not only variable in intensity. Thus, better performance might be achieved by *personalization*, *contextualization*, and *complementarity*, instead of searching for an overall optimum that may be spurious or counterproductive. Experiments with more users and images should be conducted to strengthen, nuance, and perhaps diversify the findings.

The findings also lead to a paradigm of legibility enhancement for critical applications, in view of a future computer-aided critical vision system.

**Author contributions** — V. A. conceived the algorithms, programmed the software, designed the experiment, analyzed the results, and wrote the article. I. M.-S., project leader, defined, planned, and supervised the research (data collection, guidance, tests, experiment coordination, critical revision of the manuscript).

**Acknowledgments** — This article is an output of the project n° PZ00P1_174149, *Reuniting fragments, identifying scribes and characterizing scripts: the digital paleography of Greek and Coptic papyri*, funded by the Swiss National Science Foundation. The experimental participants are heartily thanked for their valuable efforts. Gorge Nagy, Andreas Fischer, Irvin Schick, and Bernice Rogowitz, as well as the three anonymous reviewers, offered much appreciated critique. The authors wish to acknowledge the proofreaders' help in improving the manuscript readability: Jacqueline M. Huwyler, Thomas Hardy of Science Journal Editors, and K. F. of Proofed. — Dedicated by V. A. to Aries Arditi and Ruth Rosenholtz, for the vision from the lighthouse.

## References

1. Abidi BR, Zheng Y, Gribok AV, Abidi MA (2006). "Improving Weapon Detection in Single Energy X-Ray Images Through Pseudocoloring", *IEEE Trans. on Systems, Man, and Cybernetics, Part C (Applications and Reviews)*, 36 (6): 784–796.
2. Adelson EH (2000). "Lightness Perception and Lightness Illusions", in Gazzaniga MS, ed., *The New Cognitive Neurosciences*, Cambridge, MA: MIT Press, 2$^{nd}$ ed., pp. 339–351.
3. Afruz J, Wilson V, Umbaugh SE (2010). "Frequency Domain Pseudo-color to Enhance Ultrasound Images", *Computer and Information Science*, 3 (4): 24–34.
4. Alaei A, Raveaux R, Conte D, Stantic B (2018). " "Quality" vs. "Readability" in Document Images: Statistical Analysis of Human Perception", *Proc. 13th IAPR Intl. Workshop on Document Analysis Systems (DAS), 24–27 April 2018, Vienna, Austria*, pp. 363–368.
5. Alexopoulou AA, Kaminari AA, Panagopoulos A, Pöhlmann E (2013). "Multispectral documentation and image processing analysis of the papyrus of tomb II at Daphne, Greece", *J. of Archaeological Science*, 40 (2): 1242–1249.
6. Allberry CRC *et al.* (1938). *Actes du V$^e$ Congrès international de papyrologie, Oxford, 30 Août – 3 Septembre 1937*, Bruxelles: Fondation égyptologique Reine Élisabeth.
7. Allegra D, Ciliberto E, Ciliberto P, Milotta FLM, Petrillo G, Stanco F, Trombatore C (2015). "Virtual Unrolling Using X-ray Computed Tomography", *Proc. 23$^{rd}$ European Signal Processing Conf. (EUSIPCO), 31 August – 4 September 2015, Nice, France*, pp. 2864–2868.
8. Alley RE (1996). *Algorithm Theoretical Basis Document for Decorrelation Stretch*, ASTER-AST06, Pasadena, CA: Jet Propulsion Laboratory, http://www.dstretch.com/DecorrelationStretch.pdf.
9. Altamura O, Berardi M, Ceci M, Malerba D, Varlaro A (2007). "Using colour information to understand censorship cards of film archives", *Intl. J. on Document Analysis and Recognition*, 9 (1): 281–297.
10. Alter AL, Oppenheimer DM, Epley N, Eyre RN (2007). "Overcoming Intuition: Metacognitive Difficulty Activates Analytic Reasoning", *J. of Experimental Psychology*, 136 (4): 569–576.
11. Alvo M, Yu PhLH (2014). *Statistical Methods for Ranking Data*, New York, NY: Springer.
12. Arsene CTC, Church S, Dickinson M (2018). "High Performance software in Multidimensional Reduction Methods for Image Processing with Application to Ancient Manuscripts", *manuscript cultures*, 11: 73–96.
13. Asano Y, Fairchild MD, Blondé L (2016). "Individual Colorimetric Observer Model", *PLoS ONE*, 11 (2): e0145671.
14. Atanasiu V (2018). "Ugraphia: The Utopia of a Perfectly Legible Script", /gʁafematik/ *Collection of extended abstracts of the conference Graphemics in the 21st century: From graphemes to knowledge, 14–15 June 2018, Brest, France*, pp. 113–140.
15. Atanasiu V (2020). "Non-parametric rating and ranking functions", *MATLAB Central File Exchange*, https://www.mathworks.com/matlabcentral/fileexchange/78384.
16. Atanasiu V, Marthot-Santaniello I (2020). *Hierax Legibility Enhancement Software*, https://hierax.ch.
17. Bagnall RS (2009). *The Oxford Handbook of Papyrology*, Oxford: Oxford Univ. Press.
18. Balinski M, Rida L (2010). *Majority Judgment Measuring, Ranking, and Electing*, Cambridge, MA: MIT Press.
19. Barten PGJ (1999). *Contrast Sensitivity of the Human Eye and Its Effects on Image Quality*, Bellingham, WA: SPIE Optical Engineering Press.
20. Bauer D, Cavonius C (1980). "Improving the legibility of visual display units through contrast reversal", in Grandjean E and Vigliani E, eds., *Ergonomic aspects of visual display terminals*, London: Taylor & Francis, pp. 137–142.
21. Belthangady Ch, Royer LA (2019). "Applications, promises, and pitfalls of deep learning for fluorescence image reconstruction", *Nature Methods*, 16: 1215–1225.
22. Berns RS (2014). "Extending CIELAB: Vividness, $V^*_{ab}$, Depth, $D^*_{ab}$, and Clarity, $T^*_{ab}$", *Color Research and Application*, 39 (4): 322–330.
23. Beutel J, Van Metter RL, Kundel HL, eds (2000). *Handbook of Medical Imaging*, Bellingham, WA: SPIE, vol. 1.
24. Bhowmik Sh., Sarkar R, Nasipuri M, Doermann D (2018). "Text and non-text separation in offline document images: a survey", *Intl. J. on Document Analysis and Recognition*, 21 (1–2): 1–20.
25. Blair Sh (2006). *Islamic Calligraphy*, Edinburgh: Edinburgh Univ. Press.
26. Boff KR, Lincoln JE, eds. (1988). *Engineering Data Compendium: Human Perception and Performance*. Ohio, OH: Harry G. Armstrong Aerospace Medical Research Laboratory, Wright-Patterson

Air Force Base, vol. 1.
27. Bouwmans Th, Javed S, Zhang H, Lin Zh, Otazo R (2018). “On the Applications of Robust PCA in Image and Video Processing”, *Proceedings of the IEEE*, 106 (8): 1427–1457.
28. Brenner S, Sablatnig R (2021). “Subjective Assessments of Legibility in Ancient Manuscript Images - The SALAMI Dataset”, *Proc. ICPR Intl. Workshops and Challenges, 10–15 January 2021, Virtual Event*, part VII: 68–82.
29. Breuil C, Jennings BJ, Barthelmé S, Guyader N, Kingdom FAA (2019). “Color improves edge classification in human vision”, *PLoS Computational Biology*, 15 (10): e1007398.
30. Brink A, van der Klauw H, Schomaker L (2008). “Automatic removal of crossed-out handwritten text and the effect on writer verification and identification”, *Proc. SPIE–IS&T Electronic Imaging Conf., Document Recognition and Retrieval XV, 29–31 January 2008, San Jose, CA, USA*, vol. 6815: 68150A.
31. Broersen PMT (2006). *Automatic Autocorrelation and Spectral Analysis*, London: Springer.
32. Buchner A, Mayr S, Brandt M (2009). “The advantage of positive text-background polarity is due to high display luminance”, *Ergonomics*, 52 (7): 882–886.
33. Buckle CE, Udawatta V, Straus CM (2013). “Now You See It, Now You Don’t: Visual Illusions in Radiology”, *RadioGraphics*, 33 (7): 2087–2102.
34. Bülow-Jacobsen A (2009). “Writing Materials in the Ancient World”, in Bagnall RS, *The Oxford Handbook of Papyrology*, Oxford: Oxford Univ. Press, pp. 3–29.
35. Bülow-Jacobsen A (2021). “Photography of Papyri and Ostraca”, in Caputo C, Lougovaya J, eds., *Using Ostraca in the Ancient World. New Discoveries and Methodologies*, Berlin: De Gruyter, pp. 59–83.
36. Cannon M, Hochberg J, Kelly P (1999). “Quality assessment and restoration of typewritten document images”, *Intl. J. on Document Analysis and Recognition*, 2 (2–3): 80–88.
37. Capasso M, Davoli P (2019). *Abstracts of the 29th Intl. Cong. of Papyrology, 28 July – 8 August 2019, Lecce, Italy*, Lecce: Università del Salento.
38. Chen J, Lopresti D, Nagy G (2016). “Conservative preprocessing of document images”, *Intl. J. on Document Analysis and Recognition*, 19 (4): 321–333.
39. Christen M, Vitacco DA, Huber L, Harboe J, Fabrikant SI, Brugger P (2013). “Colorful brains: 14 years of display practice in functional neuroimaging”, *NeuroImage*, 73 (6): 30–39.
40. CIE (2004). *Colorimetry*, TR 15:2004, Vienna: CIE.
41. Ciortan I, Deborah H, George S, Hardeberg JY (2015). “Color and Hyperspectral Image Segmentation for Historical Documents”, *Proc. 2015 Digital Heritage Conf., 28 September – 2 October 2015, Granada, Spain*, pp. 199–206.
42. Csikszentmihalyi M (2008). *Flow: The Psychology of Optimal Experience*, New York, NY: HarperPerennial.
43. Davies G, Valentine T (2014). “Facial Composites: Forensic Utility and Psychological Research”, Lindsay RCL, Ross DF, Read JD, Toglia MP (eds.), *Handbook Of Eyewitness Psychology: Volume 2, Memory For People*, New York, NY: Psychology Press, pp. 59–82.
44. DIB (2021). *Document Image Binarization*, https://dib.cin.ufpe.br.
45. Donofrio RL (2011). “Review Paper: The Helmholtz-Kohlrausch Effect”, *J. of the Society for Information Display*, 19 (10): 658–664.
46. Drira F, LeBourgeois F (2017). “Mean-Shift segmentation and PDE-based nonlinear diffusion: toward a common variational framework for foreground/background document image segmentation”, *Intl. J. on Document Analysis and Recognition*, 20 (3): 201–222.
47. Easton RL, Knox KT, Christens-Barry WA (2003). “Multispectral imaging of the Archimedes palimpsest”, *Proc. 32nd Applied Imagery Pattern Recognition Workshop (IEEE-AIPR’03), 15–17 October 2003, Washington, DC, USA*, pp. 111–116.
48. Elhedda W, Mehri M,Mahjoub MA (2020). “Hyperkernel-based intuitionistic fuzzy c-means for denoising color archival document images”, *Intl. J. on Document Analysis and Recognition*, 23 (3): 161–181.
49. Elliot AJ, Fairchild MD, Franklin A, eds. (2015). *Handbook of Color Psychology*, Cambridge: Cambridge Univ. Press.
50. Fairchild MD (1996). “Refinement of the RLAB Color Space”, *Color Research and Application*, 21 (5): 338–346.
51. Fairchild MD (2013). *Color Appearance Models*. Chichester: Wiley, 3rd ed.
52. Fairchild MD, Pirrotta E (1991). “Predicting the lightness of chromatic object colors using CIELAB”, *Color Research and Application*, 16 (6): 385–393.
53. Farup I (2020). “Individualised Halo-Free Gradient-Domain Colour Image Daltonisation”, *J. of Imaging*, 6 (116): 1–10.
54. Fiorucci M, Khoroshiltseva M, Pontil M, Traviglia A, Del Bue A, James S (2020). “Machine Learning for Cultural Heritage: A Survey”, *Pattern Recognition Letters*, 133 (5): 102–108.
55. Fitts PM, ed. (1947). *Psychological Research on Equipment Design*, Report no. 19, Army Air Forces: Washington D.C.
56. de la Flor G, Luff P, Jirotka M, Pybus J, Kirkham R, Carusi A (2010). “The Case of the Disappearing Ox: Seeing Through Digital Images to an Analysis of Ancient Texts”, *Proc. Conf. on Human Factors in Computing Systems (CHI 2010), 10–15 April 2010, Atlanta, GA, USA*, pp. 473–482.
57. Fréchette GR, Schotter A (2015). *Handbook of Experimental Economic Methodology*, Oxford: Oxford University Press.
58. Fu Sh, Lowry GD, Yonemura A (1986). *From Concept to Context: Approaches to Asian and Islamic Calligraphy*, Washington, DC: Smithsonian Institution.
59. Gao XW, Loomes M (2016). “A new approach to image enhancement for the visually impaired”, *Proc. IS&T Intl. Symp. on Electronic Imaging 2016, Color Imaging XXI: Displaying, Processing, Hardcopy, and Applications, 14–18 February 2016, San Francisco, CA, USA*, COLOR-325: 1–7.
60. Gargano M, Bertani D, Greco M, Cupitt J, Gadia D, Rizzi A (2015). “A perceptual approach to the fusion of visible and NIR images in the examination of ancient documents”, *J. of Cultural Heritage*, 16 (4): 518–525.
61. George S, Grecicosei AM, Waaler E, Hardeberg JY (2014). “Spectral Image Analysis and Visualisation of the Khirbet Qeiyafa Ostracon”, *Proc. 6th Intl. Conf. on Image and Signal Processing (ICISP 2014), 30 June – 2 July 2014, Cherbourg, France*, 272–279.
62. Gevers Th, Gijsenij A, van de Weijer J, Geusebroek J-M (2012). *Color in Computer Vision: Fundamentals and Applications*. Chichester: Wiley.
63. Ghadiyaram D, Bovik AC (2016). “Massive Online Crowdsourced Study of Subjective and Objective Picture Quality”, *IEEE Trans. on Image Processing*, 25 (1): 372–387.
64. Gibson A, Piquette KE, Bergmann U, Christens-Barry W, Davis G, Endrizzi M, Fan Sh, *et al.* (2018). “An assessment of multimodal imaging of subsurface text in mummy cartonnage using surrogate papyrus phantoms”, *Heritage Science*, 6: 7.
65. Gilchrist A (2006). *Seeing Black and White*, Oxford: Oxford Univ. Press.
66. Gillespie AR, Kahle AB, Walker, RE (1986). “Color enhancement of highly correlated images: I. Decorrelation and HSI contrast stretches”, *Remote Sensing of Environment*, 20: 209–235.
67. Gonzalez A, Benavente R, Penacchio O, Vazquez-Corral J, Vanrell M, Parraga CA (2013). “Coloresia: An Interactive Colour Perception Device for the Visually Impaired”, in Sappa AD, Vitrià J, eds., *Multimodal Interaction in Image and Video Applications*, Berlin: Springer, pp.47-66.
68. Govindaraju V, Srihari SN (1995). “Image Quality and Readability”,

*Proc. Intl. Conf. on Image Processing (ICIP), 23–26 October 1995, Washington, DC, USA*, vol. 3, pp. 324–327.
69. Hansen Th, Gegenfurtner KR (2009). "Independence of color and luminance edges in natural scenes", *Visual Neuroscience*, 26: 35–49.
70. Hariharan H, Koschan A, Abidi B, Gribok A, Abidi M (2006). "Fusion of Visible and Infrared Images using Empirical Mode Decomposition to Improve Face Recognition", *Proc. Intl. Conf. on Image Processing, 8–11 October 2006, Atlanta, GA, USA*, pp. 2049–2052.
71. Harman J (2008). "Using Decorrelation Stretch to Enhance Rock Art Images", *DStretch: Rock Art Digital Enhancement*, website, http://www.dstretch.com/AlgorithmDescription.html.
72. He Sh, Schomaker L (2019). "DeepOtsu: Document enhancement and binarization using iterative deep learning", *Pattern Recognition*, 91 (7): 379–390.
73. Hernández F-J (2010). *Retinex for ImageJ*, https://web.archive.org/web/20130323053256/http://www.dentistry.bham.ac.uk/landinig/software/retinex/retinex.html.
74. Hidalgo HL, España S, Castro MH, Pérez HA (2005). "Enhancement and Cleaning of Handwritten Data by Using Neural Networks", *Proc. Iberian Conference on Pattern Recognition and Image Analysis (IbPRIA), 7-9 June 2005, Estoril, Portugal*, part I, pp. 376–383.
75. Hofstadter D, Fluid Analogies Research Group (1995). *Fluid Concepts & Creative Analogies*, New York, NY: BasicBooks.
76. Holleman GA, Hooge ITC, Kemner C, Hessels RS (2020). "The 'Real-World Approach' and Its Problems: A Critique of the Term Ecological Validity", *Frontiers in Psychology*, 11: 721.
77. Homer (1990). *The Iliad*, trans. Robert Fagles, New York, NY: Penguin.
78. Hoßfeld T, Keimel Ch, Hirth M, Gardlo B, Habigt J, Diepold K, Tran-Gia Ph (2014). "Best Practices for QoE Crowdtesting: QoE Assessment With Crowdsourcing", *IEEE Trans. on Multimedia*, 16 (2): 541–558.
79. Hummel R (1977). "Image enhancement by histogram transformation", *Computer Graphics and Image Processing*, 6: 184–195.
80. James AP, Dasarathy BV (2014). "Medical image fusion: A survey of the state of the art", *Information Fusion*, 19: 4–19.
81. Jobson DJ, Rahman Z, Woodell GA (1997). "A Multiscale Retinex for Bridging the Gap Between Color Images and the Human Observation of Scenes", *IEEE Trans. on Image Processing*, 6 (7): 965–976.
82. Hyoung-Gook K, Moreau N, Sikora Th (2005). *MPEG-7 Audio and Beyond: Audio Content Indexing and Retrieval*, Chichester: Wiley.
83. Kahneman D (2011). *Thinking, Fast and Slow*, London: Penguin.
84. Kambara A (2017). "Effects of Experiencing Visual Illusions and Susceptibility to Biases in One's Social Judgments", *Perceptual and Motor Skills*, 7 (4): 1–6.
85. Kapoor A, Caicedo JC, Lischinski D, Kang SB (2014). "Collaborative Personalization of Image Enhancement", *Intl. J. of Computer Vision*, 108 (1–2): 148–164.
86. Karatzas D, Antonacopoulos A (2007). "Colour text segmentation in web images based on human perception", *Image and Vision Computing*, 25: 564–577.
87. Keelan BW (2002). *Handbook of Image Quality: Characterization and Prediction*, New York, NY: Marcel Dekker.
88. Kemeny JG, Snell JL (1972). *Mathematical Models in the Social Sciences, Cambridge*, MA: MIT Press.
89. Kendall MG (1970). *Rank Correlation Methods*, New York, NY: McGraw-Hill, 4th ed.
90. Kihlstrom JF (2021). Ecological Validity and "Ecological Validity", *Perspectives on Psychological Science*, 16 (2): 466–471.
91. Kim SJ, Deng F, Brown MS (2011). "Visual enhancement of old documents with hyperspectral imaging", *Pattern Recognition*, 44: 1461–1469.
92. Kim HU, Koh YJ, Kim CS (2020). "PieNet: Personalized Image Enhancement Network", *Proc. Computer Vision (ECCV 2020), 23–28 August 2020, Glasgow, UK*, pp. 374–390.
93. Kim J, Vora A, Peli E (2004). "MPEG-based image enhancement for the visually impaired", *Optical Engineering*, 43 (6): 1318–1328.
94. Kim SJ, Zhuo Sh., Deng F, Fu Ch-W, Brown MS (2010). "Interactive Visualization of Hyperspectral Images of Historical Documents", *IEEE Trans. on Visualization and Computer Graphics*, 16 (6): 1441–1448.
95. Kligler N, Katz S, Tal A (2018). "Document Enhancement using Visibility Detection", *Proc. 2018 IEEE/CVF Conf. on Computer Vision and Pattern Recognition, 18–23 June 2018, Salt Lake City, UT, USA*, pp. 2374–2382.
96. Koo TK, Li MY (2016). "A Guideline of Selecting and Reporting Intraclass Correlation Coefficients for Reliability Research", *J. of Chiropractic Medicine*, 15 (2): 155–163.
97. Koschan A, Abidi M (2008). *Digital Image Processing*, Chichester: Wiley.
98. Kovesi P (2015). "Good Colour Maps. How to Design Them", *arXiv.org*, https://arxiv.org/abs/1509.03700.
99. Kowalik G, Nielek R (2016). "Senior Programmers: Characteristics of Elderly Users from Stack Overflow", *Proc. 8th Intl. Conf. on Social Informatics (SocInfo 2016), 11–14 November 2016, Bellevue, WA, USA*, part II, pp. 87–96.
100. Krupinski EA (2009). "Practical Applications of Perceptual Research", in Beutel J, Van Metter RL, Kundel HL, eds., *Handbook of Medical Imaging, Bellingham*, WA: SPIE, vol. 1, pp. 895–929.
101. Kruse P, Stadler M, eds. (1995). *Ambiguity in Mind and Nature: Multistable Cognitive Phenomena*, Berlin: Springer.
102. Kuhlthau CC (2004). *Seeking Meaning: A Process Approach to Library and Information Services*, Westport, CT: Libraries Unlimited.
103. Kuncheva LI (2014). *Combining pattern classifiers : methods and algorithms*, New Jersey, NJ: Wiley, 2nd ed.
104. Kundel HL (2000). "Visual Search in Medical Images", in Beutel J, Van Metter RL, Kundel HL, eds., *Handbook of Medical Imaging, Bellingham*, WA: SPIE, vol. 1, pp. 837–858.
105. Labaune J, Jackson JB, Pagès-Camagna S, Menu M, Mourou GA (2010). "Terahertz investigation of Egyptian artifacts", *Proc. 35th Intl. Conf. on Infrared, Millimeter, and Terahertz Waves, 5–10 September 2010, Rome, Italy*, pp. 1–3.
106. Lakoff G, Johnson M (1980). *Metaphors We Live By*, Chicago: The University of Chicago Press.
107. Land EH (1977). *The Retinex Theory of Color Vision*, Scientific American, 237 (6): 108–128.
108. Langville AN, Meyer CD (2012). *Who's #1?: The Science of Rating and Ranking*, Princeton, NJ: Princeton Univ. Press.
109. Legge GE, Rubin GS, Schleske MM (1987). "Contrast-polarity effects in low-vision reading", in Woo GC, ed., *Low Vision: Principles and Applications*, New York, NY: Springer, pp. 288–307.
110. Lettner M, Sablatnig R (2009). "Spatial and Spectral Based Segmentation of Text in Multispectral Images of Ancient Documents", *Proc. 10th Intl. Conf. on Document Analysis and Recognition, 26–19 July 2009, Barcelona, Spain*, pp. 813–817.
111. Li X (2013). "Image Denoising: Past, Present, and Future", Gunturk BK, Li X, eds., *Image Restoration: Fundamentals and Advances*, Boca Raton, FL: CRC Press, pp. 1–19.
112. Lin Y, Seales WB (2005). "Opaque Document Imaging: Building Images of Inaccessible Texts", *Proc. 10th IEEE Intl. Conf. on Computer Vision (ICCV'05), 17–21 October 2005, Beijing, China*, vol. 1, pp. 662–669.
113. Lins RD, Almeida MM, Bernardino RB, Jesus D, Oliveira JM (2017). "Assessing Binarization Techniques for Document Images", *Proc. of the ACM Document Engineering Conf. (DocEng'17), 4–7 September 2017, Valletta, Malta*, pp. 183–192.
114. Lund O (1999). *Knowledge construction in typography: the case of legibility research and the legibility of sans serif typefaces*, PhD thesis, University of Reading, Reading.

115. Lombardi F, Marinai S (2020). “Deep Learning for Historical Document Analysis and Recognition—A Survey”, *J. of Imaging*, 6 (10): 110.
116. Machado GM, Oliveira MM, Fernandes LAF (2009). “A Physiologically-based Model for Simulation of Color Vision Deficiency”, *IEEE Trans. on Visualization and Computer Graphics*, 15 (6): 1291–1298.
117. Marinai S, Karatzas D (2011). “Report from the AND 2009 working group on noisy text datasets”, *Intl. J. on Document Analysis and Recognition*, 14 (2): 113–116.
118. MathWorks (2020). “Contrast Adjustment”, *Matlab: Image Processing Toolbox*, https://www.mathworks.com/help/images/contrast-adjustment.html.
119. McCann JJ (2017). “Retinex at 50: color theory and spatial algorithms, a review,” *J. of Electronic Imaging*, 26 (3): 031204.
120. McIntyre D (2002). *Colour blindness: Causes and Effects*, Chester: Dalton Publishing.
121. Mehta R, Zhu R (2009). “Blue or Red? Exploring the Effect of Color on Cognitive Task Performances”, *Science*, 323 (5918): 1226–1229.
122. Miller E, Lee NJ, Uprichard K, Daniels V (2000). “The Examination and Conservation of the Rosetta Stone at the British Museum”, in Roy A, Smith P, eds., *Tradition and Innovation: Advances in Conservation*, The Intl. Institute for Conservation of Historic and Artistic Works, pp. 128–132.
123. Mindermann S (2018). *Hyperspectral imaging for readability enhancement of historic manuscripts*, MA thesis, Technical University München, München.
124. Mitchell G (2012). “Revisiting Truth or Triviality: The External Validity of Research in the Psychological Laboratory”, *Perspectives on Psychological Science*, 7 (2): 109–117.
125. Mocella V, Brun E, Ferrero C, Delattre D (2015). “Revealing letters in rolled Herculaneum papyri by X-ray phase-contrast imaging”, *Nature Communications*, 6:5895.
126. Moriyama D, Azetsu T, Ueda Ch, Suetake N, Uchino E (2020). “Image enhancement with lightness correction and image sharpening based on characteristics of vision for elderly persons”, *Optical Review*, 27 (4): 352–360.
127. Morovič J (2008). *Color Gamut Mapping*, Chichester: Wiley.
128. Mullen KT (1985). “The contrast sensitivity of human colour vision red-green and blue-yellow chromatic gratings”, *The J. of Physiology*, 359: 381–400.
129. Mussell J (2012). *The Nineteenth-Century Press in the Digital Age*, London: Palgrave Macmillan.
130. Neji H, Nogueras-Iso J, Lacasta J, Ben Halima M, Alimi A (2019). “Adversarial Autoencoders for Denoising Digitized Historical Documents: The Use Case of Incunabula”, *Proc. Intl. Conf. on Document Analysis and Recognition (ICDAR 2019), 20–25 September 2019, Sydney, Australia*, pp. 31–34.
131. Nguyen CKh, Nguyen CT, Hotta S, Nakagawa M (2019). “A Character Attention Generative Adversarial Network for Degraded Historical Document Restoration”, *Proc. Intl. Conf. on Document Analysis and Recognition (ICDAR 2019), 20–25 September 2019, Sydney, Australia*, pp. 420–425.
132. Oppenheimer DM (2006). “Consequences of Erudite Vernacular Utilized Irrespective of Necessity: Problems with Using Long Words Needlessly”, *Applied Cognitive Psychology*, 20: 139–156.
133. Pan X-B, Brady M, Bowman AK, Crowther Ch, Tomlin RSO (2004). “Enhancement and feature extraction for images of incised and ink texts”, *Image and Vision Computing*, 22: 443–451.
134. Paris S, Hasinoff SW, Kautz J (2015). “Local Laplacian Filters: Edge-Aware Image Processing with a Laplacian Pyramid”, *Communications of the ACM*, 58 (3): 81–91.
135. Parry N. R. A. (2015). “Color vision deficiencies”, in Elliot A. J., Fairchild M. D., Franklin A., eds., *Handbook of Color Psychology*, Cambridge: Cambridge University Press, pp. 216–242.
136. Peli E (1992). “Limitations of Image Enhancement for the Visually Impaired”, *Optometry and Vision Science*, 15–24.
137. Peli E, Lee E, Trempe CL, Buzney Sh (1994). “Image enhancement for the visually impaired: the effects of enhancement on face recognition”, *J. of the Optical Society of America A*, 11 (7): 1929–1939.
138. Peli E, Woods RL (2009). “Image enhancement for impaired vision: the challenge of evaluation”, *Intl. J. on Artificial Intelligence Tools*, 18 (3): 415–438.
139. Perry S (2018). “Image and Video Noise: An Industry Perspective”, in Bertalmío M, ed., *Denoising of Photographic Images and Video: Fundamentals, Open Challenges and New Trends*, Cham: Springer, pp. 217–230.
140. Petro AB, Sbert C, Morel J-M (2014). “Multiscale Retinex”, *Image Processing On Line*, 4: 71–88.
141. Pham Th-A, Delalandre M (2017). “Post-processing coding artefacts for JPEG documents”, *Intl. J. on Document Analysis and Recognition*, 20 (3): 189–200.
142. Piquette KE (2018). “Revealing the Material World of Ancient Writing: Digital Techniques and Theoretical Considerations”, in Hoogendijk FA J, van Gompel SM T, eds., *The Materiality of Texts from Ancient Egypt: New Approaches to the Study of Textual Material from the Early Pharaonic to the Late Antique Period*, Leiden: Brill, pp. 94–118.
143. Plateau J (1839). *Mémoire sur l’irradiation*, Bruxelles: Hayez.
144. Ponomarenko N, Jin L, Ieremeiev O, Lukin V, Egiazarian K, Astola J, Vozel B, Chehdi K, Carli M, Battisti F, Kuo C-CJ (2015). “Image database TID2013: Peculiarities, results and perspectives”, *Image Communication*, 30 (1): 57–77.
145. Pons C, Mazade R, Jin J, Dul MW, Zaidi Q, Alonso J–M (2017). “Neuronal mechanisms underlying differences in spatial resolution between darks and lights in human vision”, *J. of Vision*, 17 (14): 5, 1–24.
146. Popowicz A, Smolka B (2015). “Overview of Grayscale Image Colorization Techniques”, in Celebi ME, Lecca M, Smolka B, eds., *Color Image and Video Enhancement*, Cham: Springer, pp. 345–370.
147. Da Pos O, Zambianchi E (1996). *Visual Illusions and Effects: A Collection*, Milano: Guerini.
148. Poston AM (1974). *A literature review of cockpit lighting*, technical memorandum 10-64, Aberdeen Proving Ground, MD: US Army Human Engineering Laboratory.
149. Pratikakis I, Zagoris K, Karagiannis X, Tsochatzidis L, Mondal T, Marthot-Santaniello I (2019). “ICDAR 2019 Competition on Document Image Binarization (DIBCO 2019)”, *Proc. 2019 Intl. Conf. on Document Analysis and Recognition (ICDAR), 20–25 September 2019, Sydney, Australia*, pp. 1547–1556.
150. Raj V, Arunkumar C (2014). “Content Restoration of Degraded Termite Bitten Document Images”, *Intl. J. of Engineering Research and Applications*, 4 (5): 151–155.
151. Rasche K, Geist R, Westall J (2005). “Detail Preserving Reproduction of Color Images for Monochromats and Dichromats”, *IEEE Computer Graphics and Applications*, 26 (3): 22–30.
152. Ribeiro F, Florencio D., Nascimento V (2011). “Crowdsourcing Subjective Image Quality Evaluation”, *Proc. of 18th IEEE Intl. Conf. on Image Processing (ICIP 2011), 11–14 September 2011, Brussels, Belgium*, pp. 3097–3100.
153. Rodney A (2005). *Color Management for Photographers*, Oxford: Focal Point.
154. Rogowitz BE, Treinish LA (1998). “Data visualization: The end of the rainbow”, *IEEE Spectrum*, 12: 52–59.
155. Roued-Cunliffe H (2011). *A decision support system for the reading of ancient documents*, Ph.D. thesis, Faculty of Classics, Univ. of Oxford, Oxford.

156. Rudd ME (2017). "Lightness computation by the human visual system", *J. of Electronic Imaging*, 26 (3): 031209.
157. Rudd ME, Popa D (2007). "Stevens's brightness law, contrast gain control, and edge integration in achromatic color perception: a unified model", *J. Optical Soc. of America*, A/24 (9): 2766–2782.
158. Samei E, Krupinski EA, eds. (2019). *The Handbook of Medical Image Perception and Techniques*, Cambridge: Cambridge Univ. Press, 2nd ed.
159. Sassoon R (2007). *Handwriting of the Twentieth Century*, Bristol: Intellect.
160. Schivre G (2020). "Multiscale Retinex", *MATLAB Central File Exchange*, https://www.mathworks.com/matlabcentral/fileexchange/71386-multiscale-retinex.
161. Schubert P (2009). "Editing a Papyrus", in Bagnall RS, ed., *The Oxford Handbook of Papyrology*, Oxford: Oxford Univ. Press, pp. 197–215.
162. Schwartz-Shea P, Yanow D (2012). *Interpretive Research Design: Concepts and Processes*, New York, NY: Routledge.
163. Seuret M, Chen K, Eichenberger N, Liwicki M, Ingold R (2015). "Gradient-domain degradations for improving historical documents images layout analysis", *Proc. 13th Intl. Conf. on Document Analysis and Recognition (ICDAR), 23–26 August 2015, Nancy, France*, pp. 1006–1010.
164. Shapiro A, Todorović D, eds. (2017). *The Oxford Compendium of Visual Illusions*, Oxford: Oxford Univ. Press.
165. Sharma A (2018). *Understanding Color Management*, Hoboken, NJ: Wiley, 2nd ed.
166. Sharma G, Rodríguez-Pardo CE (2012). "The Dark Side of CIELAB", *Proc. SPIE–IS&T Electronic Imaging, Color Imaging XVII: Displaying, Processing, Hardcopy, and Applications*, vol. 8292: 82920D.
167. Shneiderman B (1996). "The Eyes Have It: A Task by Data Type Taxonomy for Information Visualizations", *Proc. 1996 IEEE Symp. on Visual Languages*, 3–6 September 1996, Boulder, CO, USA, pp. 336–343.
168. Shortland ND, Alison LJ, Moran JM (2019). *Conflict: How Soldiers Make Impossible Decisions*, Oxford: Oxford University Press.
169. Simons DJ, Chabris CF (1999). "Gorillas in our midst: Sustained inattentional blindness for dynamic events", *Perception*, 28 (9): 1059–1074.
170. Smith AR (1978). "Color gamut transform pairs", *SIGGRAPH '78: Proc. 5th Annual Conf. on Computer Graphics and Interactive Techniques, 23–25 August 1978, Atlanta, GA, USA*, pp. 12–19.
171. Sider D (2005). *The Library of the Villa Dei Papiri at Herculaneum*, Los Angelse, CA: Getty Publications.
172. Sobottka K, Kronenberg H, Perroud T, Bunke H (2000). "Text extraction from colored book and journal covers", *Intl. J. on Document Analysis and Recognition*, 2 (4): 163–176.
173. Sonnad S (2016). "A survey on fusion of multispectral and panchromatic images for high spatial and spectral information", *Proc. Intl. Conf. on Wireless Communications, Signal Processing and Networking (WiSPNET), 23–25 March 2016, Chennai, India*, pp. 177–180.
174. Sparavigna AC (2009). "Digital Restoration of Ancient Papyri", *arXiv.org*, http://arxiv.org/abs/0903.5045.
175. Tarte S (2011). "Papyrological investigations: transferring perception and interpretation into the digital world", *Literary and Linguistic Computing*, 26 (2): 233–247.
176. Terras MM (2006). *Image to Interpretation: An Intelligent System to Aid Historians in Reading the Vindolanda Texts*, Oxford: Oxford Univ. Press.
177. Tian Ch, Fei L, Zheng W, Xu Y, Zuo W, Lin Ch-W (2020). "Deep learning on image denoising: An overview", *Neural Networks*, 131 (11): 251–275.
178. Tonazzini A (2010). "Color Space Transformations for Analysis and Enhancement of Ancient Degraded Manuscripts", *Pattern Recognition and Image Analysis*, 20 (3): 404–417.
179. Tonazzini A, Gerace I, Martinelli F (2012). "Document Image Restoration and Analysis as Separation of Mixtures of Patterns: From Linear to Nonlinear Models", in Gunturk BK, Li X, eds., *Image Restoration: Fundamentals and Advances*, Boca Raton, FL: CRC Press, pp. 285–310.
180. Tournié A, Fleischer K, Bukreeva I, Palermo F, Perino M, Cedola A, Andraud C, Ranocchia G (2019). "Ancient Greek text concealed on the back of unrolled papyrus revealed through shortwave-infrared hyperspectral imaging", *Science Advances*, 5 (10) eaav8936.
181. Van Minnen P (2009). "The Future of Papyrology", in Bagnall RS, ed., *The Oxford Handbook of Papyrology*, Oxford: Oxford Univ. Press, pp. 644–660.
182. VanderPlas S, Hofmann H (2015). "Signs of the Sine Illusion — Why We Need to Care", *J. of Computational and Graphical Statistics*, 24 (4): 1170–1190.
183. Velázquez A, Levachkine S (2004). "Text/Graphics Separation and Recognition in Raster-Scanned Color Cartographic Maps", *Proc. Intl. Workshop on Graphics Recognition, GREC 2003: Graphics Recognition. Recent Advances and Perspectives, Lecture Notes in Computer Science*, vol 3088, pp. 63–74.
184. Ventzas D, Ntogas N, Ventza MM (2012). "Digital Restoration by Denoising and Binarization of Historical Manuscripts Images", in Ventzas D, ed., *Advanced Image Acquisition, Processing Techniques and Applications*, Rijeka: InTech, pp. 73–108.
185. Viénot F, Brettel H, Mollon JD (1999). "Digital Video Colourmaps for Checking the Legibility of Displays by Dichromats", *COLOR research and application*, 24 (4): 243–252.
186. Vilaseca M, Pujol J, Arjona M, Martínez-Verdú FM (2005). "Color Visualization System for Near-Infrared Multispectral Images", *J. of Imaging Science and Technology*, 49 (3): 246–255.
187. Wang Zh, Bovik AC (2006). *Modern Image Quality Assessment*, San Rafael, CA: Morgan & Claypool.
188. Wang Z, Simoncelli EP (2005). "Reduced-reference image quality assessment using a wavelet-domain natural image statistic model", *Proc. SPIE–IS&T Electronic Imaging Conf., Human Vision and Electronic Imaging X, 18 March 2005, San Jose, CA, USA*, pp. 149–159.
189. Webster MA (2015). "Individual differences in color vision", in Elliot AJ, Fairchild MD, Franklin A, eds., *Handbook of Color Psychology*, Cambridge: Cambridge Univ. Press, pp. 197–215.
190. Westheimer G (2007). "Irradiation, border location and the shifted-chessboard pattern", *Perception*, 36 (4): 483–494.
191. Wu HR, Rao KR, eds. (2006). *Digital Video Image Quality and Perceptual Coding*, Boca Raton, FL: CRC Press.
192. Yemelyanov KM, Lin ShSch, Luis WQ, Pugh EN Jr, Engheta N (2003). "Bio-inspired display of polarization information using selected visual cues", *Proc. SPIE Optical Science and Technology, 5158, Polarization Science and Remote Sensing, 3–8 August 2003, San Diego, California, USA*, pp. 71–84.
193. Yu FTS, Tai A, Chen H (1978). "Spatial filtered pseudocolor holographic imaging", *J. of Optics*, 9 (5): 269–273.
194. Zheng Y (2012). "An overview of night vision colorization techniques using multispectral images: From color fusion to color mapping", *Intl. Conf. on Audio, Language and Image Processing, 16-18 July 2012, Shanghai, China*, pp. 134–143.
195. Zhu Zh, Toyoura M, Go K, Kashiwagi K, Fujishiro I, Wong TT, Mao X (2021), "Personalized Image Recoloring for Color Vision Deficiency Compensation", *IEEE Transactions on Multimedia*, in press.
196. Ziyaee T (2014). "Unsupervised Denoising via Wiener Entropy Masking in the STFT Domain", *Proc. 2014 IEEE Military Communications Conf., 6–8 October 2014, Baltimore, MD, USA*, pp. 467–472.
197. Zuiderveld K (1994). "Contrast Limited Adaptive Histograph Equalization", *Graphic Gems IV*, San Diego, CA: Academic Press Professional, pp. 474–485.

## Supplementary Material

Presented herein are the ninety images used in the evaluation of papyri legibility enhancement experiment. The methods are described in the article, Sections 4 Methods and 5.1 Algorithms.

TSVP →

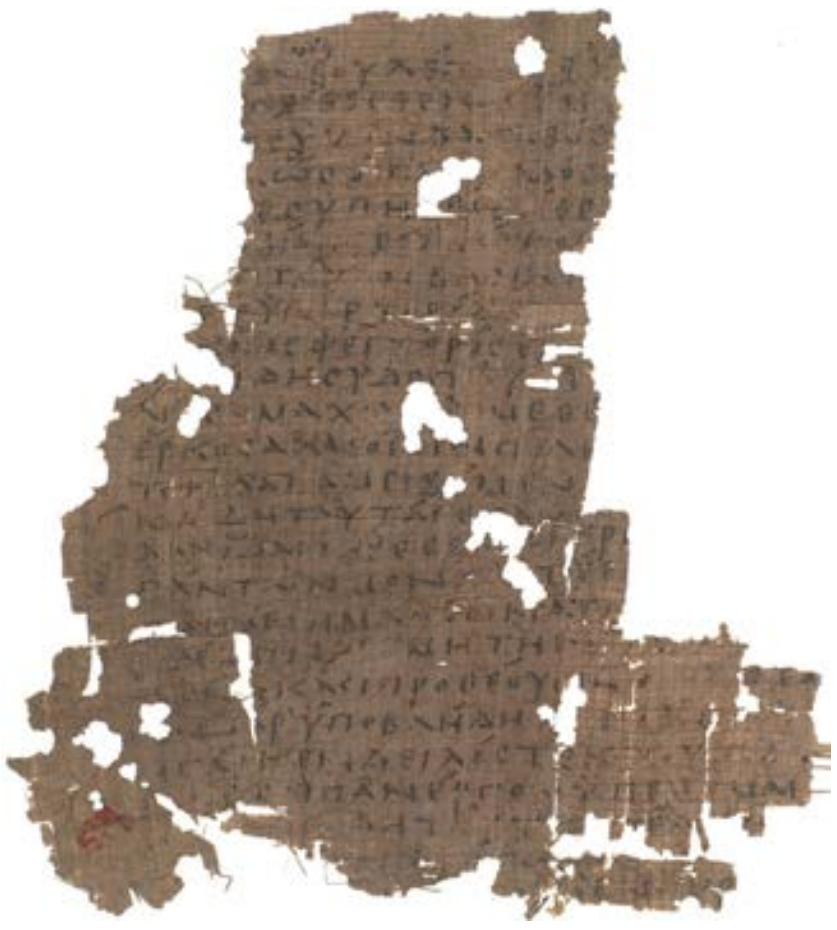

**original**

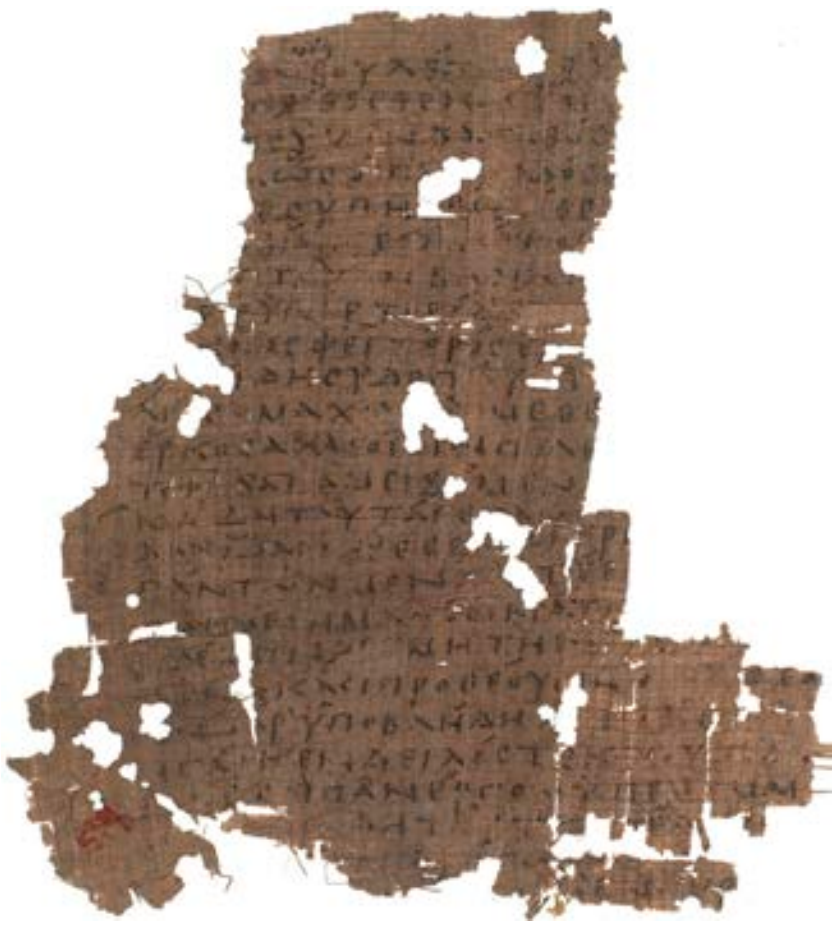

stretchlim

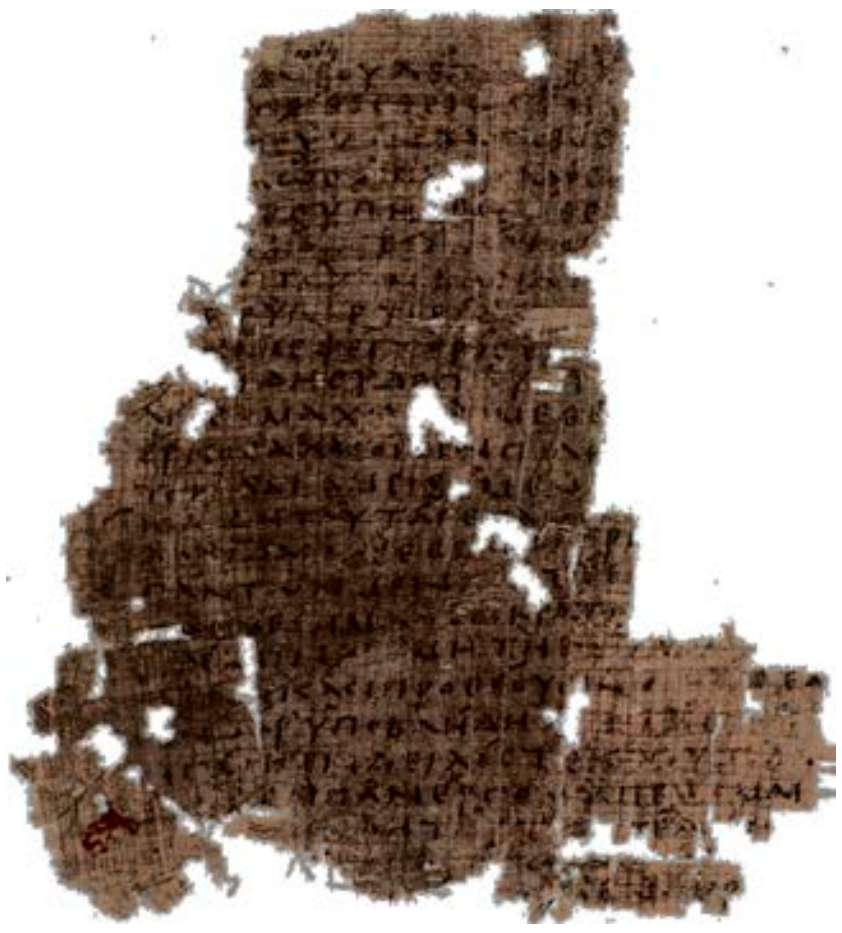

histeq

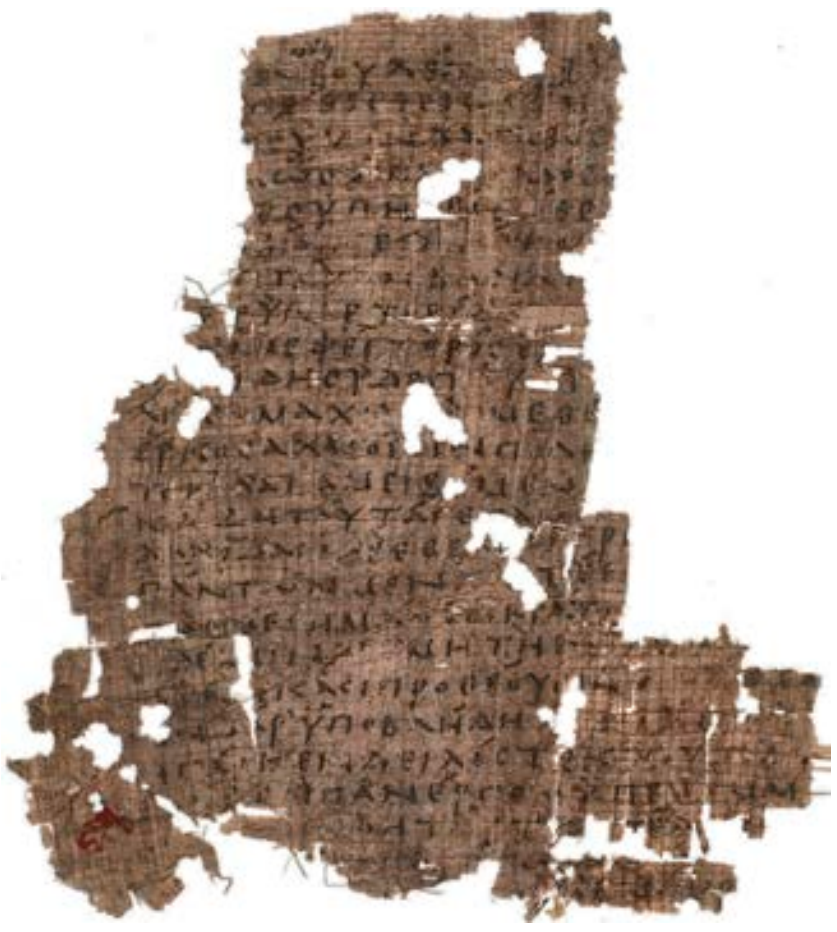

adapthisteq

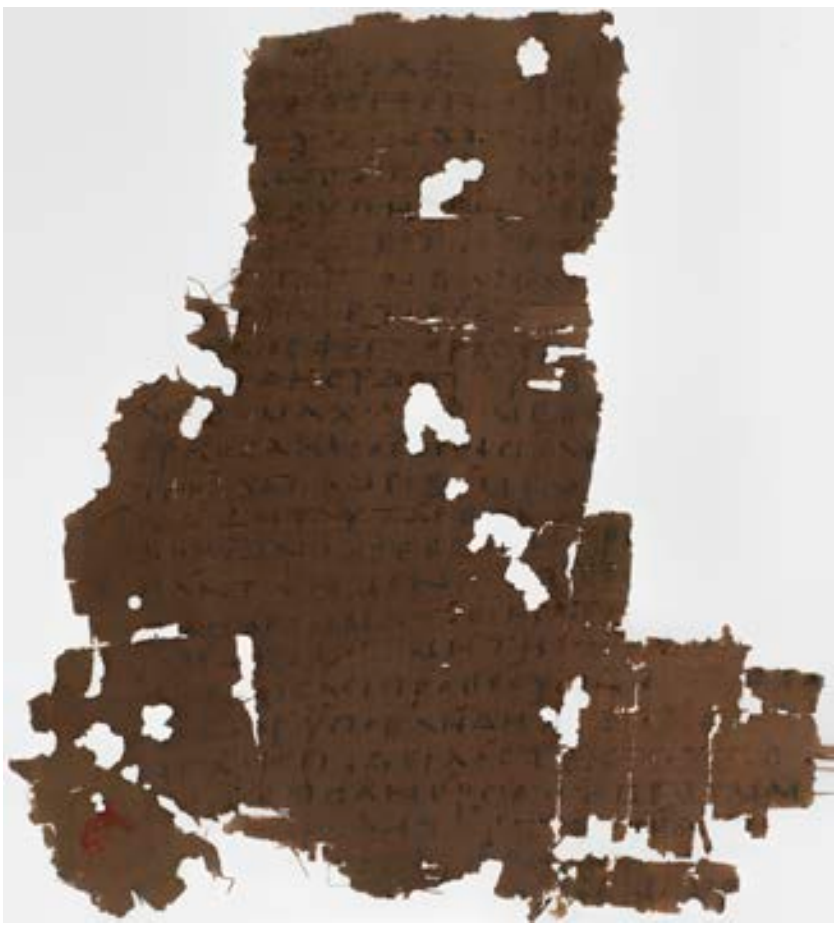

locallapfilt

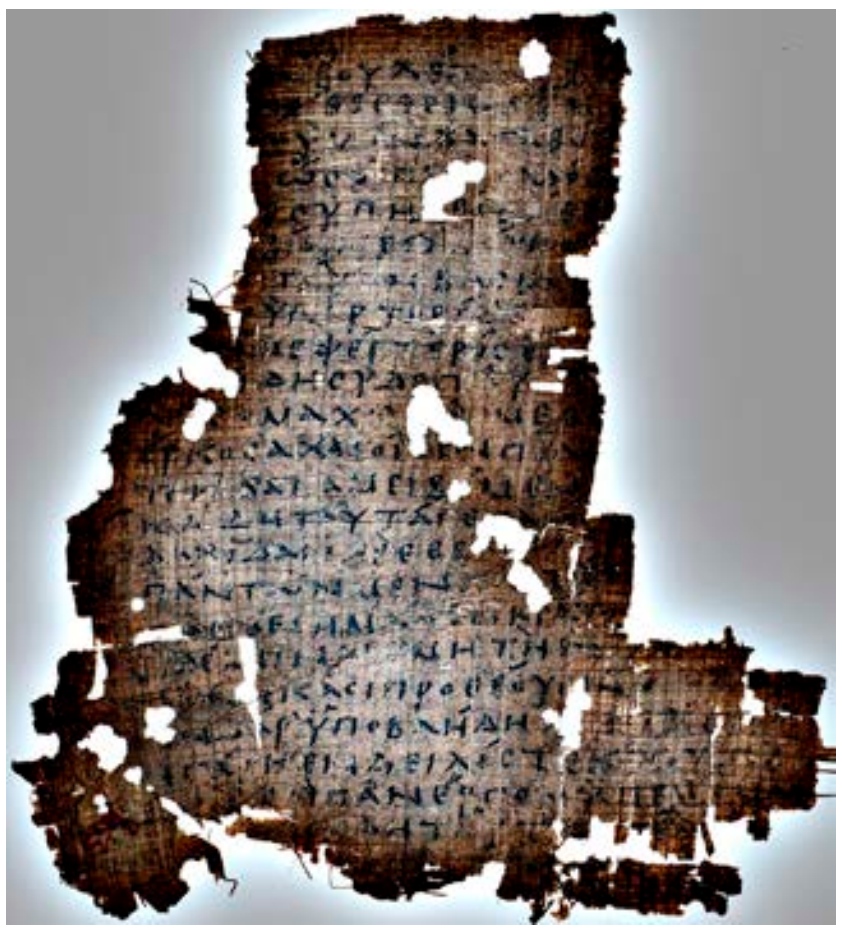

retinex

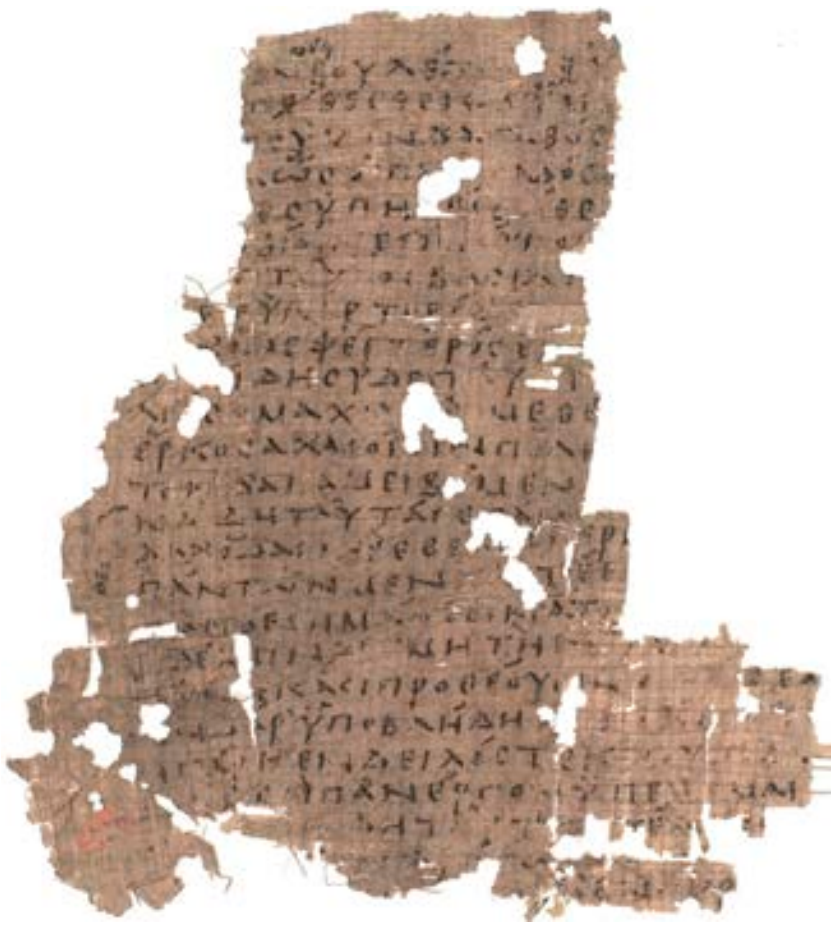

*lsv*

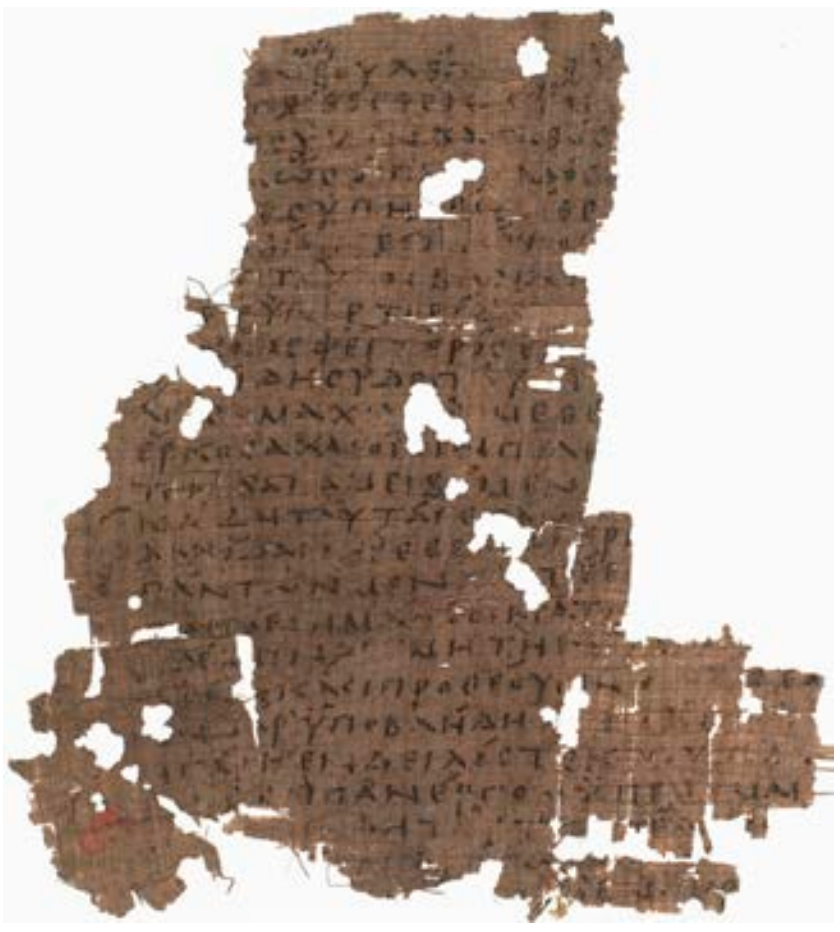

*vividness*

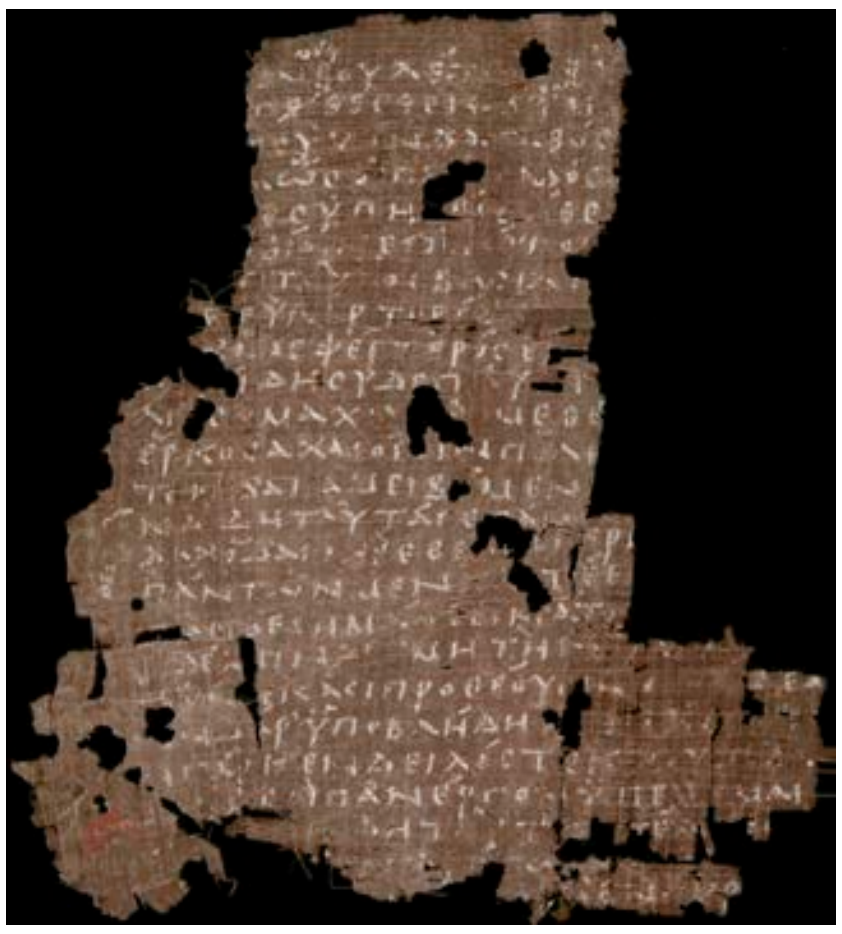

*neglsv*

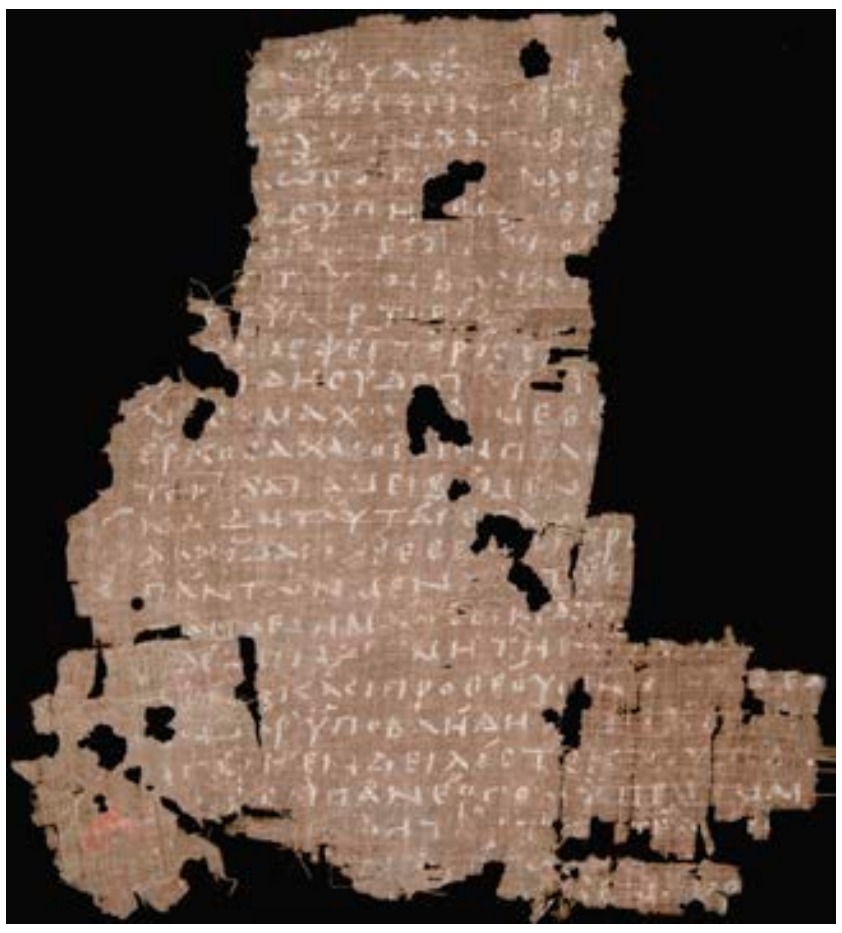

*negvividness*

**columbia.apis.p367.f.0.600**

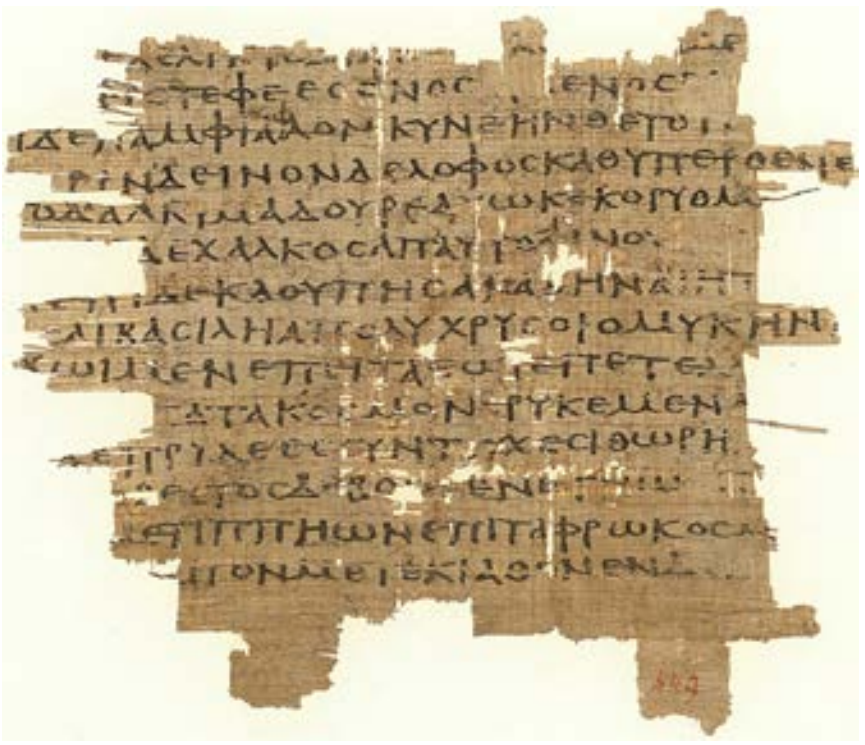

**original**

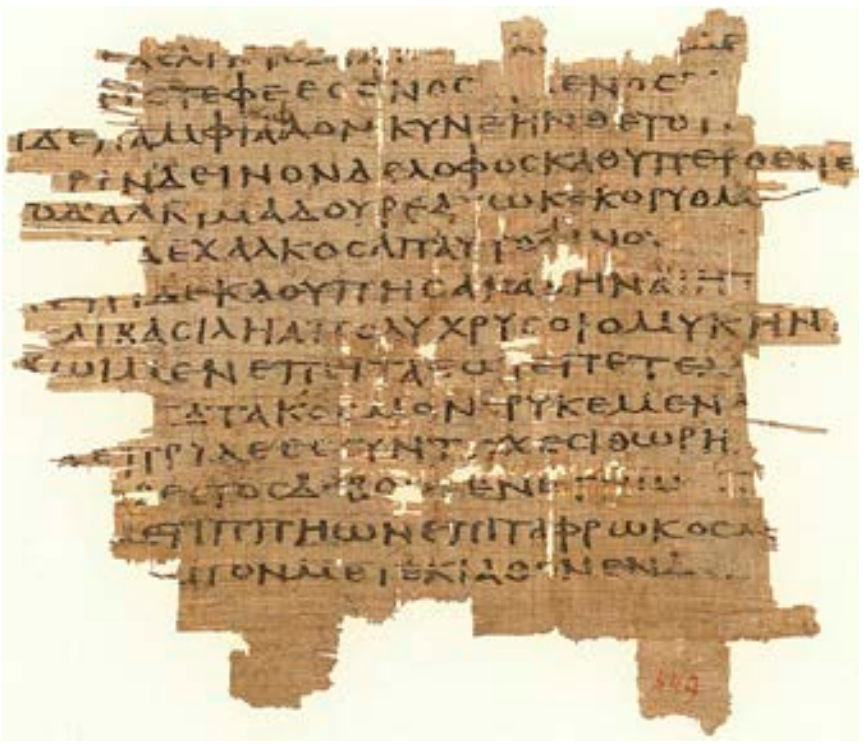

stretchlim

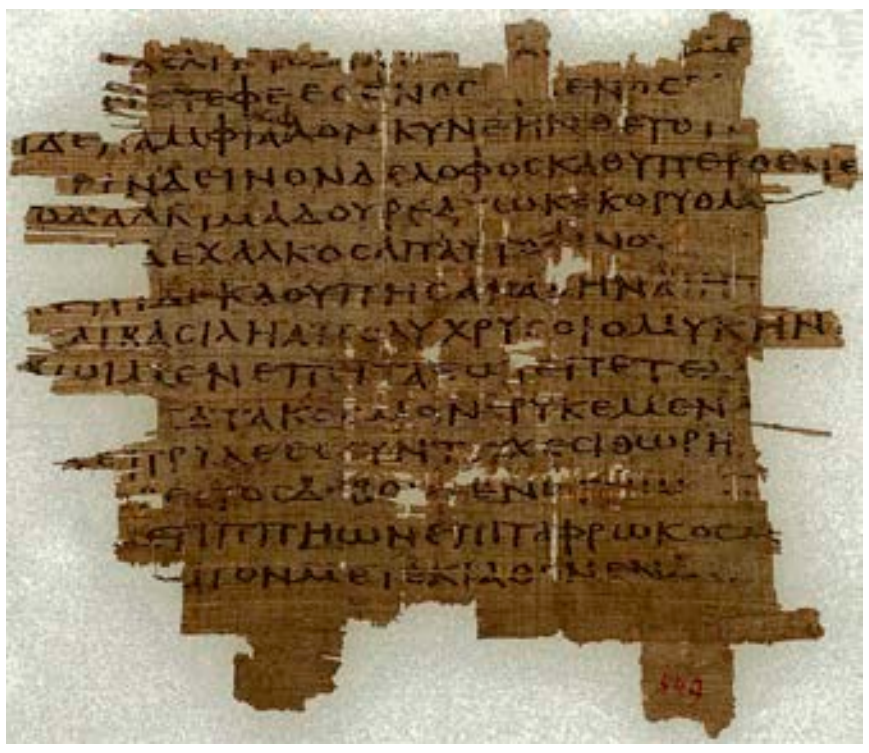

histeq

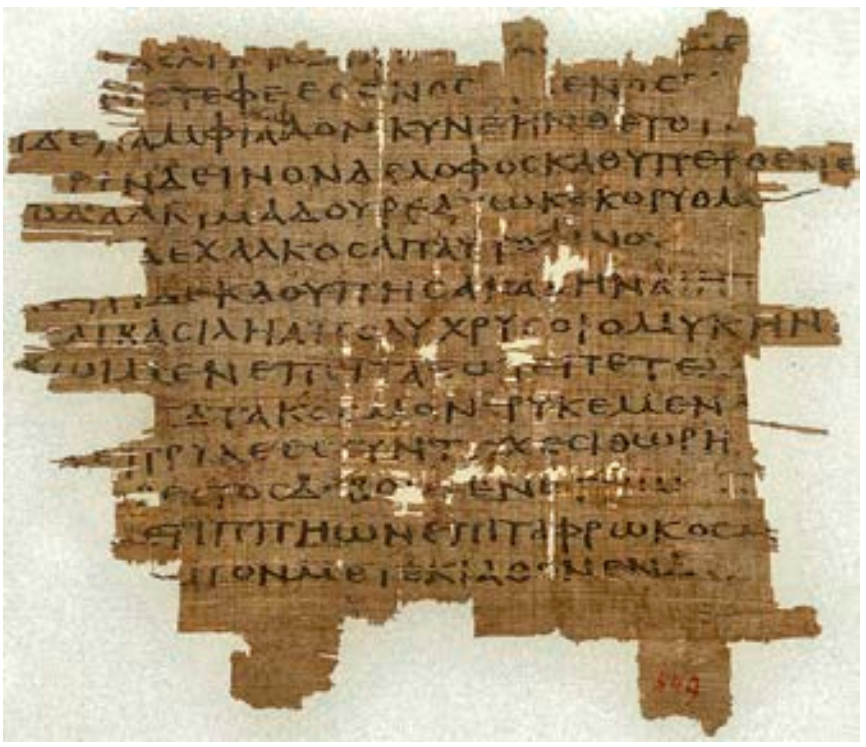

adapthisteq

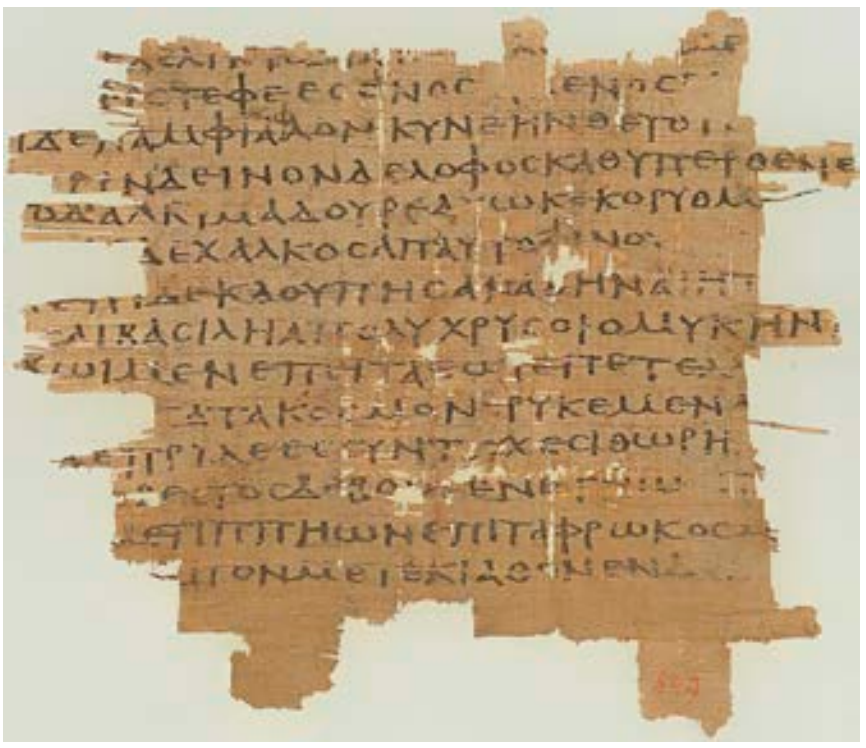

locallapfilt

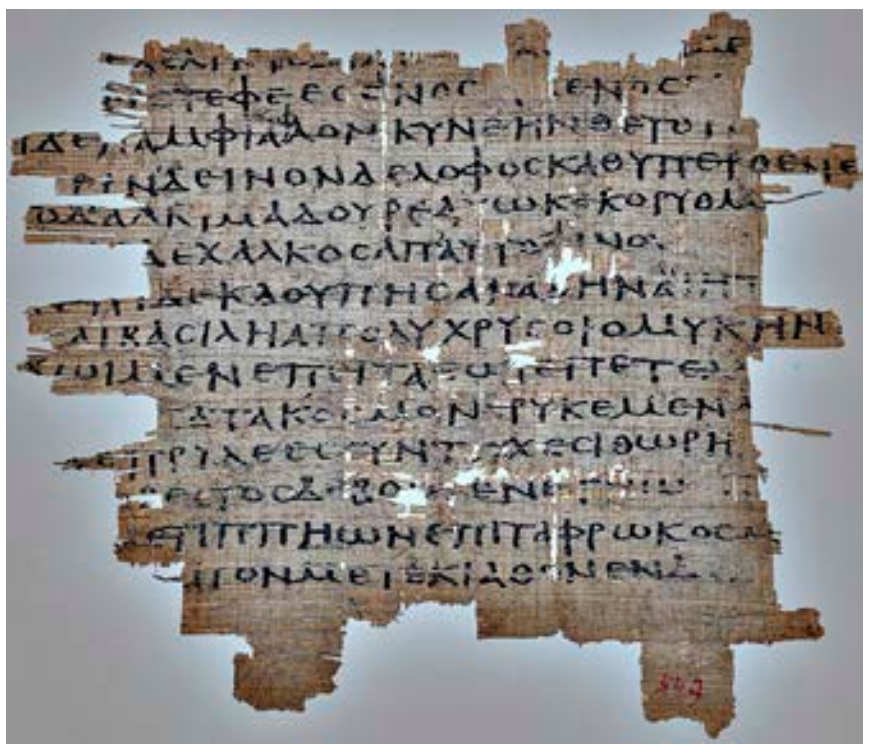

retinex

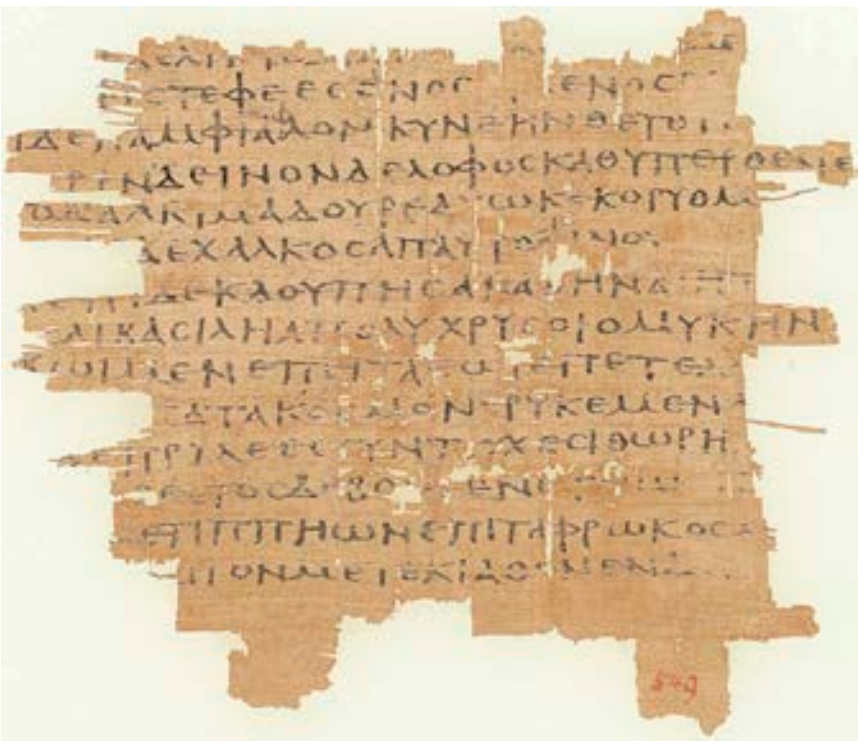

*lsv*

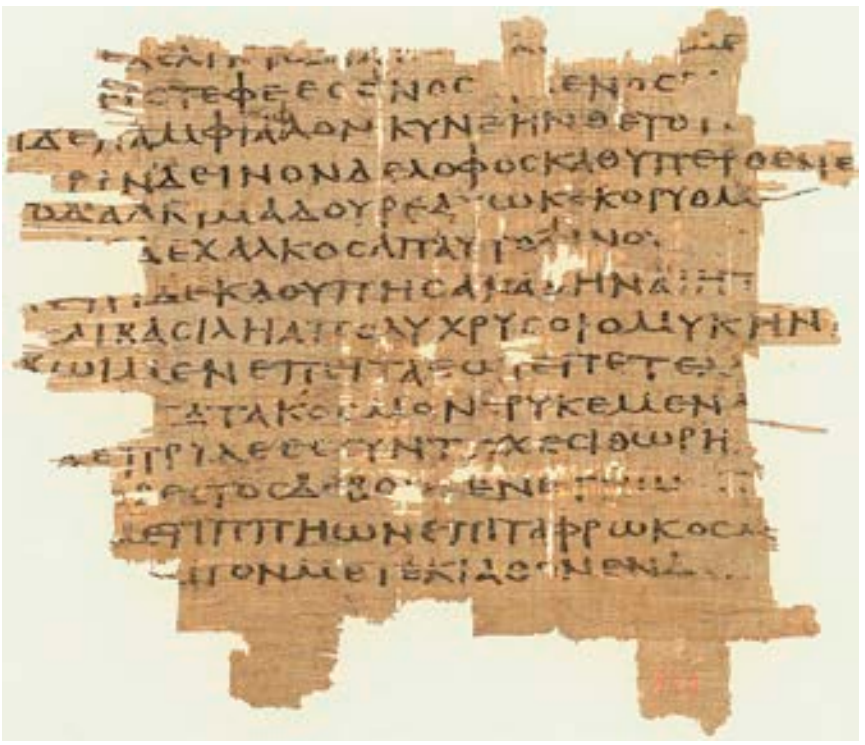

*vividness*

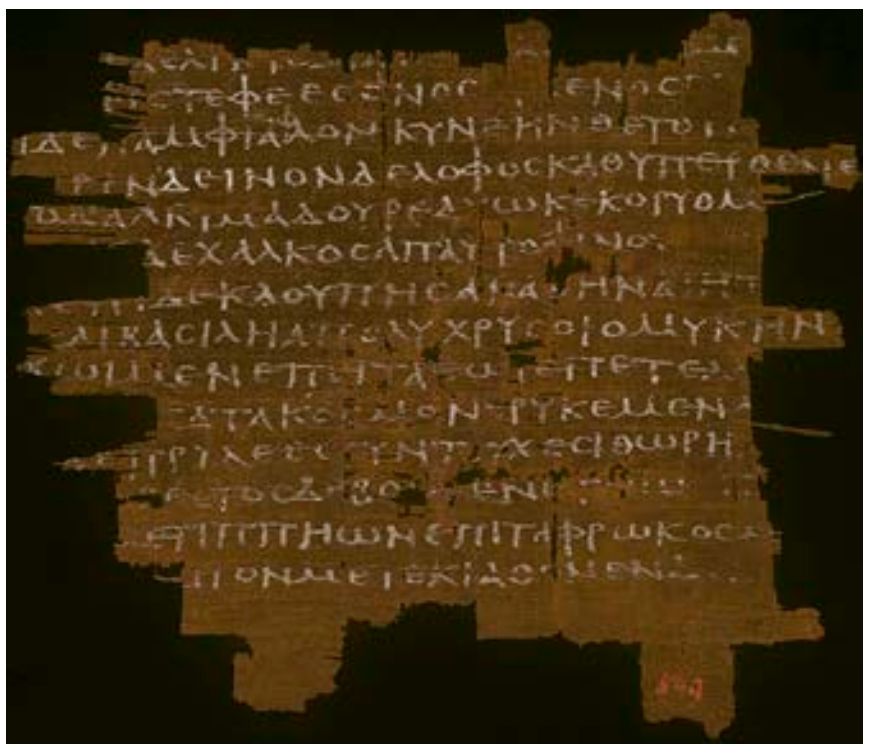

*neglsv*

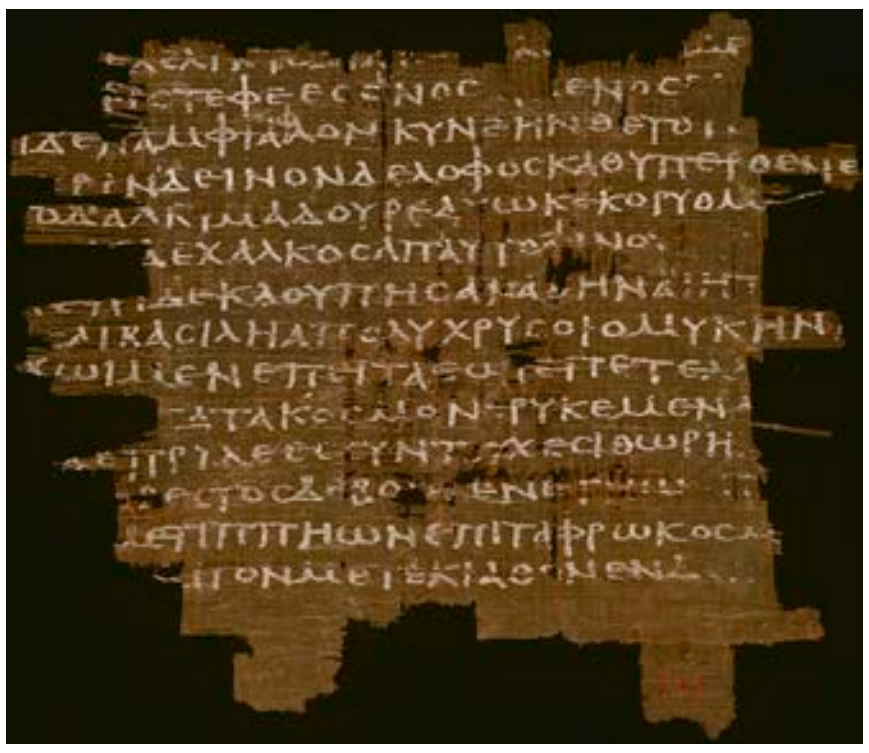

*negvividness*

**P.Corn. Inv. MSS.A 101. XIII**

Credit of original papyrus reproduction: University of Michigan Library, Public Domain Mark 1.0

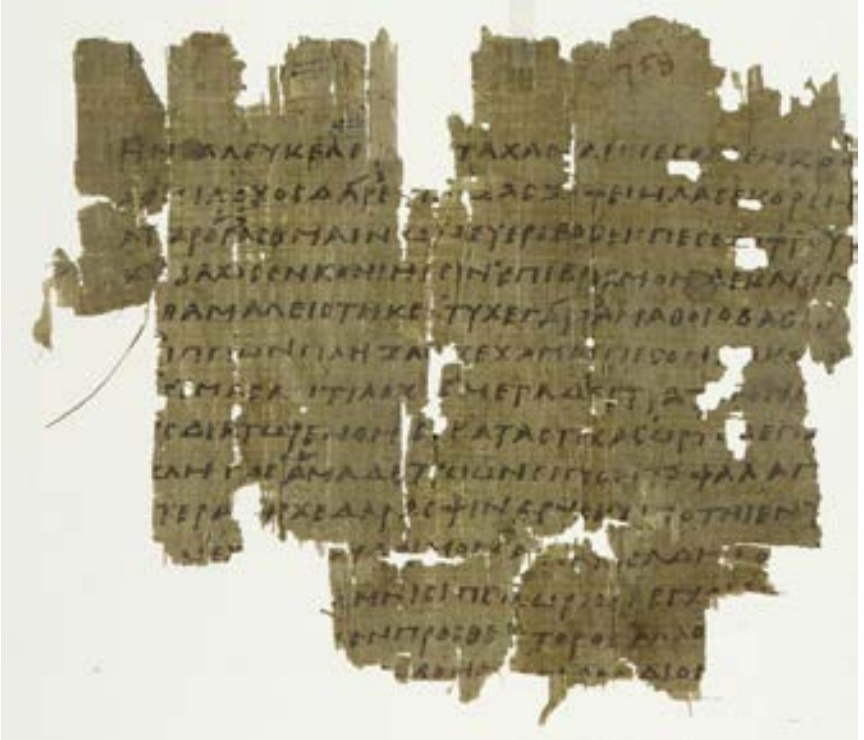

**original**

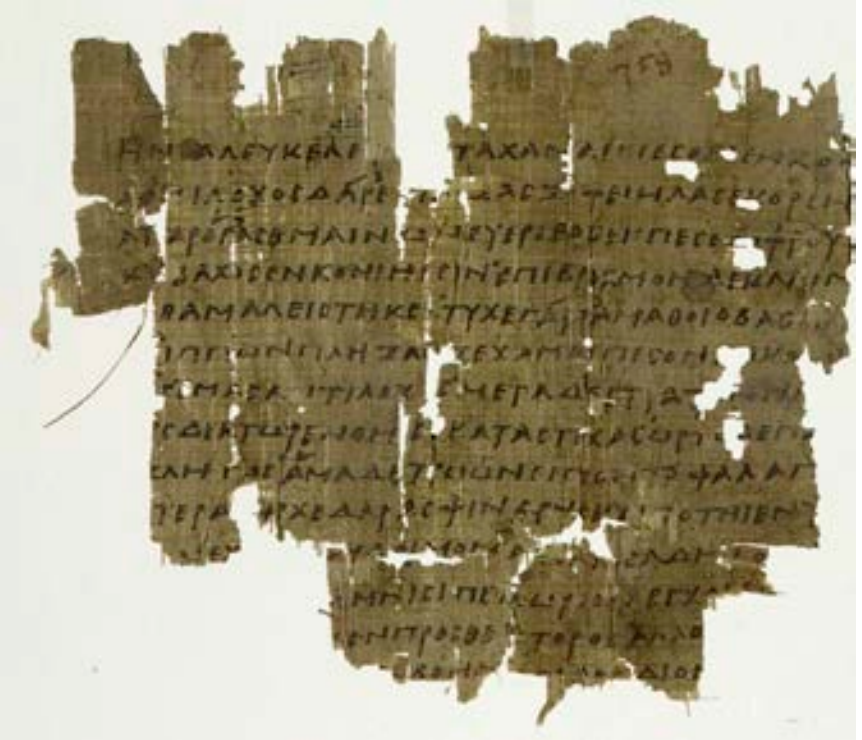

stretchlim

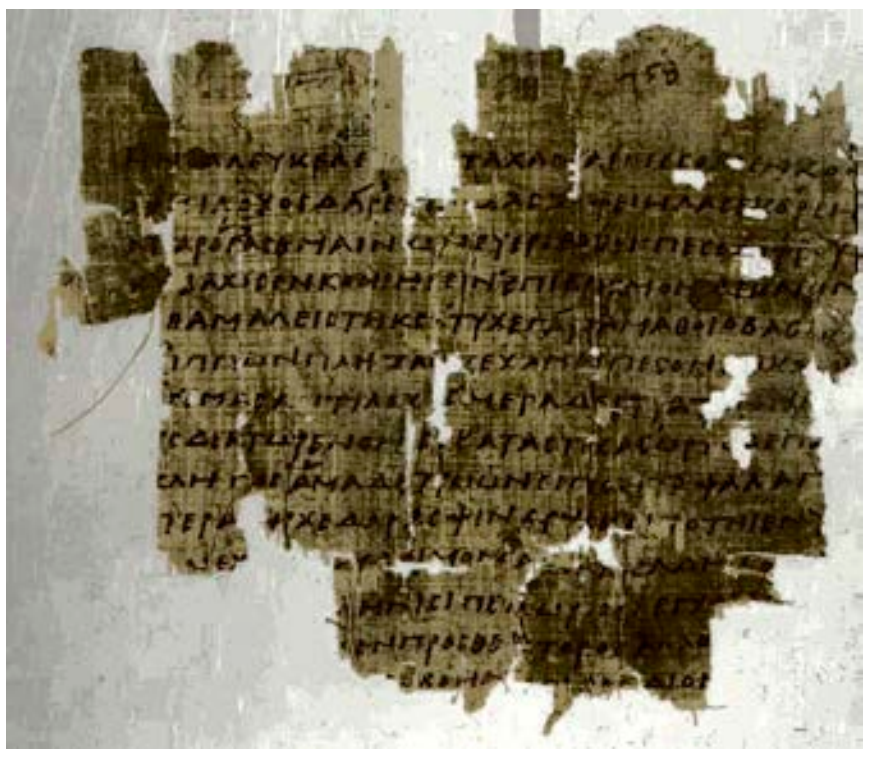

histeq

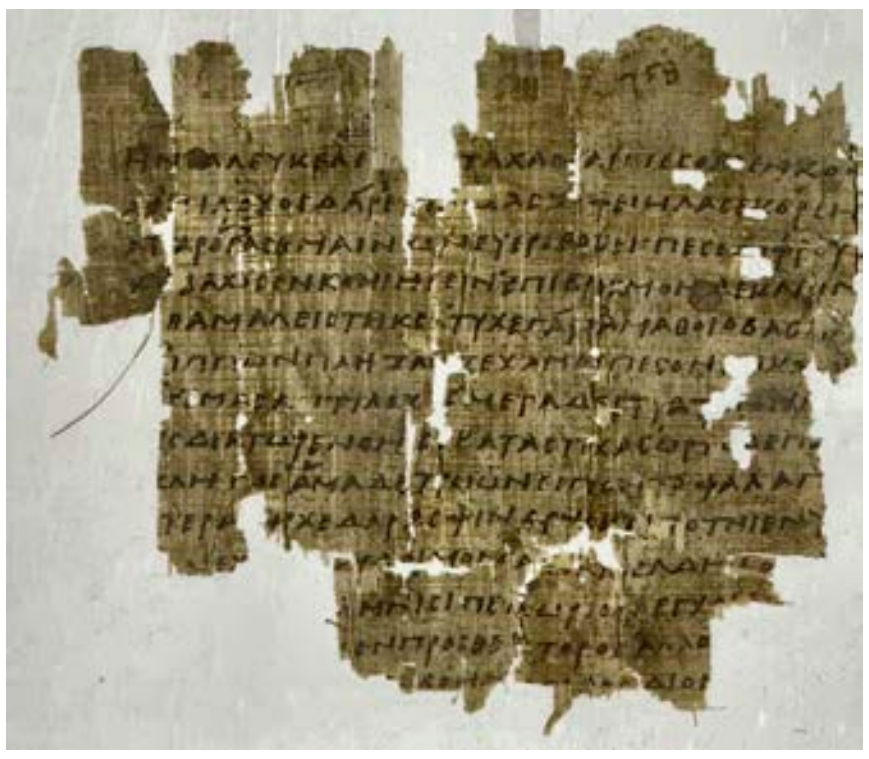

adapthisteq

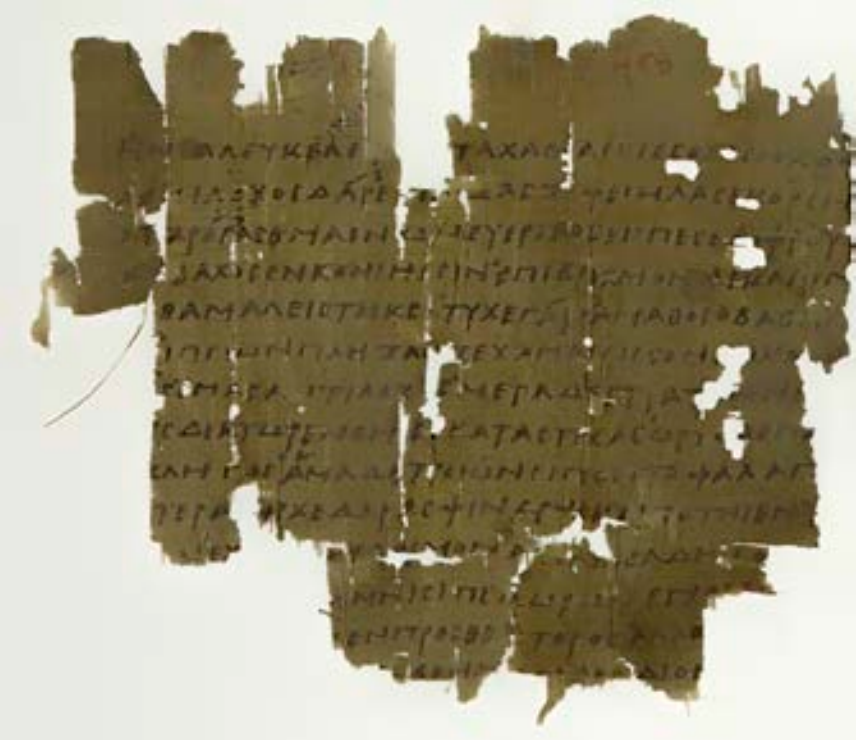

locallapfilt

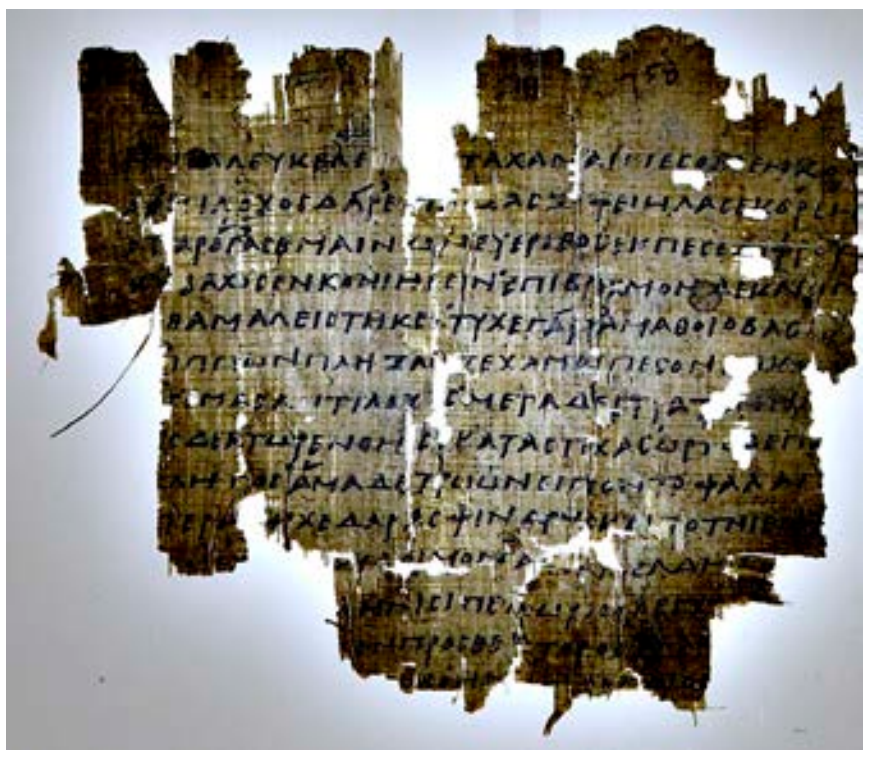

retinex

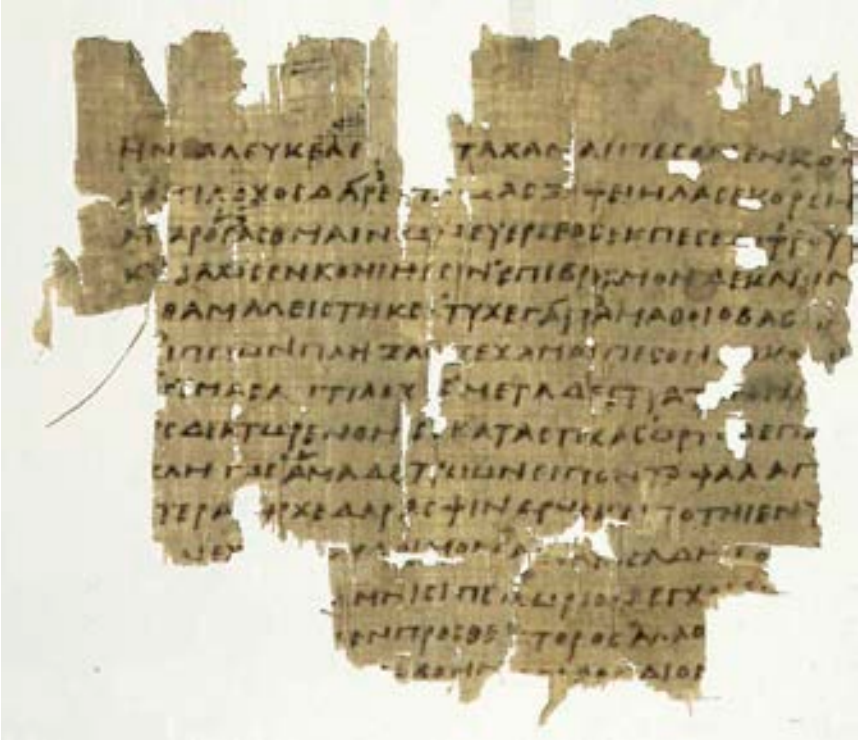

*lsv*

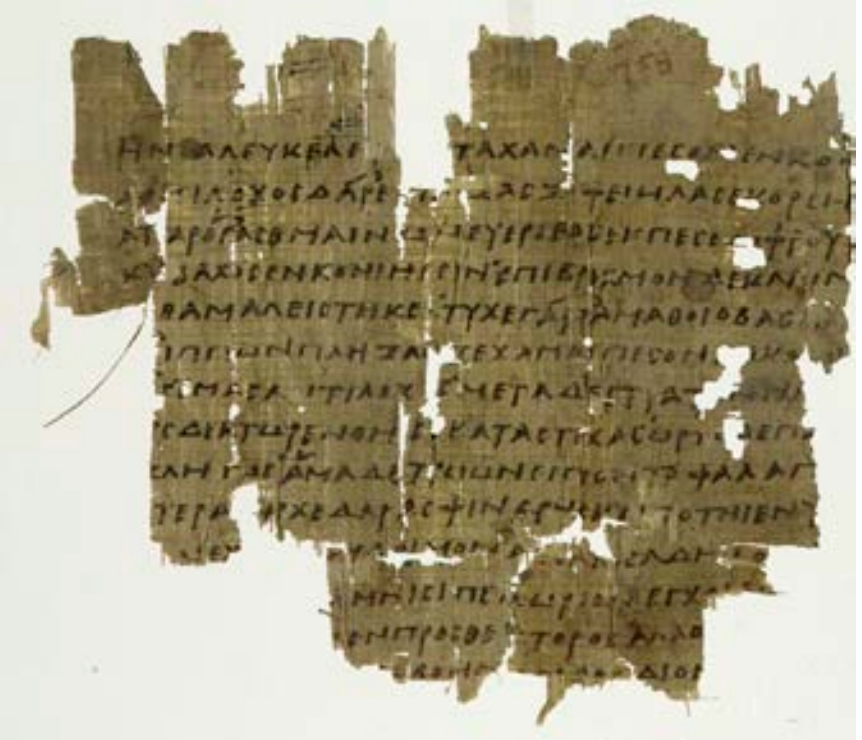

*vividness*

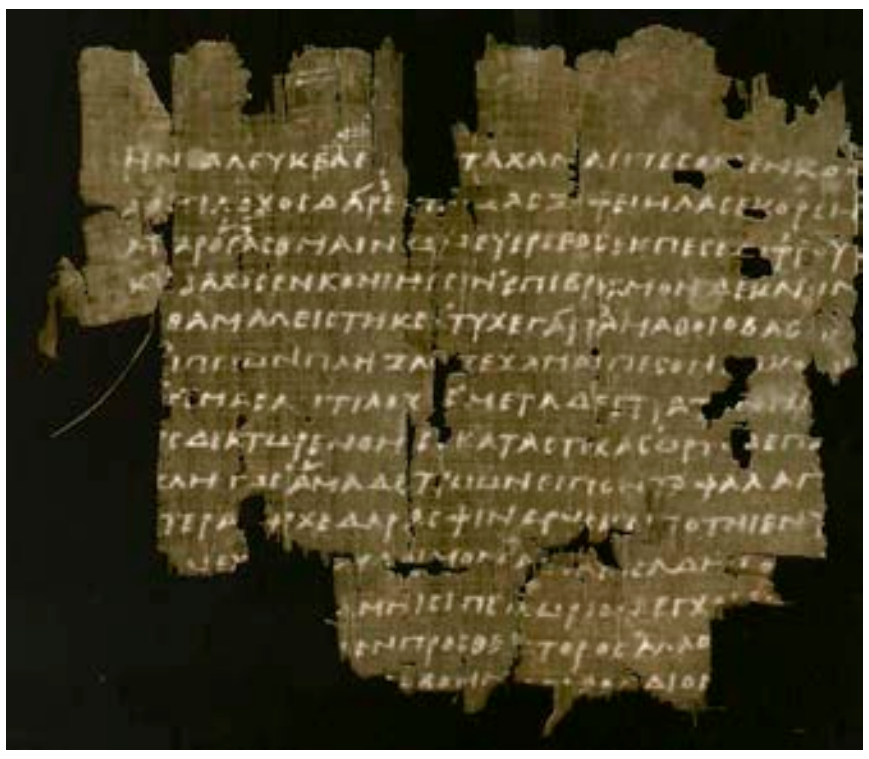

*neglsv*

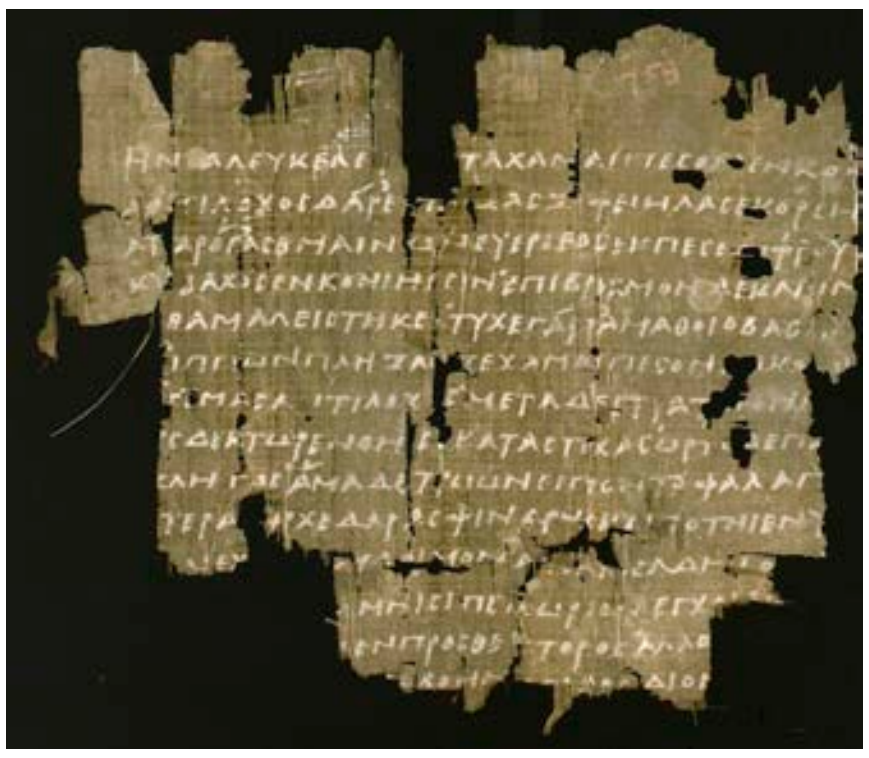

*negvividness*

**P.CtYBR inv. 69**

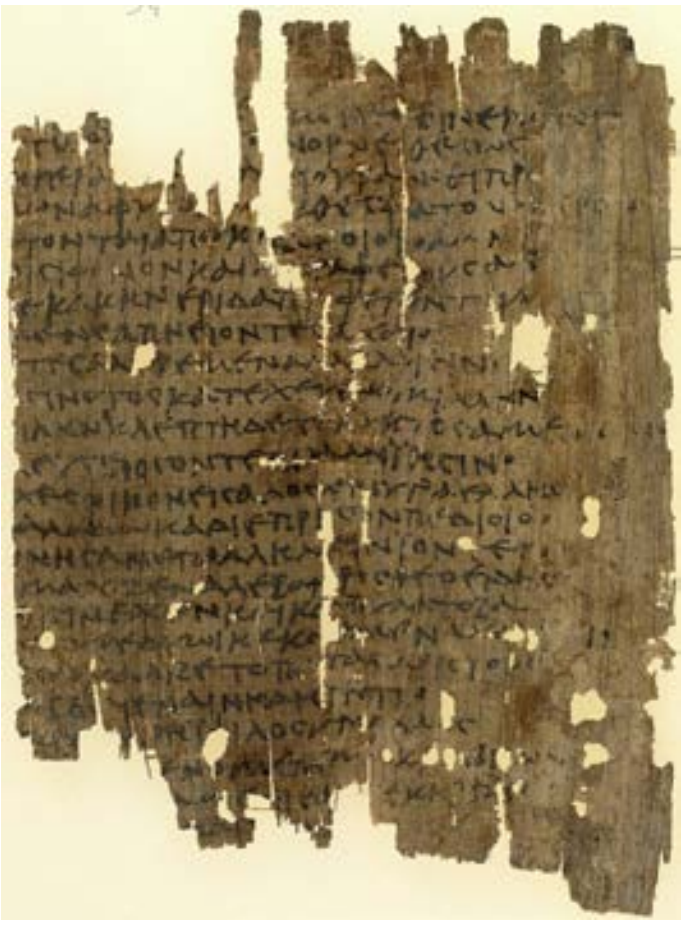

**original**

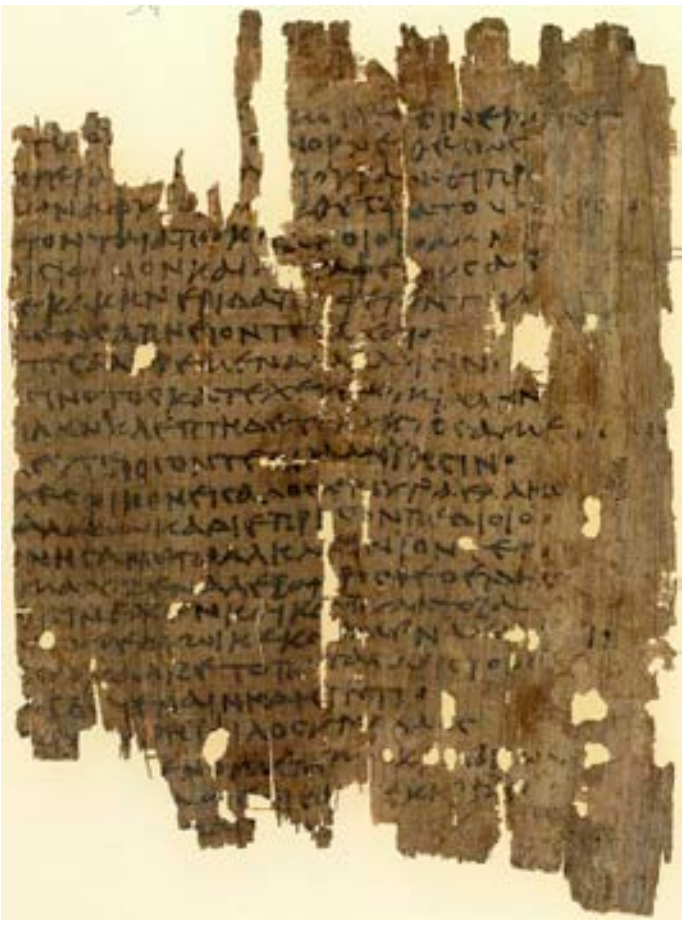

stretchlim

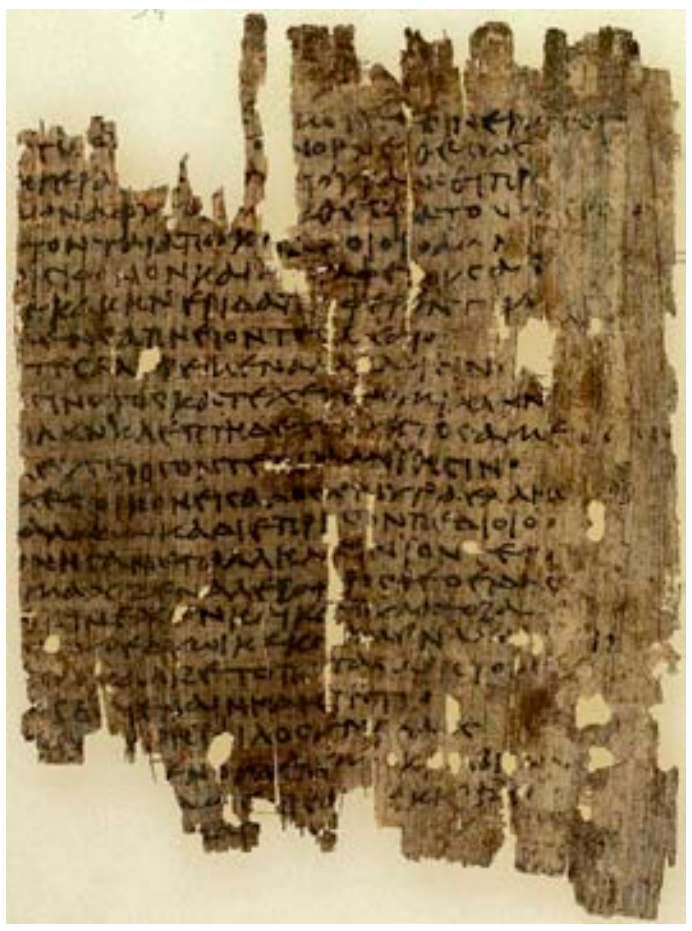

histeq

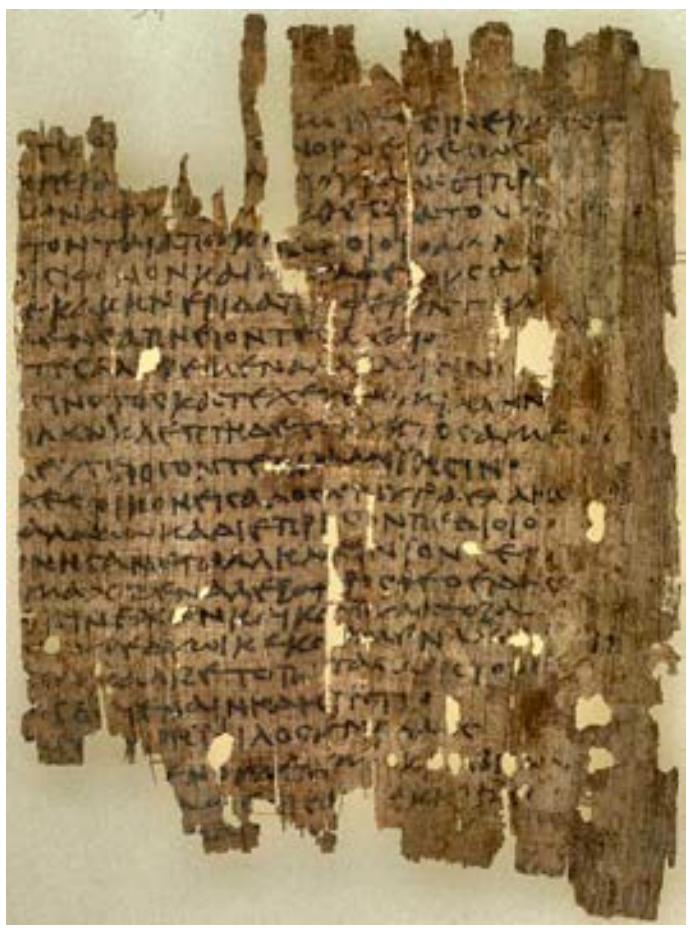

adapthisteq

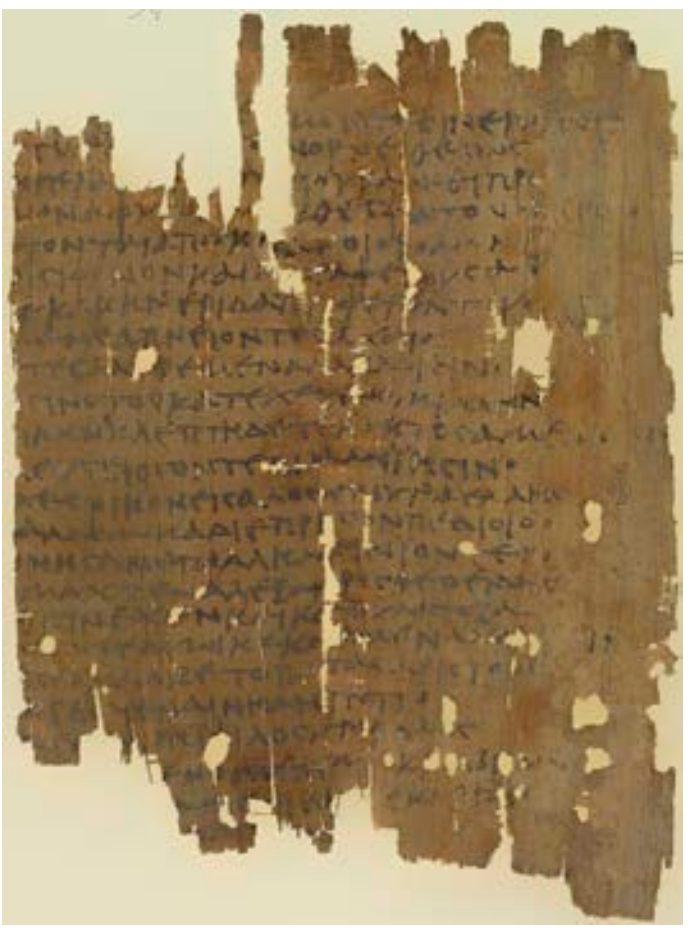

locallapfilt

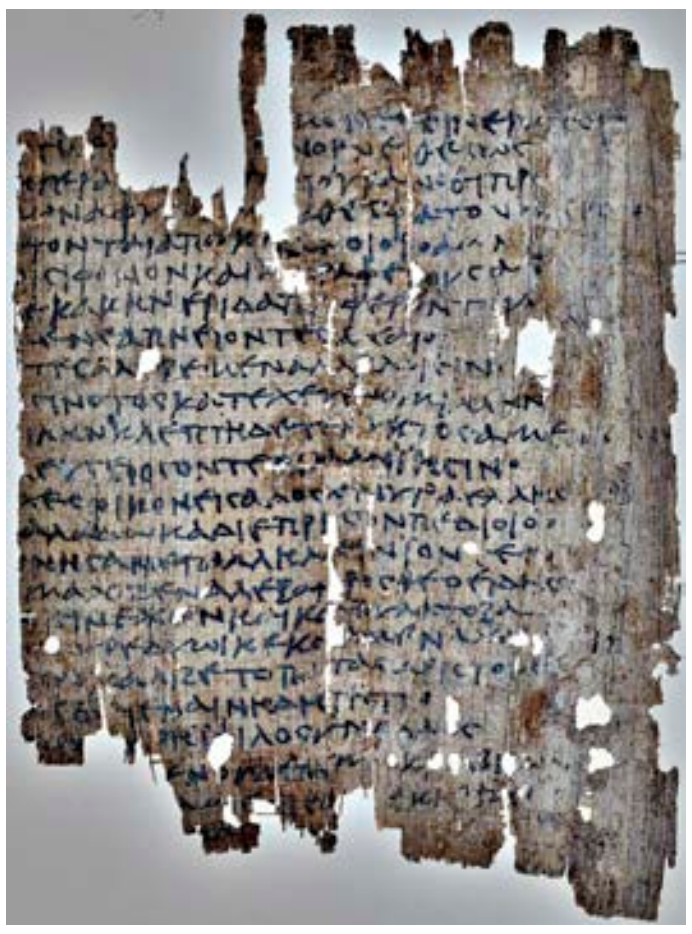

retinex

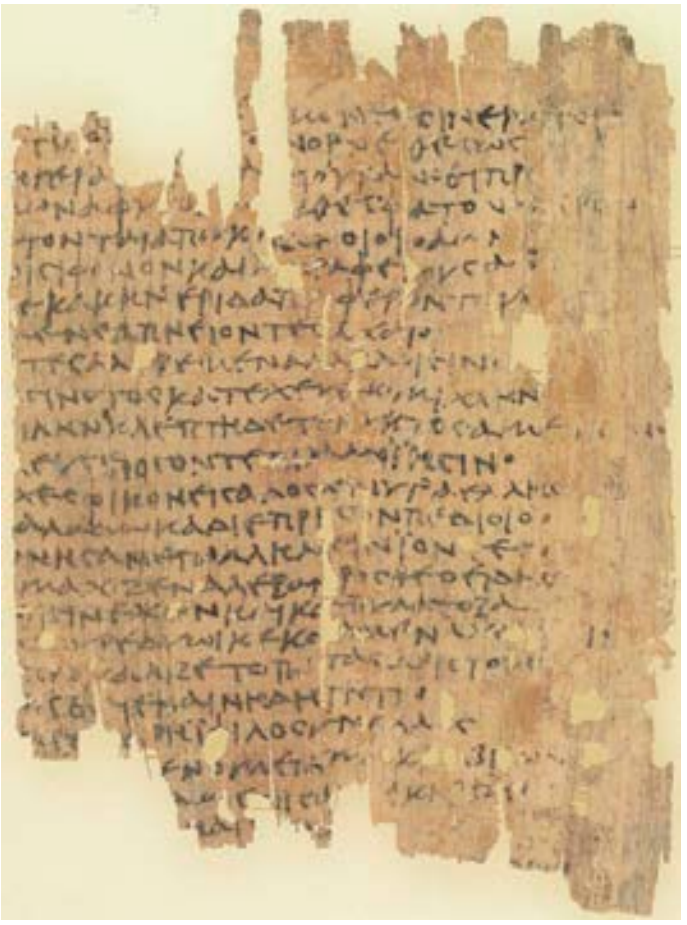

*lsv*

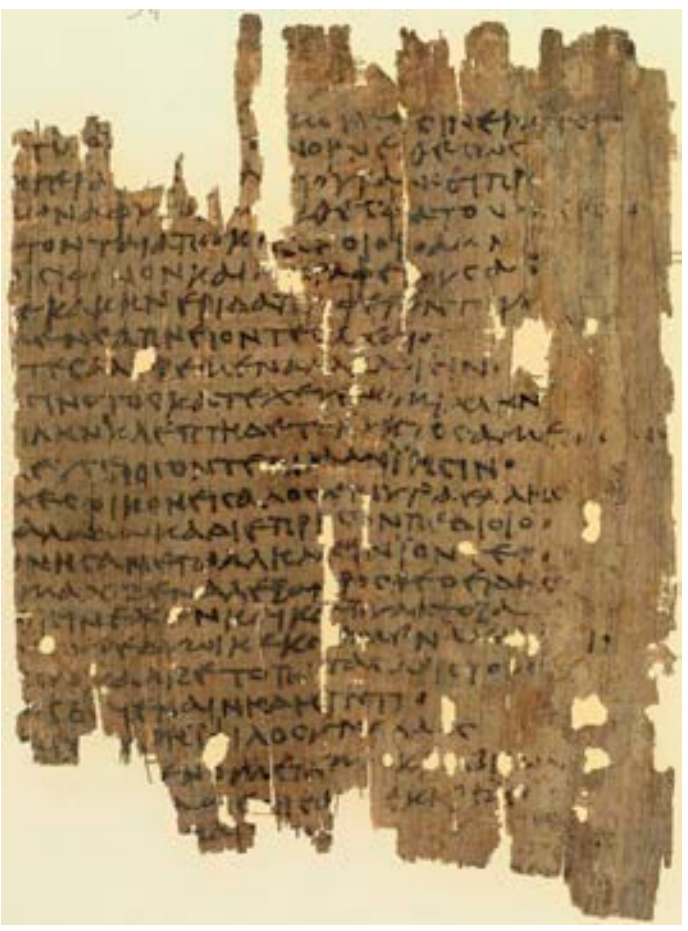

*vividness*

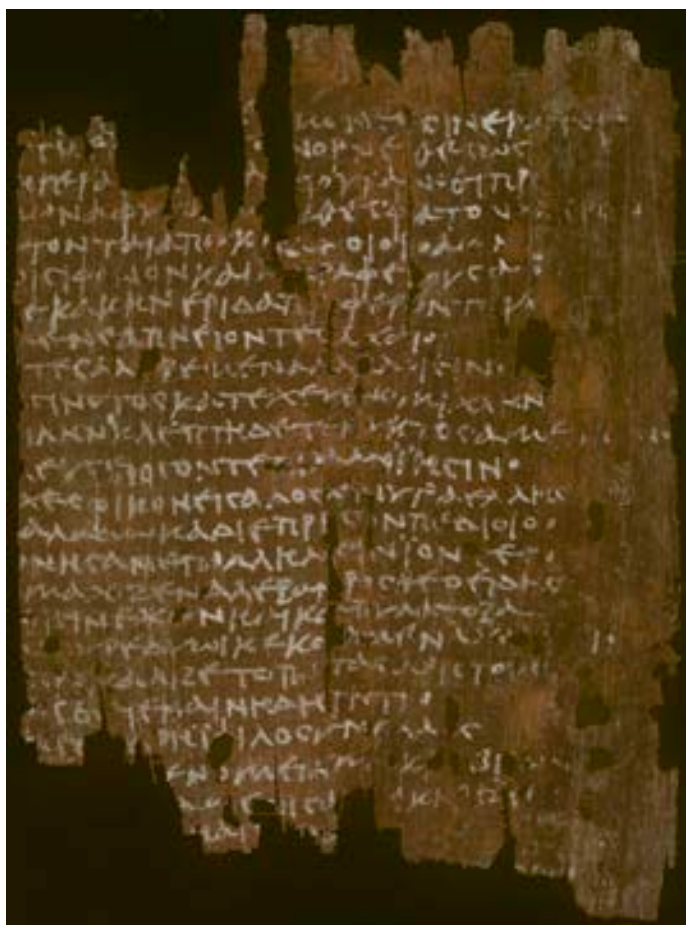

*neglsv*

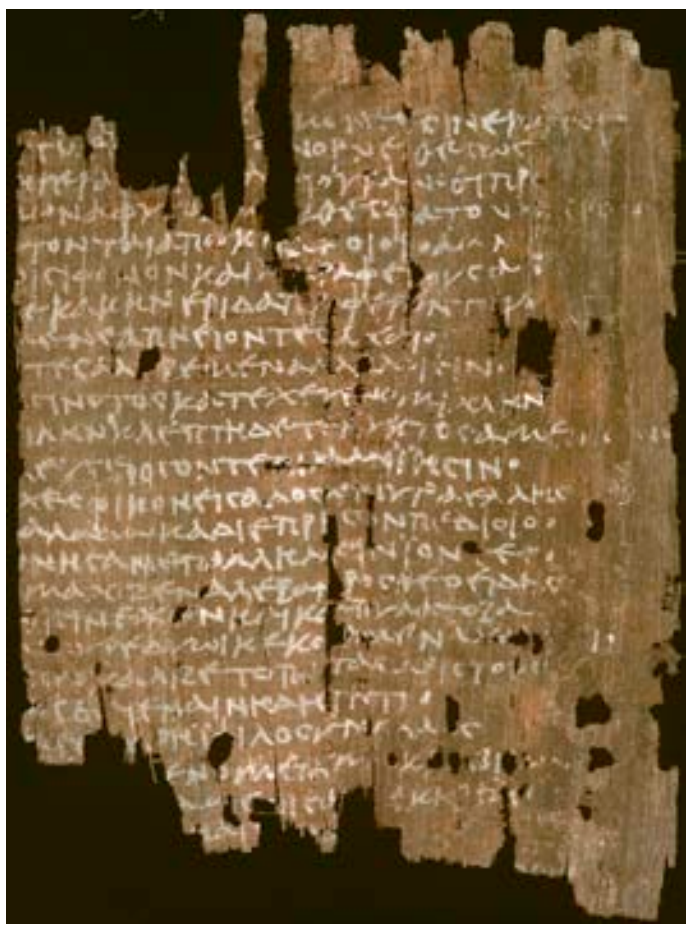

*negvividness*

**P.Mich.inv.1318v**

Credit of original papyrus reproduction: University of Michigan Library, Public Domain Mark 1.0

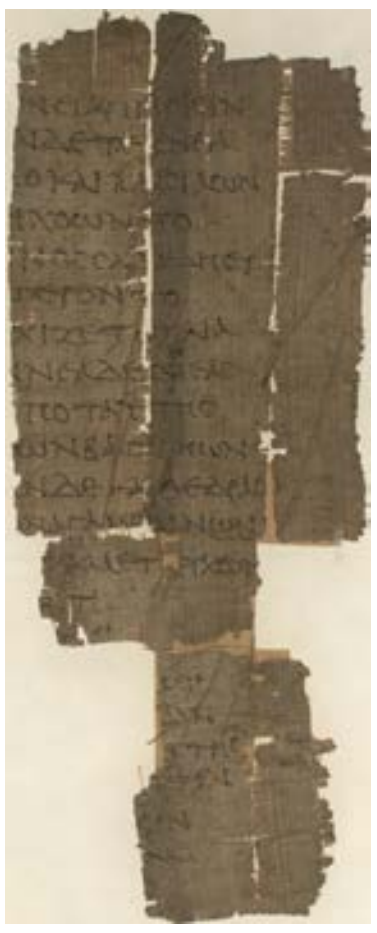
**original**

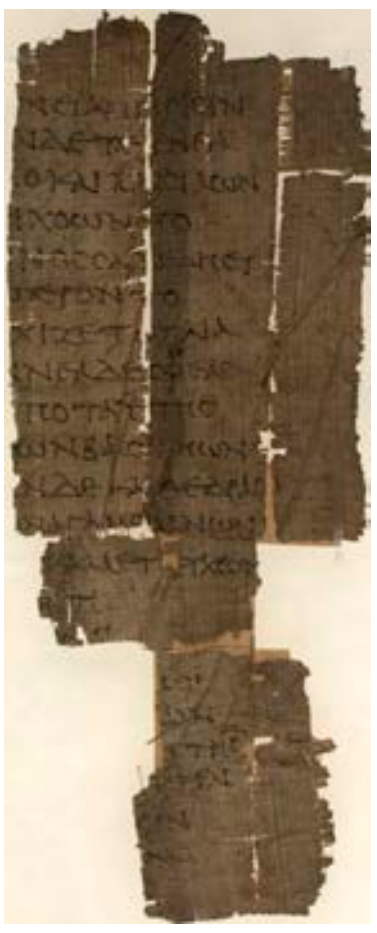
stretchlim

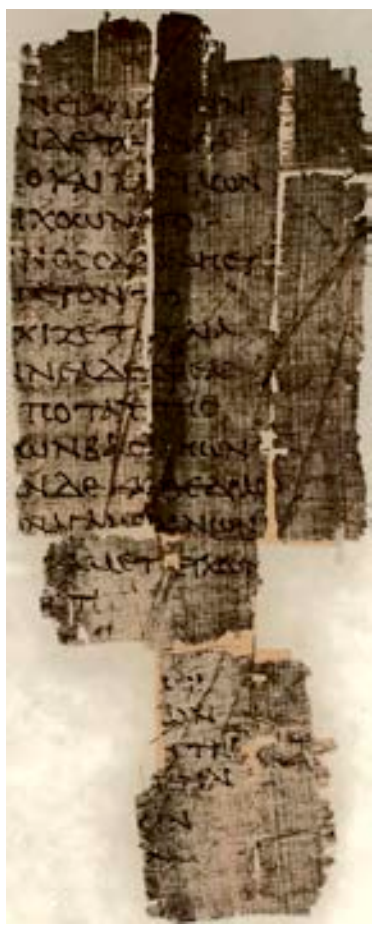
histeq

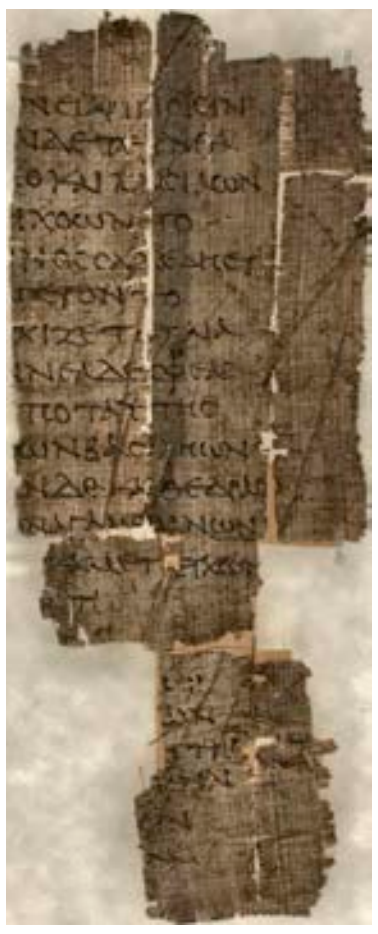
adapthisteq

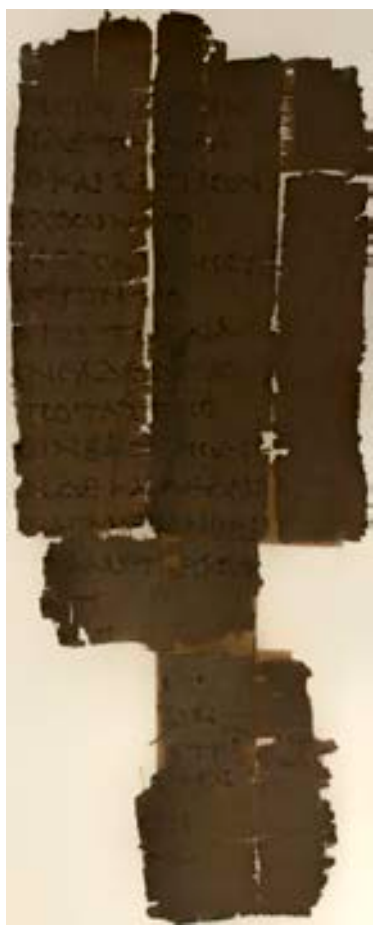
locallapfilt

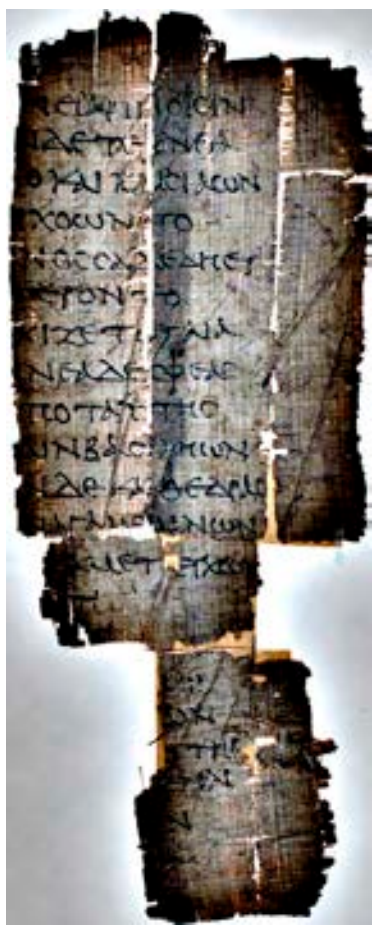
retinex

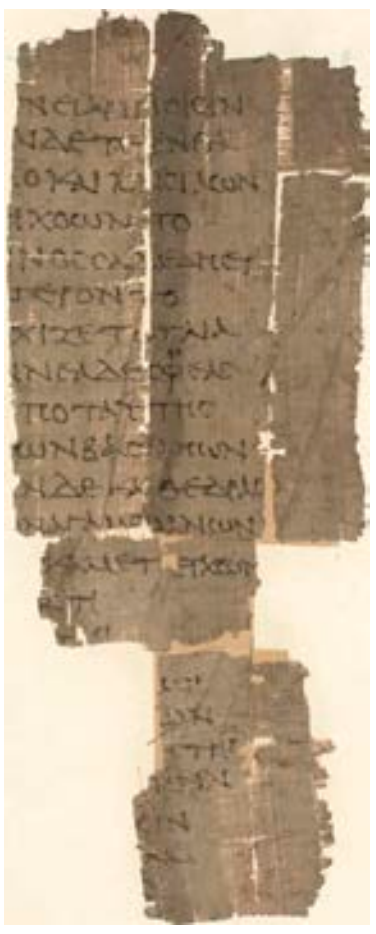
*lsv*

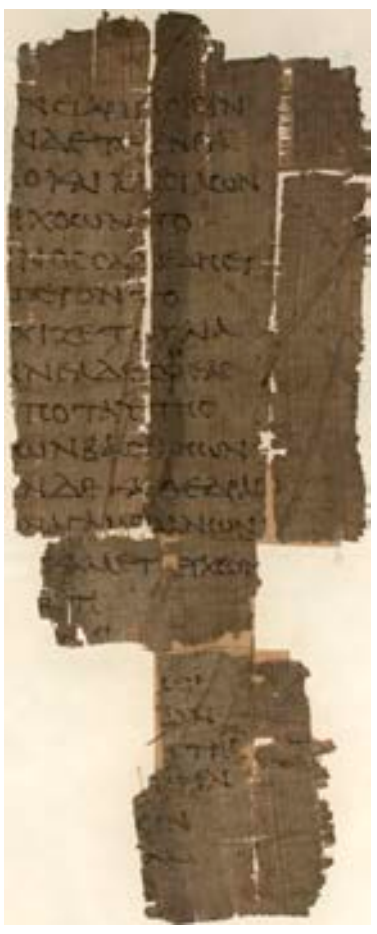
*vividness*

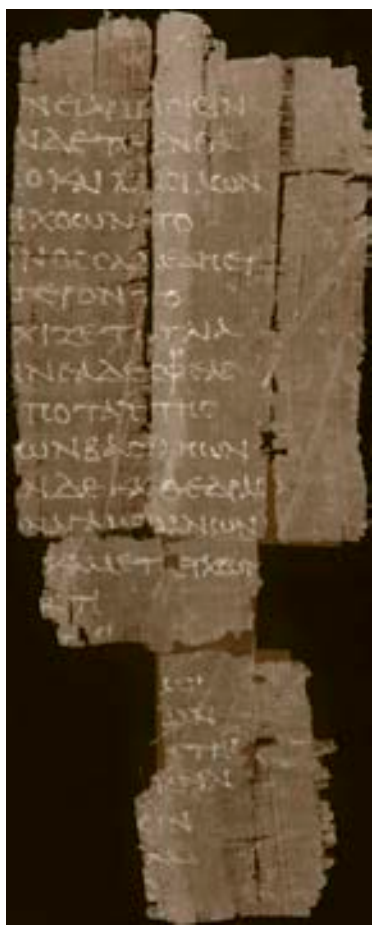
*neglsv*

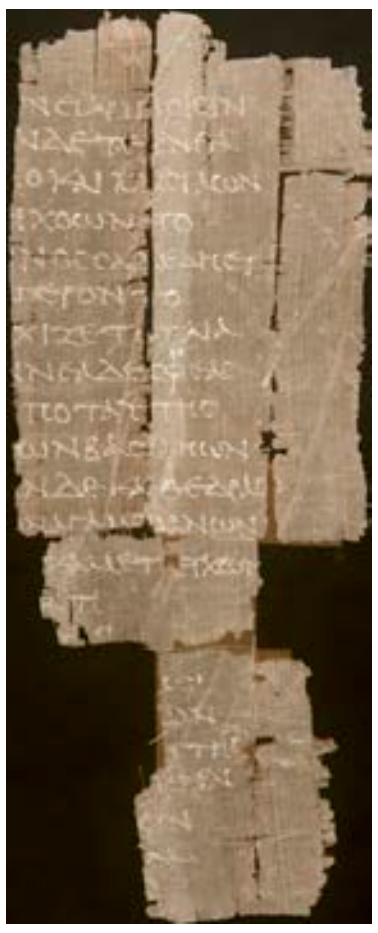
*negvividness*

**P.Mich.inv.2755**

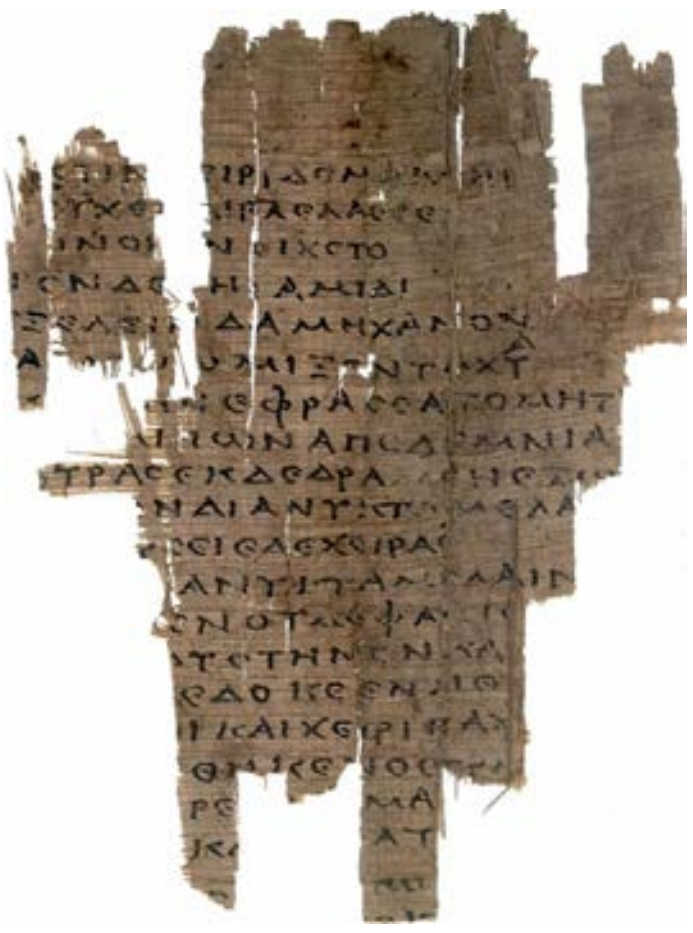

**original**

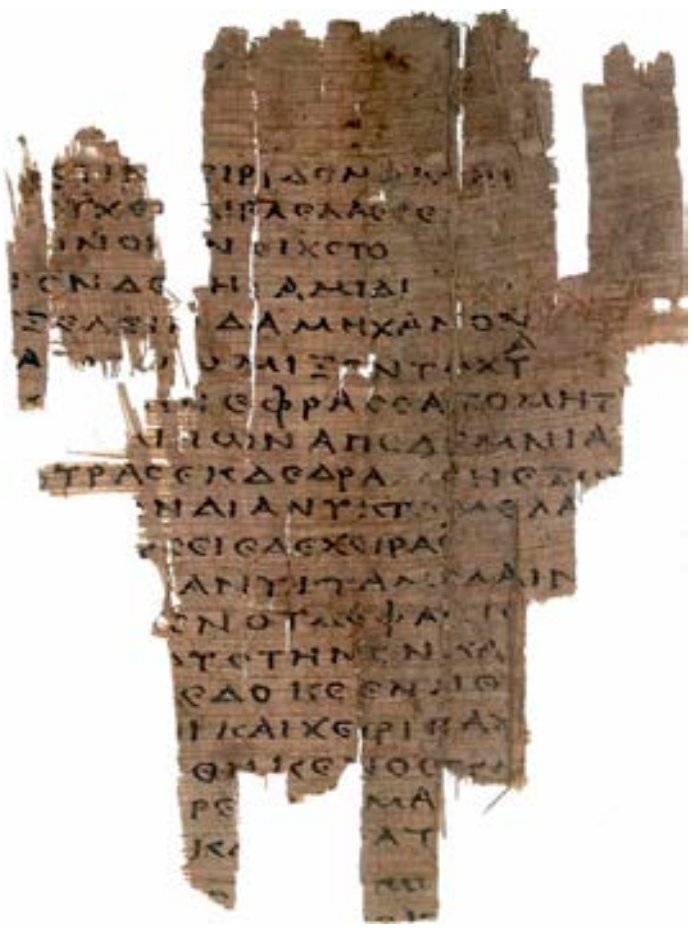

stretchlim

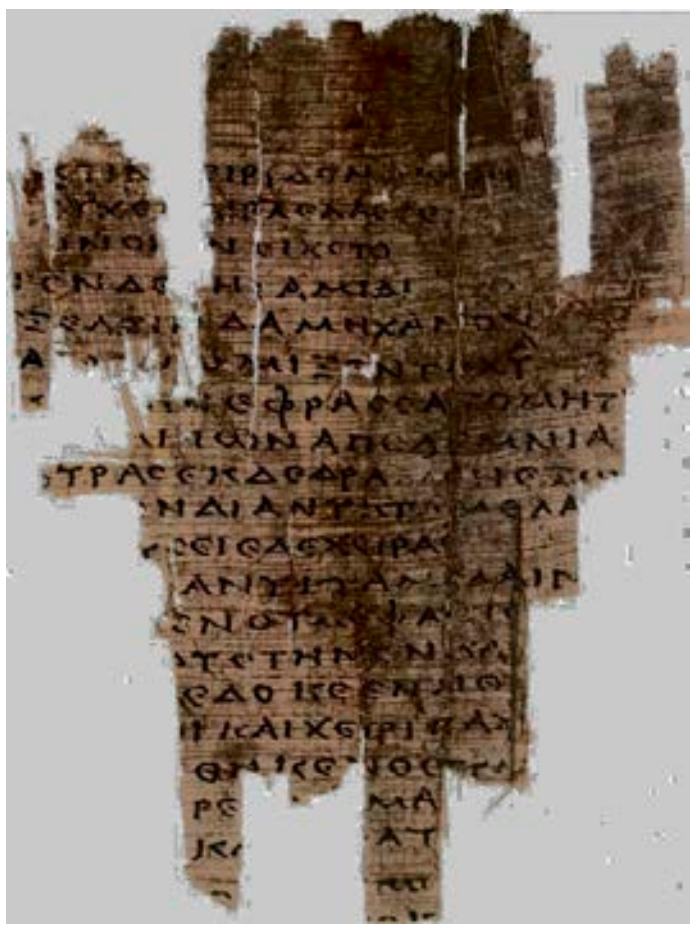

histeq

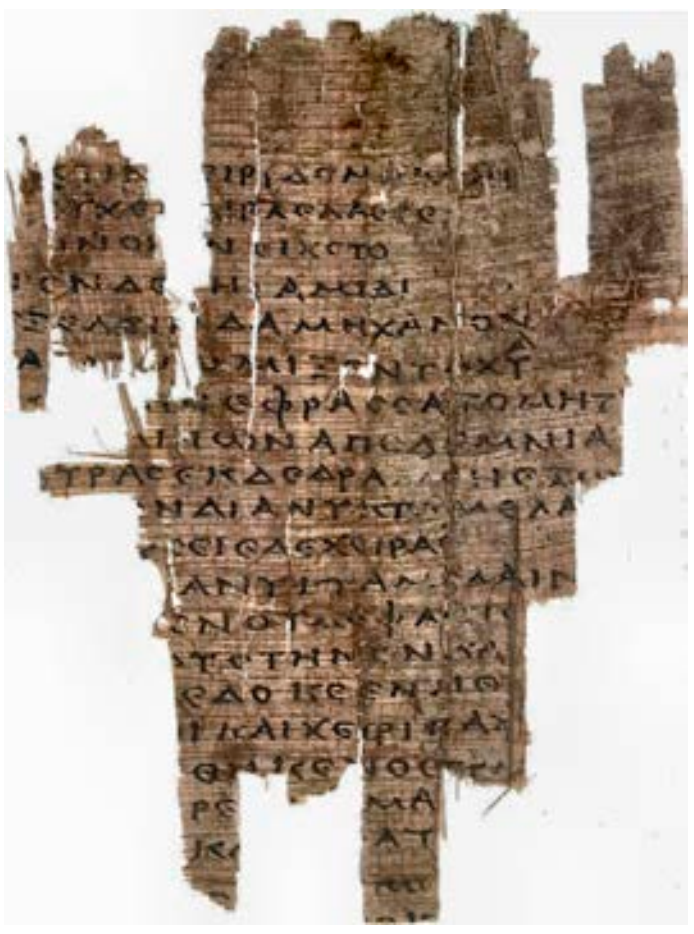

adapthisteq

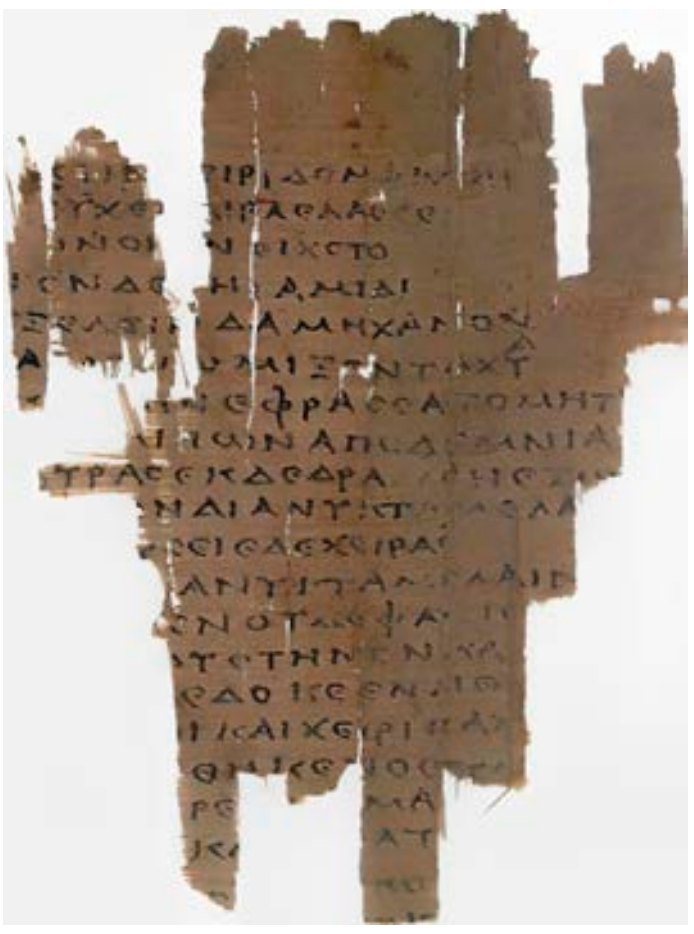

locallapfilt

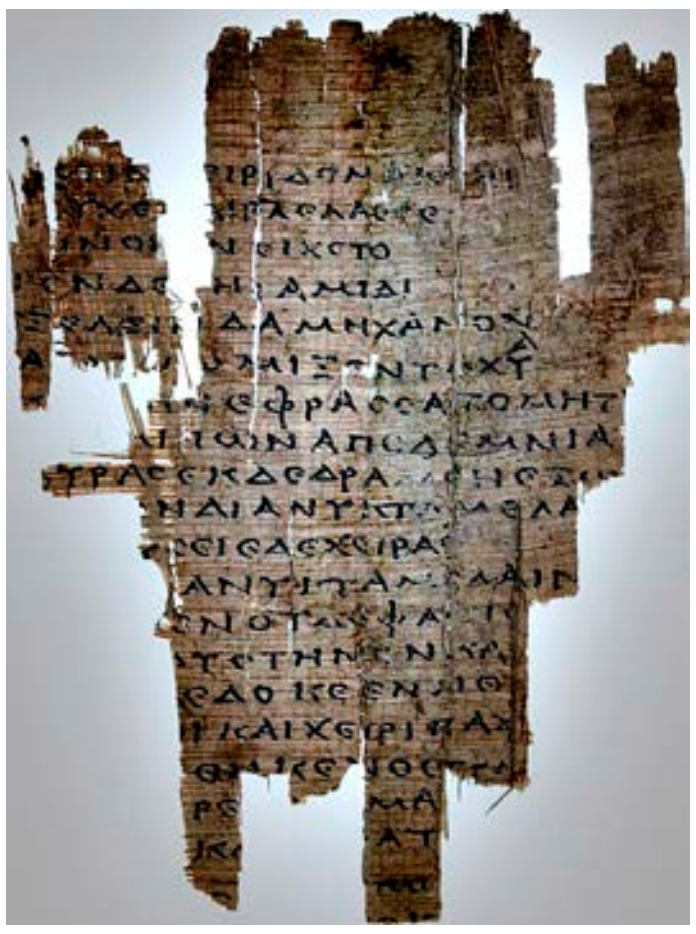

retinex

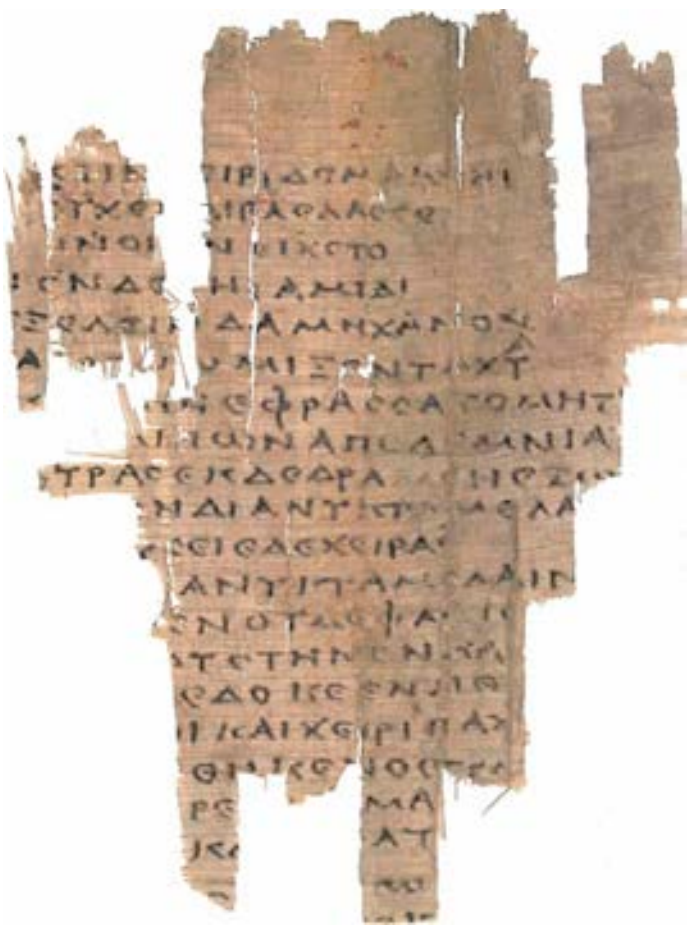

*lsv*

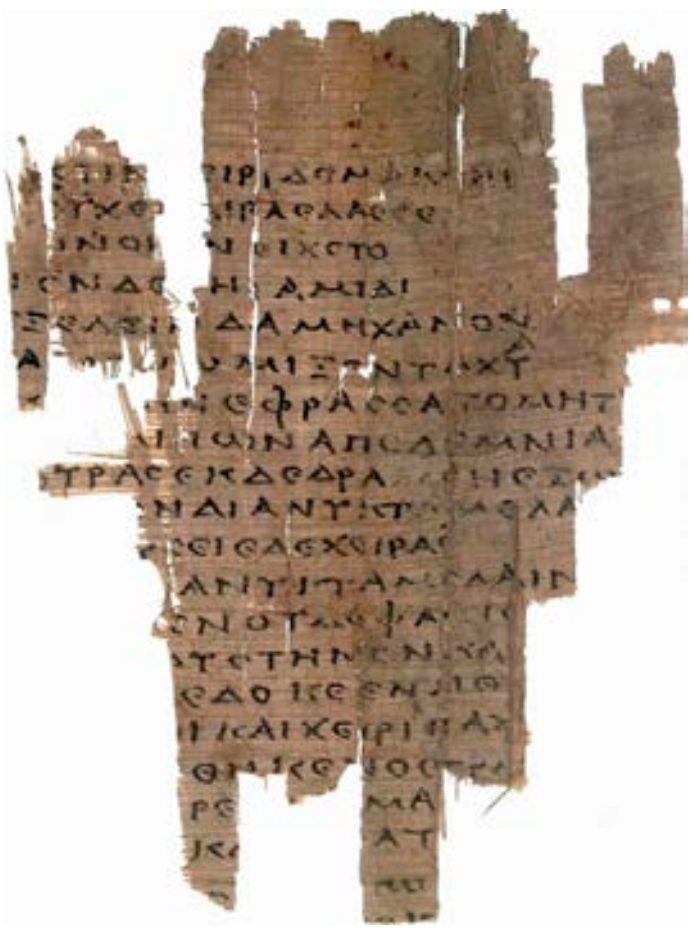

*vividness*

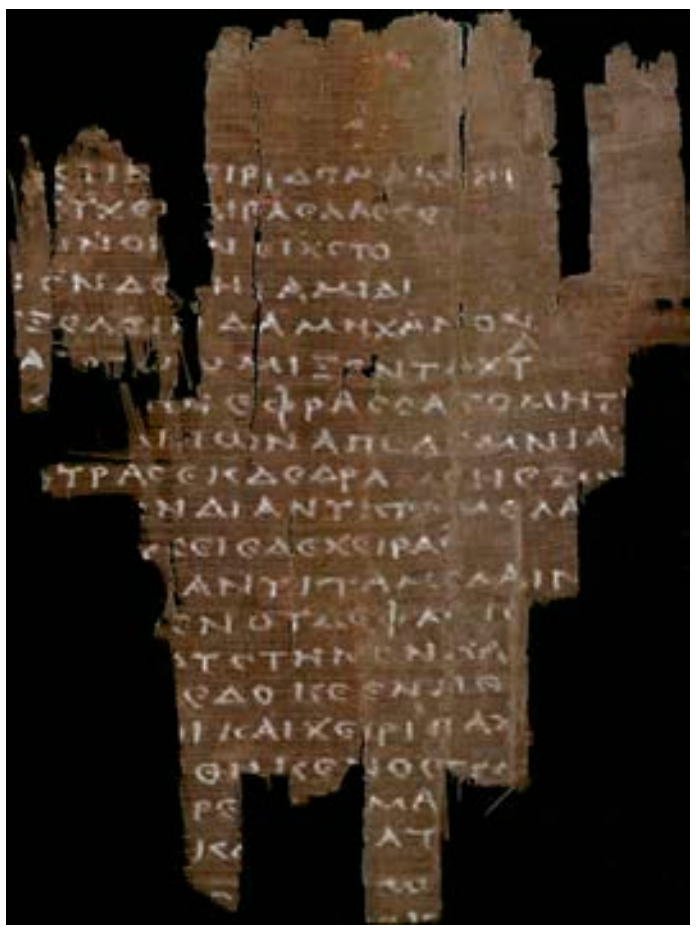

*neglsv*

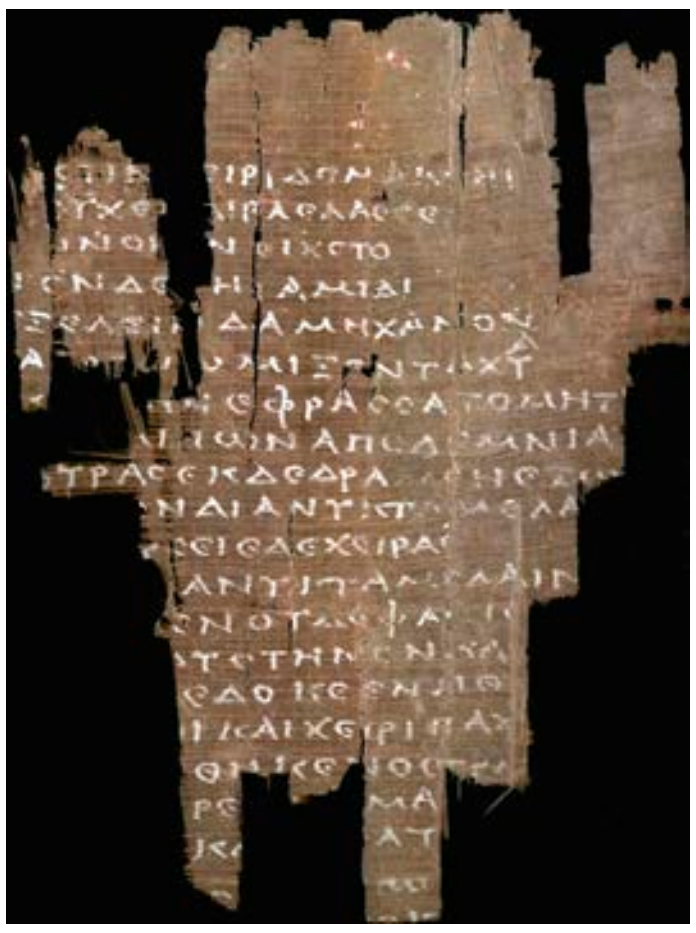

*negvividness*

**P.Oxy.XXII 2309**

Credit of original papyrus reproduction: Courtesy of The Egypt Exploration Society and the University of Oxford Imaging Papyri Project

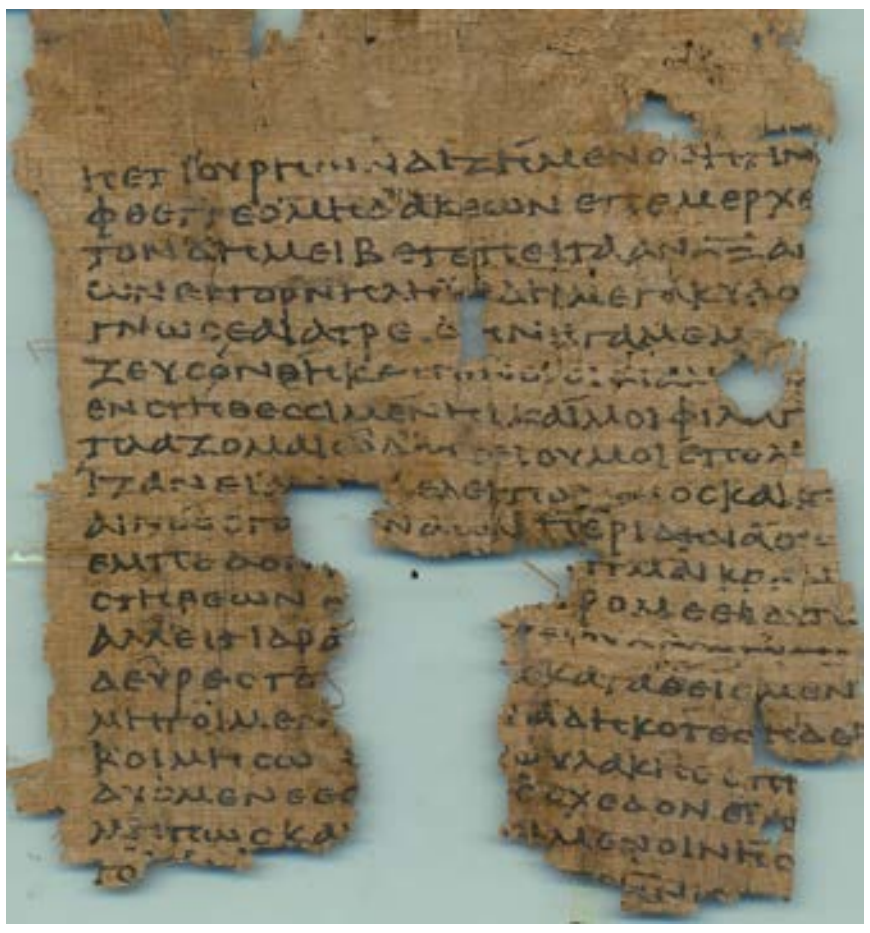

**original**

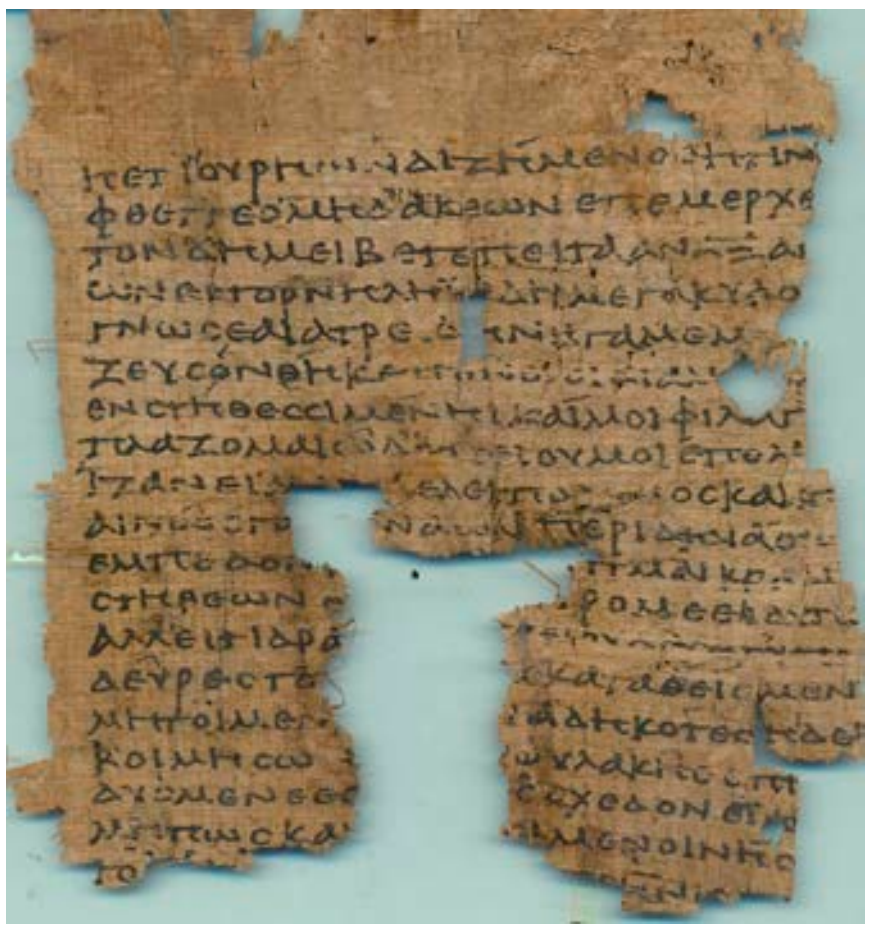

stretchlim

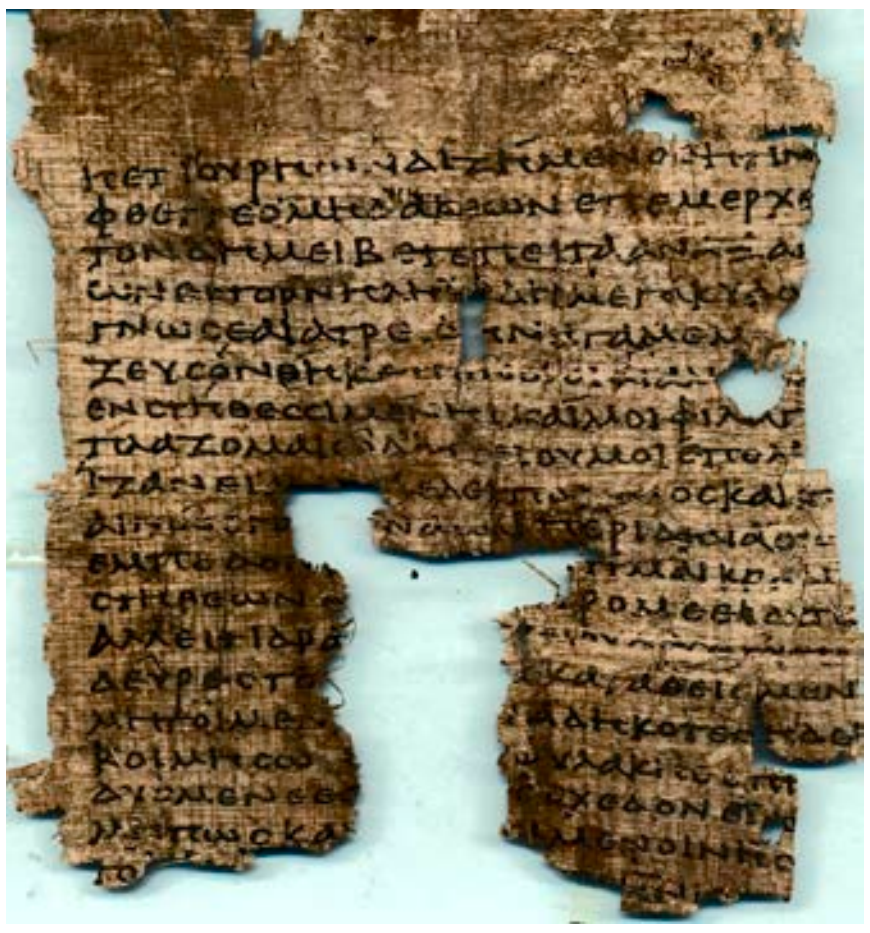

histeq

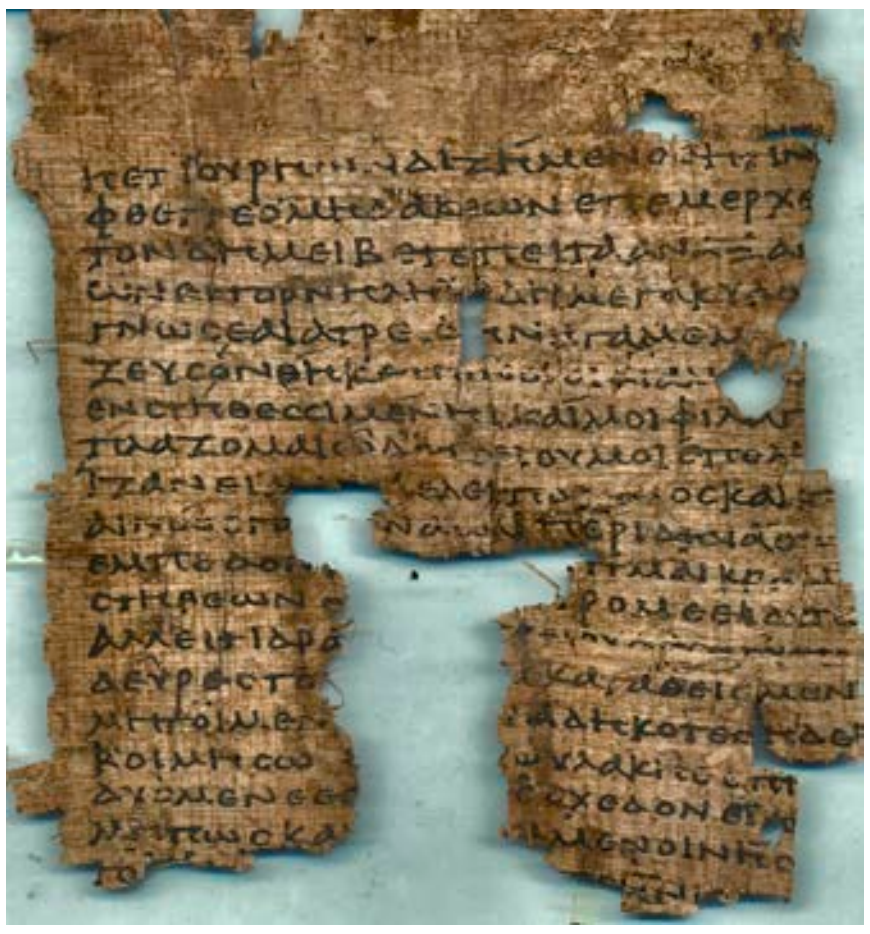

adapthisteq

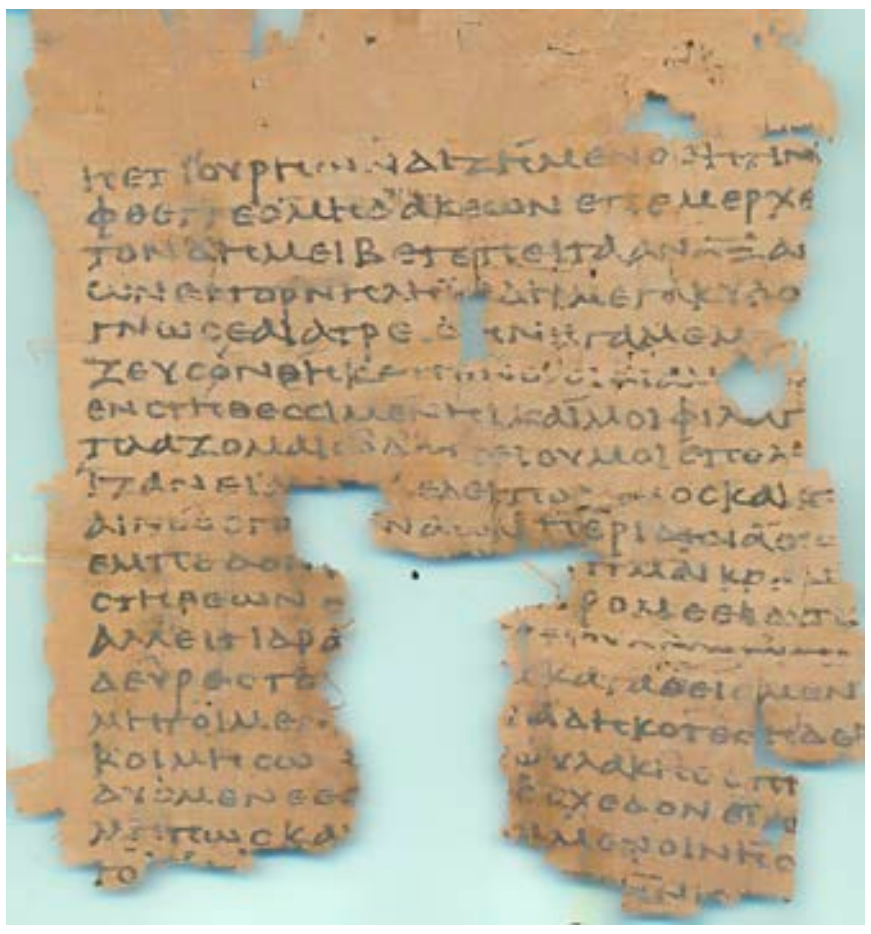

locallapfilt

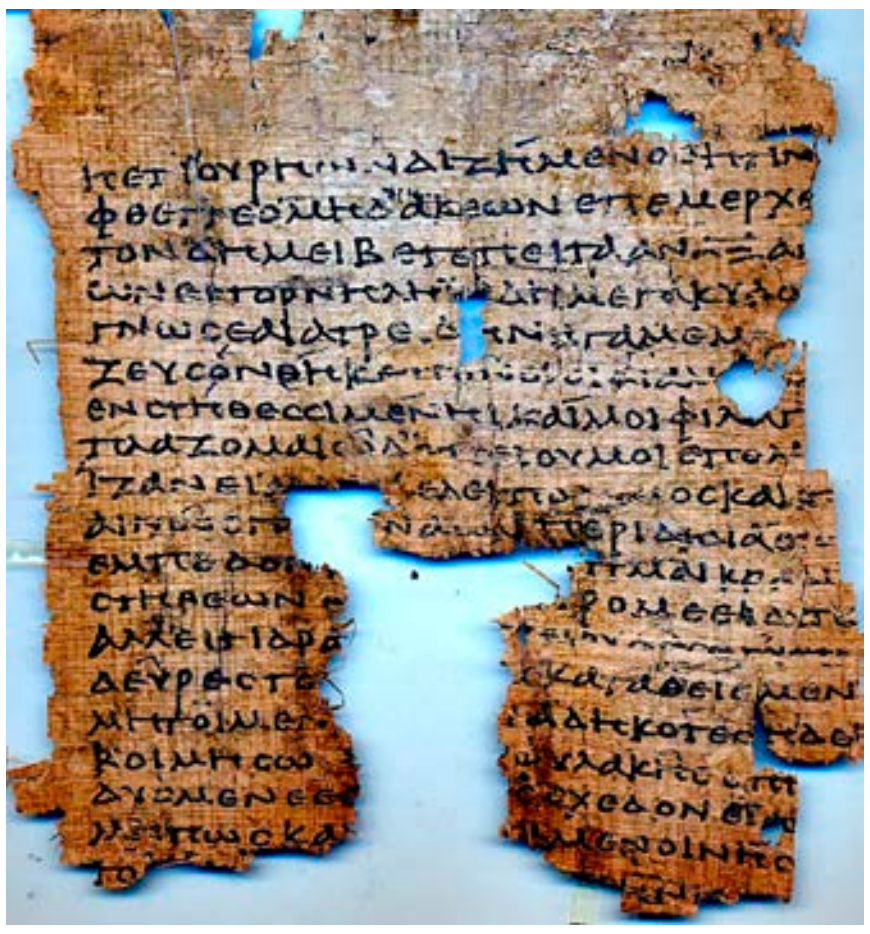

retinex

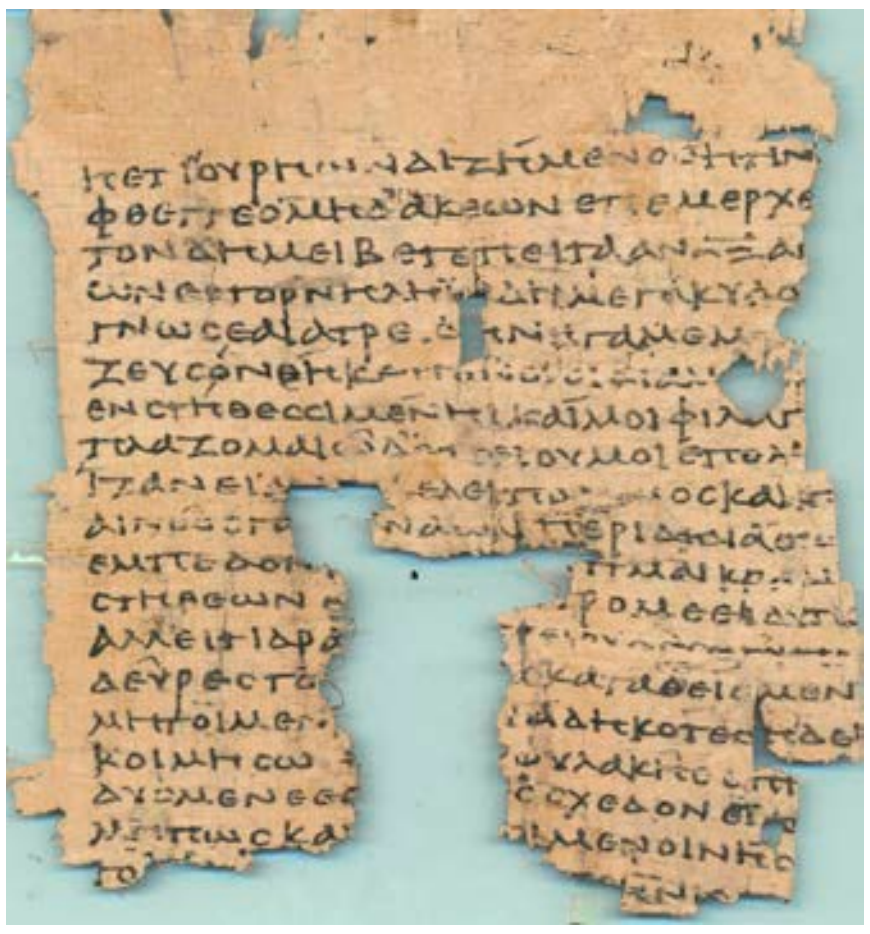

*lsv*

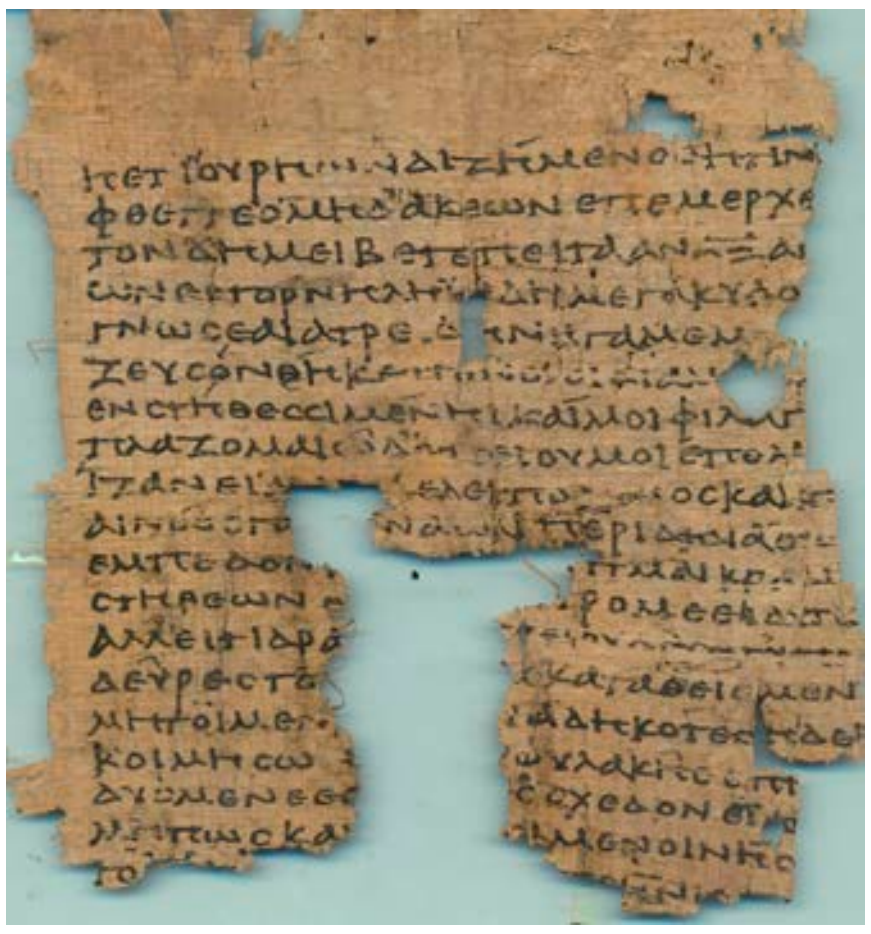

*vividness*

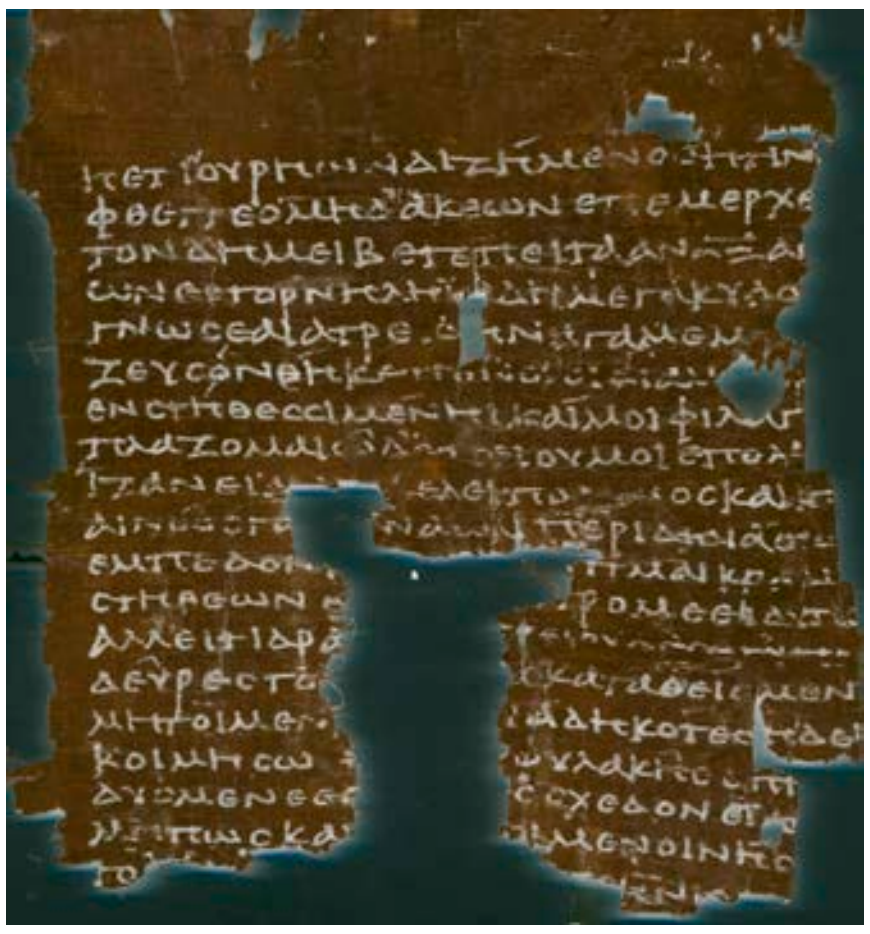

*neglsv*

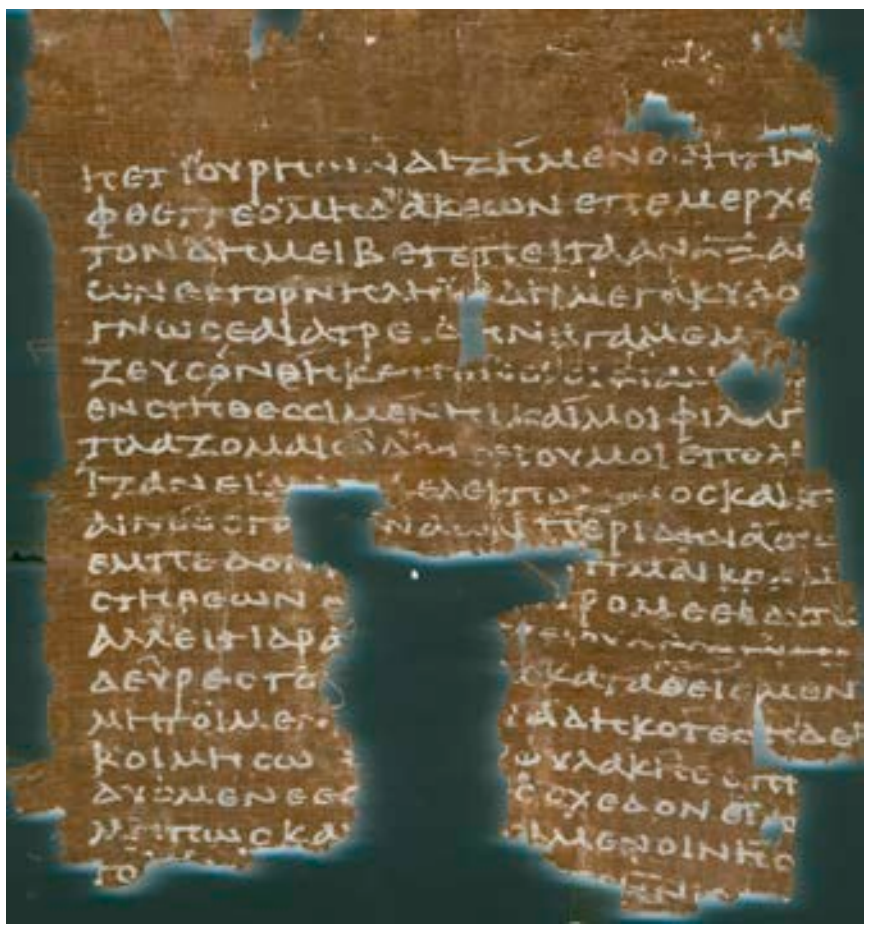

*negvividness*

**PSI XII 1274 r**

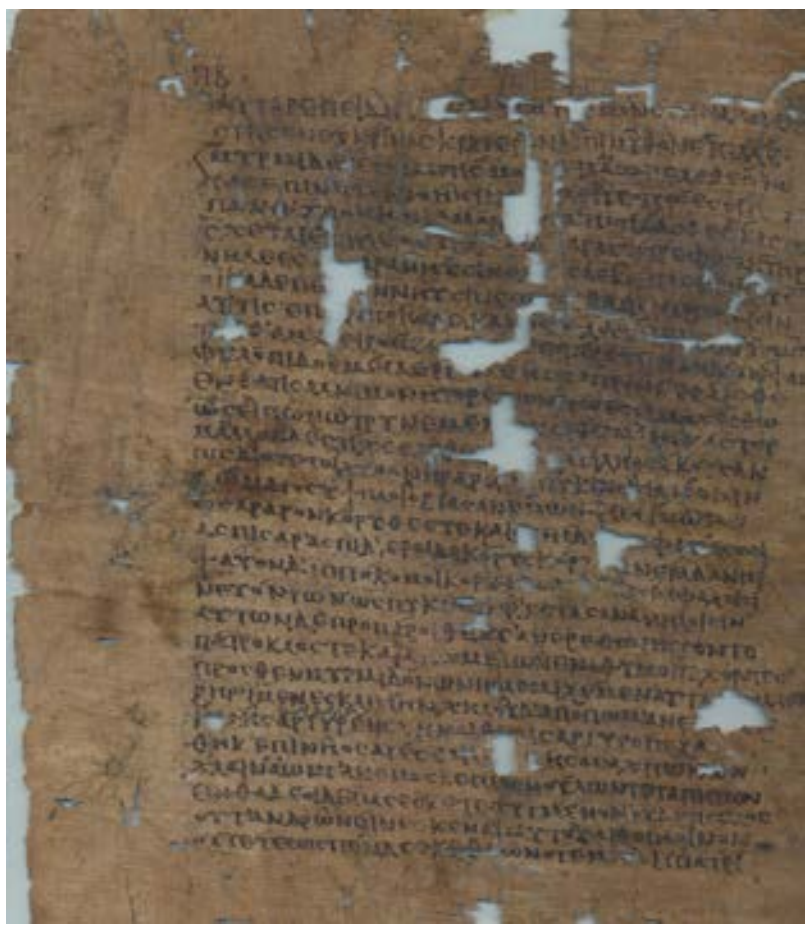

**original**

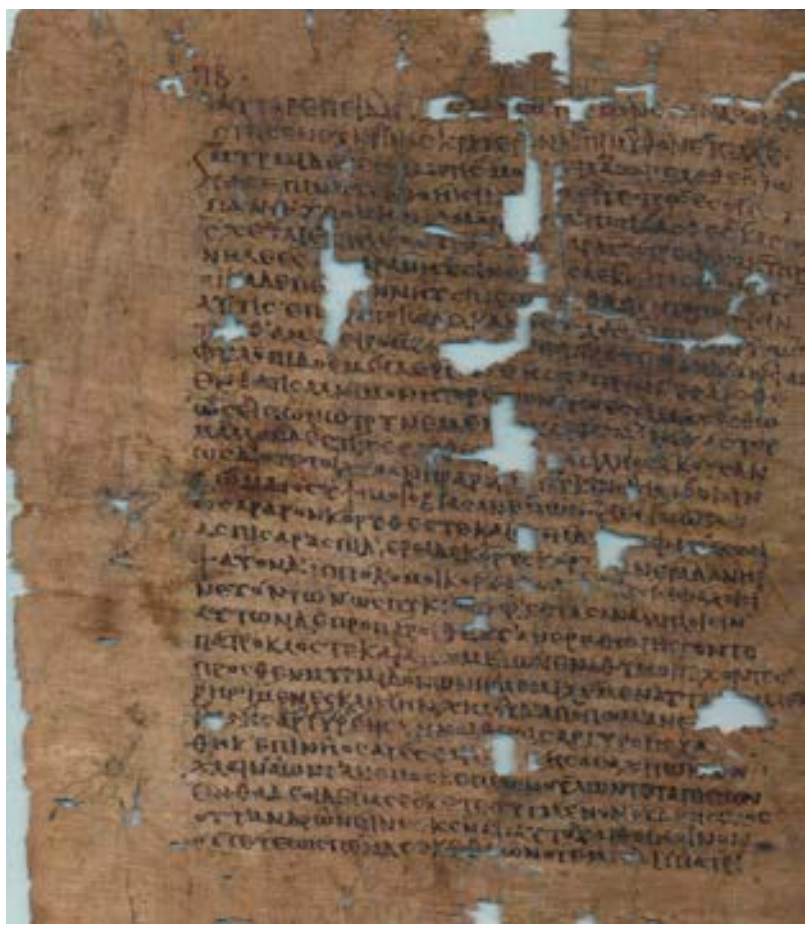

stretchlim

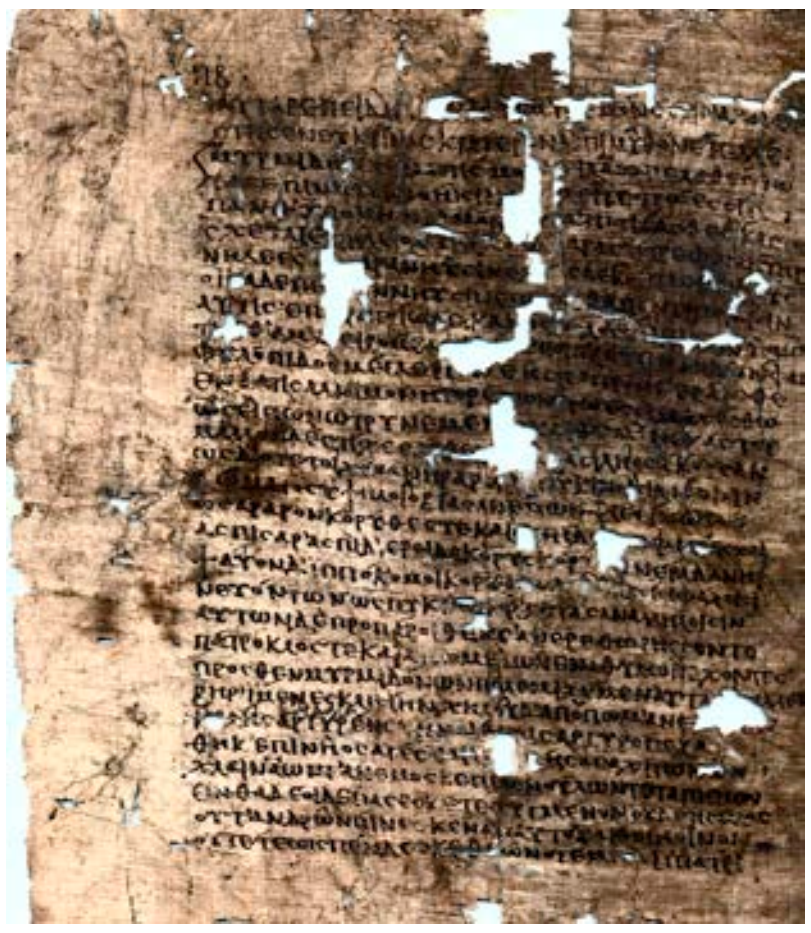

histeq

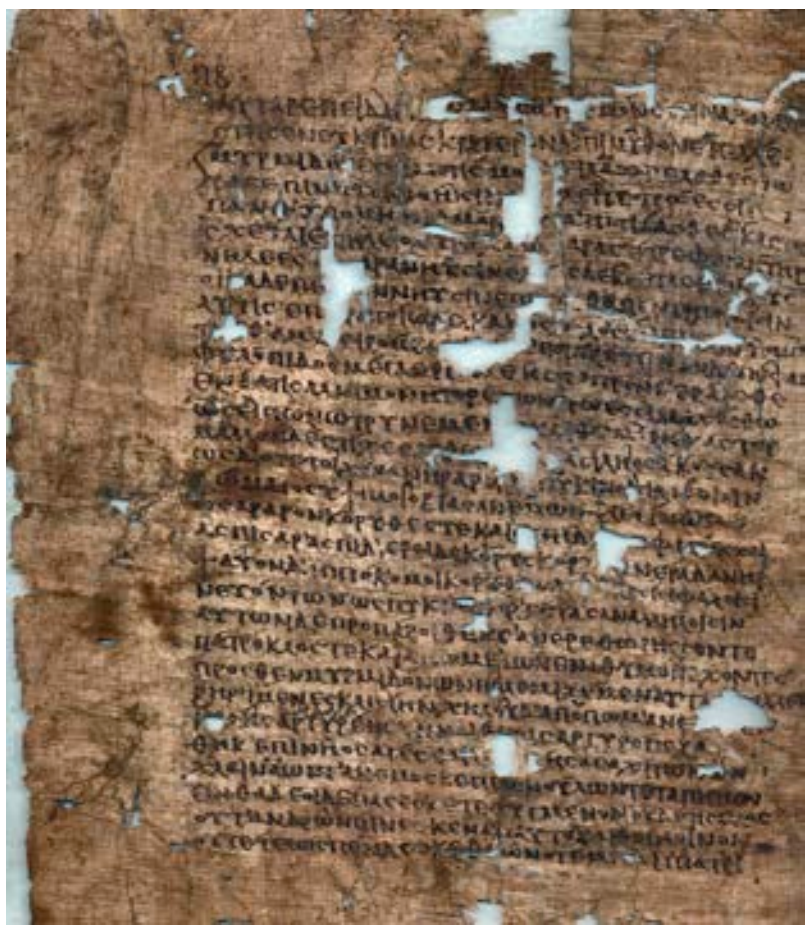

adapthisteq

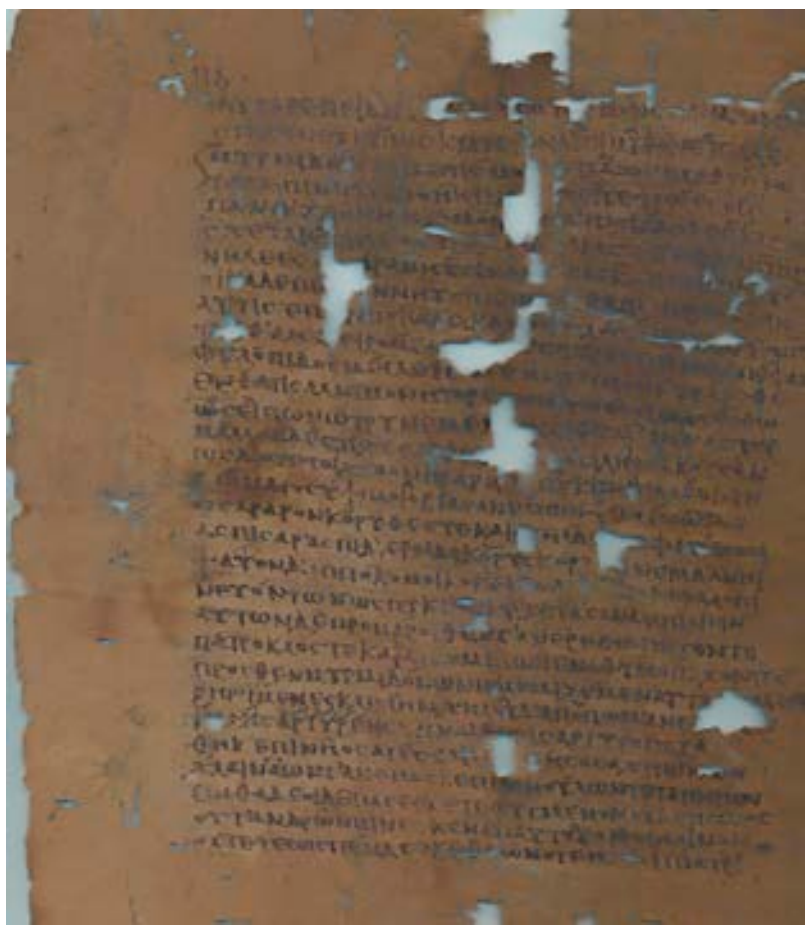

locallapfilt

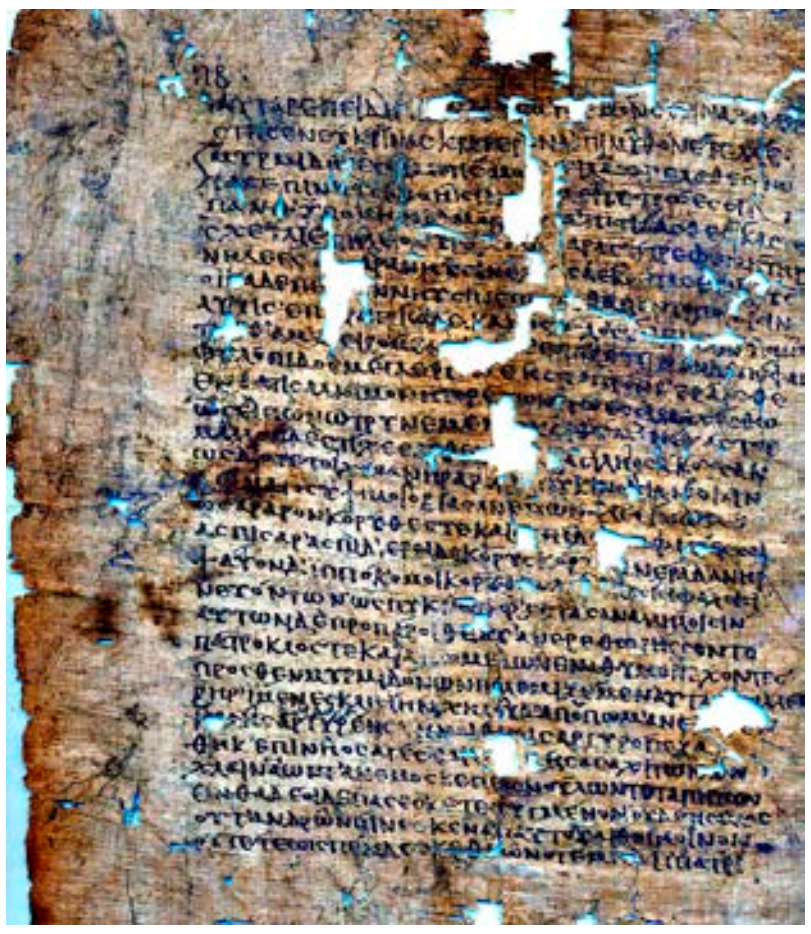

retinex

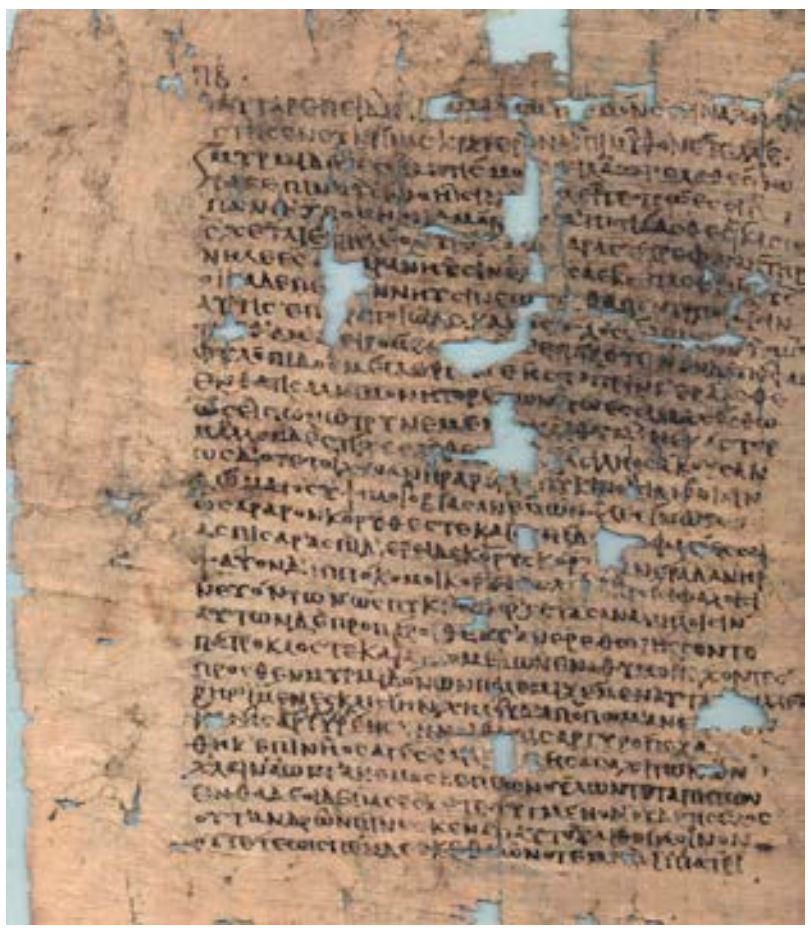

*lsv*

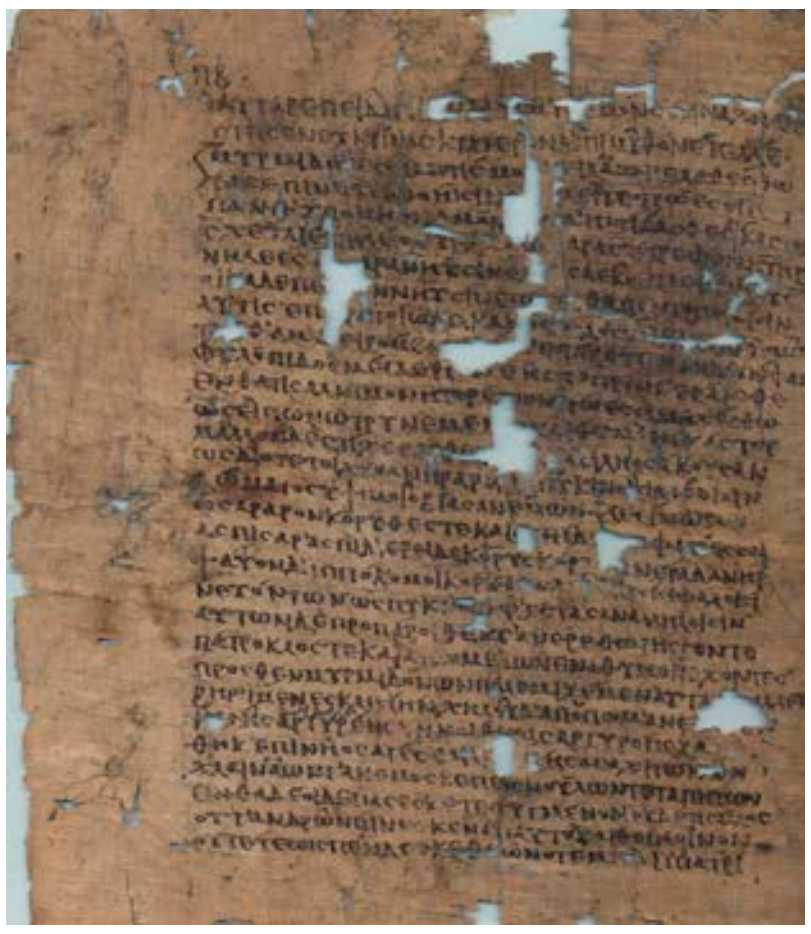

*vividness*

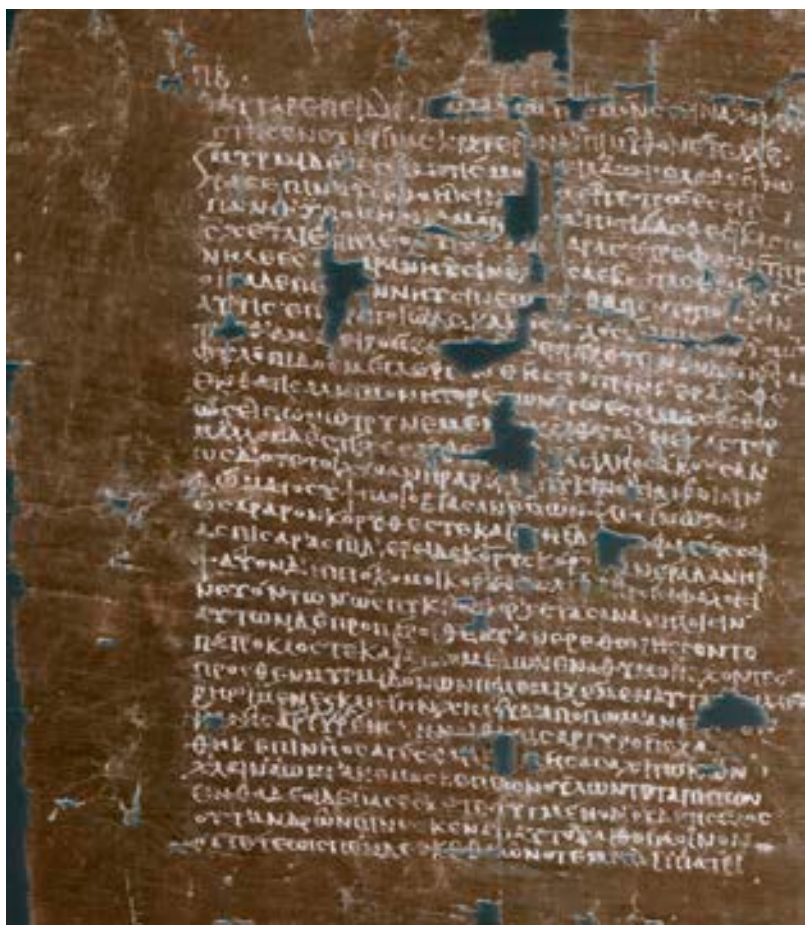

*neglsv*

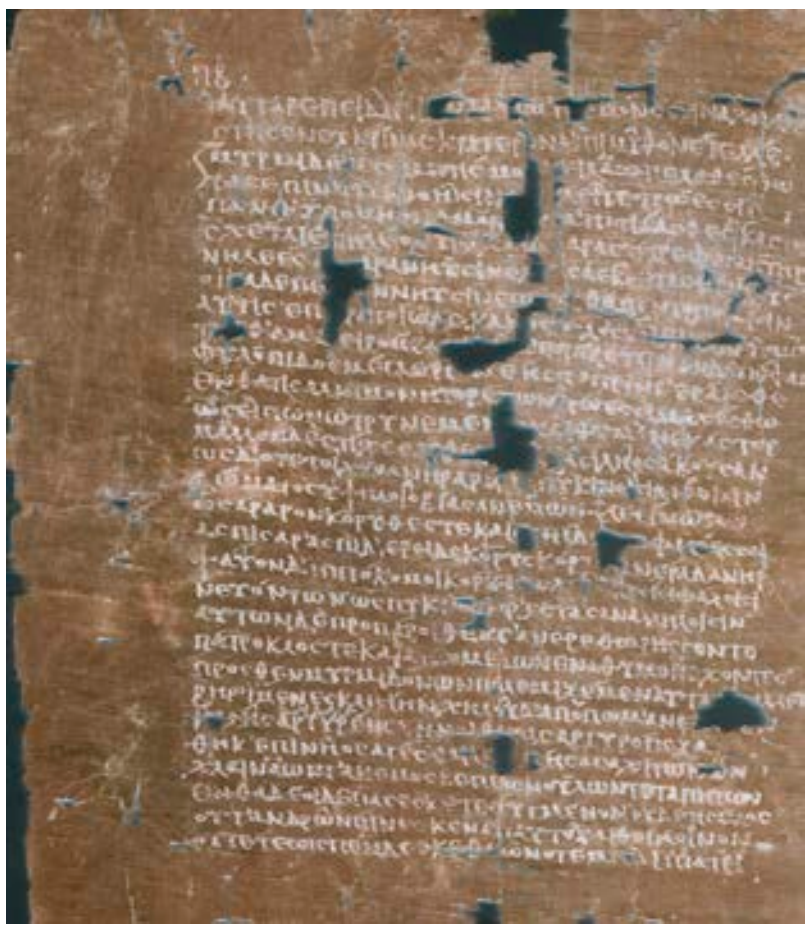

*negvividness*

**PSI XIII 1298 (15a) r 1**

Credit of original papyrus reproduction: Istituto Papirologico Vitelli, by permission

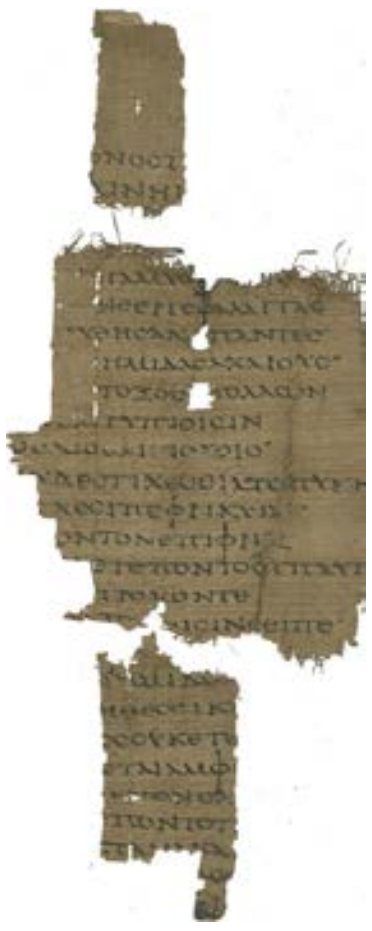

**original**

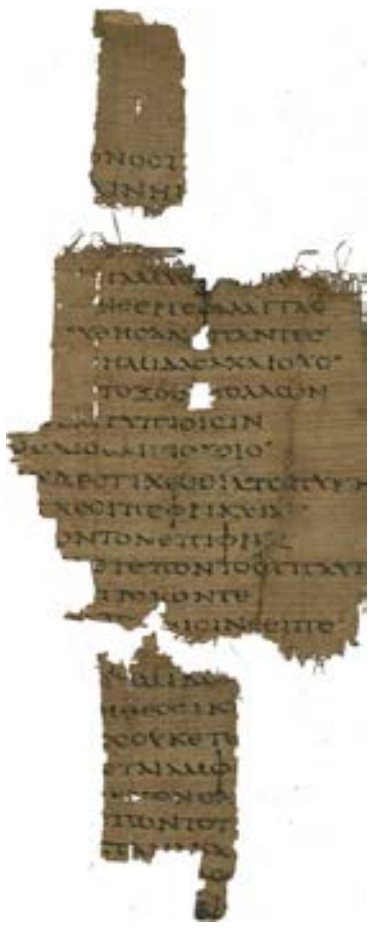

stretchlim

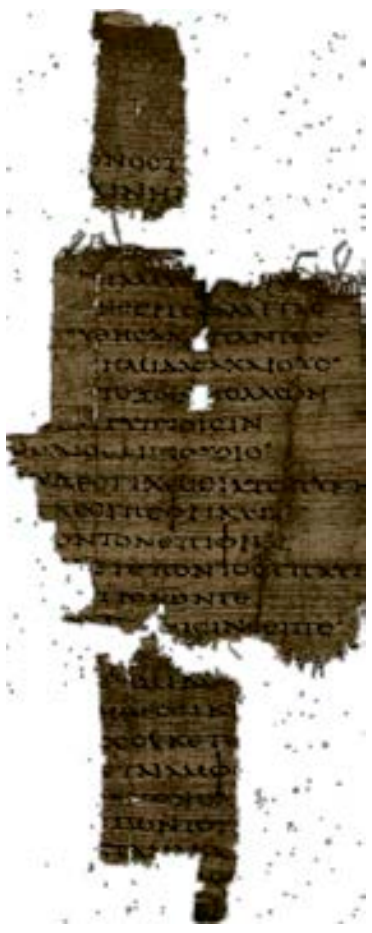

histeq

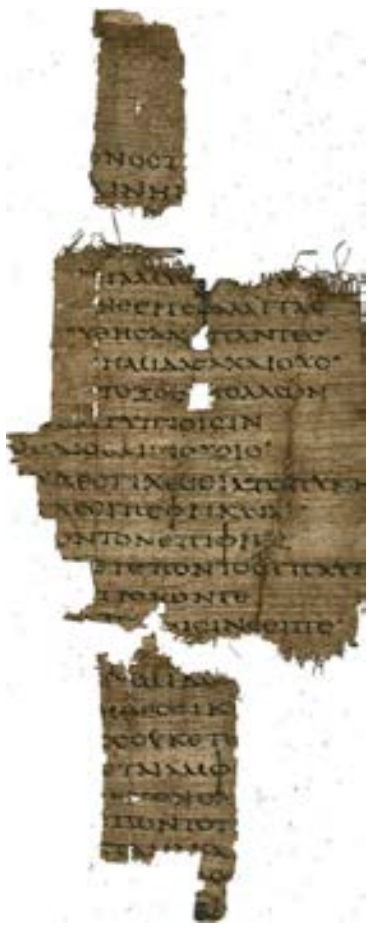

adapthisteq

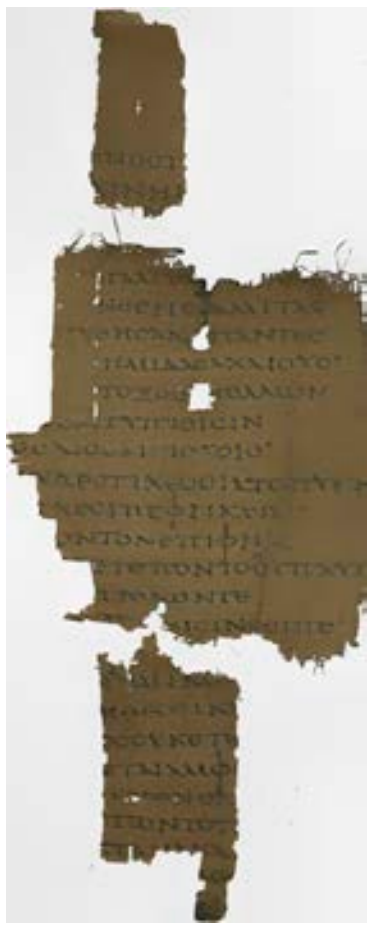

locallapfilt

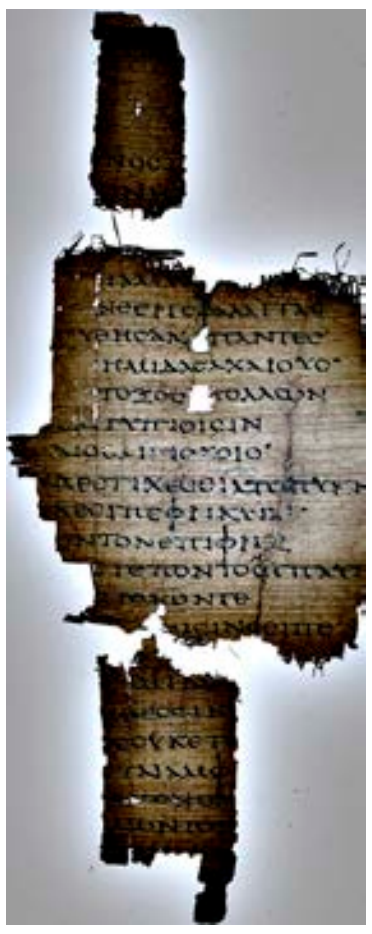

retinex

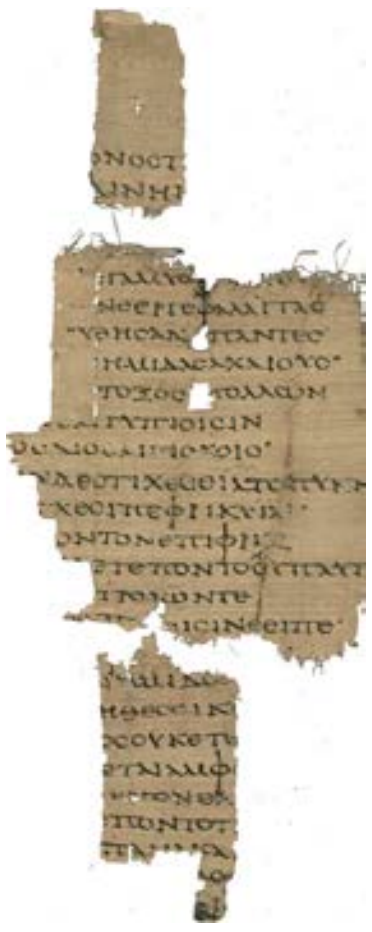

*lsv*

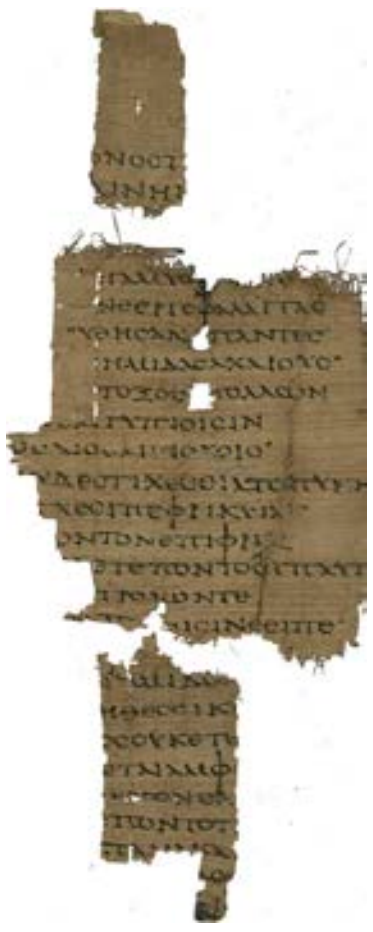

*vividness*

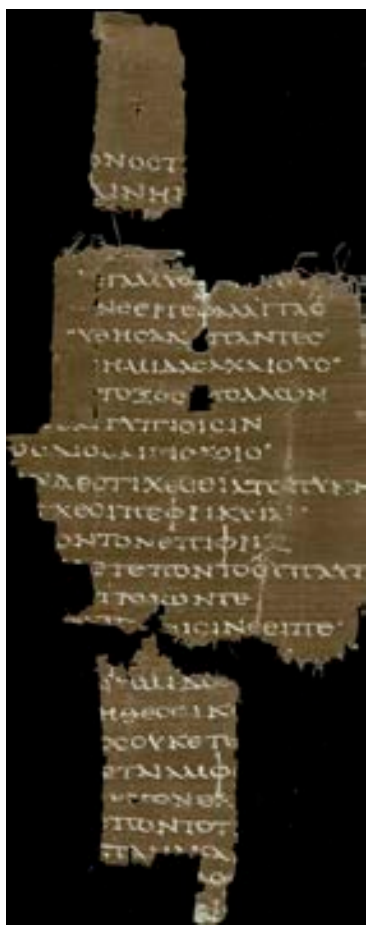

*neglsv*

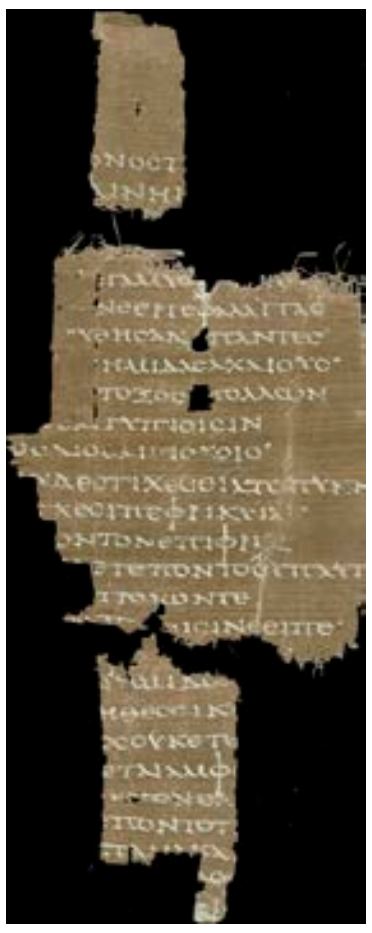

*negvividness*

**PSI XIV 1376 r**